\DocumentMetadata{}
\documentclass[acmlarge, authorsperline]{acmart}
\usepackage{graphicx}
\usepackage{longtable}
\usepackage{caption}
\usepackage{booktabs}
\usepackage{subcaption}
\usepackage{array}
\usepackage{float}
\usepackage{makecell}
\usepackage{mathtools, amsmath}
\usepackage{relsize}
\usepackage{lscape}
\usepackage{multicol, multirow}
\usepackage{enumitem}
\usepackage{tablefootnote}
\usepackage{xcolor}
\usepackage{amsmath,amsfonts}
\usepackage{algorithmic}
\usepackage{algorithm}
\usepackage{array}
\usepackage{wrapfig}
\usepackage[symbol]{footmisc}
\usepackage{bm} % allows for \bm bold math
\usepackage{url}

\definecolor{maroon}{cmyk}{0,0.87,0.68,0.32}

\newcommand{\anonymize}[1]{Study}

\newlist{questions}{enumerate}{2}

\setlist[questions,1]{label=\bf{RQ\arabic*}.,ref=RQ\arabic*}

\newcommand{\edit}[1]{{\color{black} #1}}
\AtBeginDocument{%
  \providecommand\BibTeX{{%
    \normalfont B\kern-0.5em{\scshape i\kern-0.25em b}\kern-0.8em\TeX}}}

\begin{document}


\setcopyright{acmlicensed}
\acmJournal{IMWUT}
\acmYear{2025} \acmVolume{9} \acmNumber{3} \acmArticle{126} \acmMonth{9}\acmDOI{10.1145/3749494}

% \title[Foundation Model]{A PPG Foundation Model Suitable for a Diverse Array of Mobile Health Applications in both Lab and Field Environments}
\title[Foundation Model]{Pulse-PPG: An Open-Source Field-Trained PPG Foundation Model for Wearable Applications Across Lab and Field Settings}


\author{Mithun Saha\textsuperscript{$\bm{\dagger}$}}
\email{msaha1@memphis.edu}
\affiliation{%
  \institution{University of Memphis}
  \city{Memphis}
  \state{Tennessee}
  \postcode{38152}
  \country{USA}
}

\author{Maxwell A. Xu\textsuperscript{$\bm{\dagger}$}}
\email{maxu@illinois.edu}
\affiliation{%
  \institution{University of Illinois Urbana-Champaign}
  \city{Urbana}
  \state{Illinois}
  \postcode{61801}
  \country{USA}
}

\author{Wanting Mao}
\email{wmao8@illinois.edu}
\affiliation{%
  \institution{University of Illinois Urbana-Champaign}
  \city{Urbana}
  \state{Illinois}
  \postcode{61801}
  \country{USA}
}

\author{Sameer Neupane}
\email{sameer.neupane@memphis.edu}
\affiliation{%
  \institution{University of Memphis}
  \city{Memphis}
  \state{Tennessee}
  \postcode{38152}
  \country{USA}
}

\author{James M. Rehg}
\email{jrehg@illinois.edu}
\affiliation{%
  \institution{University of Illinois Urbana-Champaign}
  \city{Urbana}
  \state{Illinois}
  \postcode{61801}
  \country{USA}
}  

\author{Santosh Kumar}
\email{skumar4@memphis.edu}
\affiliation{%
  \institution{University of Memphis}
  \city{Memphis}
  \state{Tennessee}
  \postcode{38152}
  \country{USA}
}

\thanks{\textbf{\textsuperscript{$\bm{\dagger}$}Both co-first authors contributed equally to this research.}}
% \thanks{{Corresponding Authors:} \textit{amimithun@gmail.com, maxu@illinois.edu, jrehg@illinois.edu, skumar4@memphis.edu}}

% \thanks{\noindent\rule{\linewidth}{0.4pt}}

% \thanks{$^{*}$Both authors contributed equally to this research.}
\renewcommand{\shortauthors}{Saha \& Xu et al.}
\begin{abstract}
Photoplethysmography (PPG)-based foundation models are gaining traction due to the widespread use of PPG in biosignal monitoring and their potential to track diverse health indicators. In this paper, we introduce Pulse-PPG, an open-source PPG foundation model trained exclusively on raw PPG data collected over a 100-day field study with 120 participants. Existing open-source PPG foundation models are trained on clinical data, and those trained on field data are closed source, limiting their applicability in real-world settings. Extensive evaluations demonstrate that Pulse-PPG, trained on uncurated field data, exhibits superior generalization and performance across clinical and mobile health applications in both lab and field settings, when compared with state-of-the-art PPG foundation models trained on clinical data. Exposure to real-world variability in field-collected PPG data enables Pulse-PPG to learn \edit{more robust} representations. Furthermore, pre-training Pulse-PPG on field data outperforms its \edit{own} pre-training on clinical data in many tasks, reinforcing the importance of training on real-world datasets. To encourage further advancements in robust \edit{PPG modeling}, we \edit{have open-sourced\footref{open}our} Pulse-PPG \edit{model}, 
providing researchers with a \edit{valuable} resource for developing \edit{the next generation of task-specific} PPG-based models.
\end{abstract}


\keywords{Foundation models, Wearables, Health-wellbeing, Contrastive Learning  }

\begin{CCSXML}
<ccs2012>
   <concept>
       <concept_id>10003120.10003138</concept_id>
       <concept_desc>Human-centered computing~Ubiquitous and mobile computing</concept_desc>
       <concept_significance>500</concept_significance>
       </concept>
   <concept>
       <concept_id>10010147.10010257</concept_id>
       <concept_desc>Computing methodologies~Machine learning</concept_desc>
       <concept_significance>500</concept_significance>
       </concept>
 </ccs2012>
\end{CCSXML}

\ccsdesc[500]{Human-centered computing~Ubiquitous and mobile computing}
\ccsdesc[500]{Computing methodologies~Machine learning}


\maketitle

\section{Introduction} \label{sec:intro}
Unobtrusive, ubiquitous, and cost-effective wearable sensors have demonstrated the potential to revolutionize real-time monitoring of health and wellness by enabling the detection of various physical and mental health states. Photoplethysmography (PPG) in smartwatches has emerged as a widely used modality due to its non-invasive assessment of physiology without the need for firm attachment. It is used for estimating physiological metrics such as heart rate, heart rate variability~\cite{sarhaddi2022comprehensive}, blood glucose~\cite{ali2024comparison}, oxygen saturation~\cite{rajakariar2024accuracy}, and blood pressure~\cite{he2022new,fortino2010ppg}. For diagnosis, it can detect cardiovascular conditions~\cite{ouyang2017self}, including atrial fibrillation~\cite{poh2018diagnostic} and detect hypoxia~\cite{lazazzera2020detection}. For mental health and wellness, it can track stress~\cite{zhu2023stress}, emotion~\cite{kontaxis2020photoplethysmographic}, focus~\cite{wang2024classifying}, and depression~\cite{kontaxis2020photoplethysmographic}. 

However, PPG-based inference in real-world settings remains challenging due to its high susceptibility to noise from motion artifacts~\cite{pollreisz2022detection}, ambient light~\citep{cubas2023design}, and skin conditions~\cite{ajmal2021monte}. This has slowed the progress in realizing the full potential of PPG in the natural environment. For example, high accuracies for stress classification has been reported on lab data~\cite{choi2022attention,hasanpoor2022stress,motaman2022stress,alshareef2022transformer,mitro2023ai}, but they do not generalize to the field settings~\cite{mitro2023ai}. Training models on lab data from the same participants and then applying it to their field data can lead to a better performance~\cite{toshnazarov2024sosw}, but it doesn't scale to unseen participants. Models trained using larger field datasets report low performance~\cite{zhang2024reproducible,aqajari2024enhancing}.

\begin{figure}
    \centering
    \includegraphics[width=0.95\linewidth]{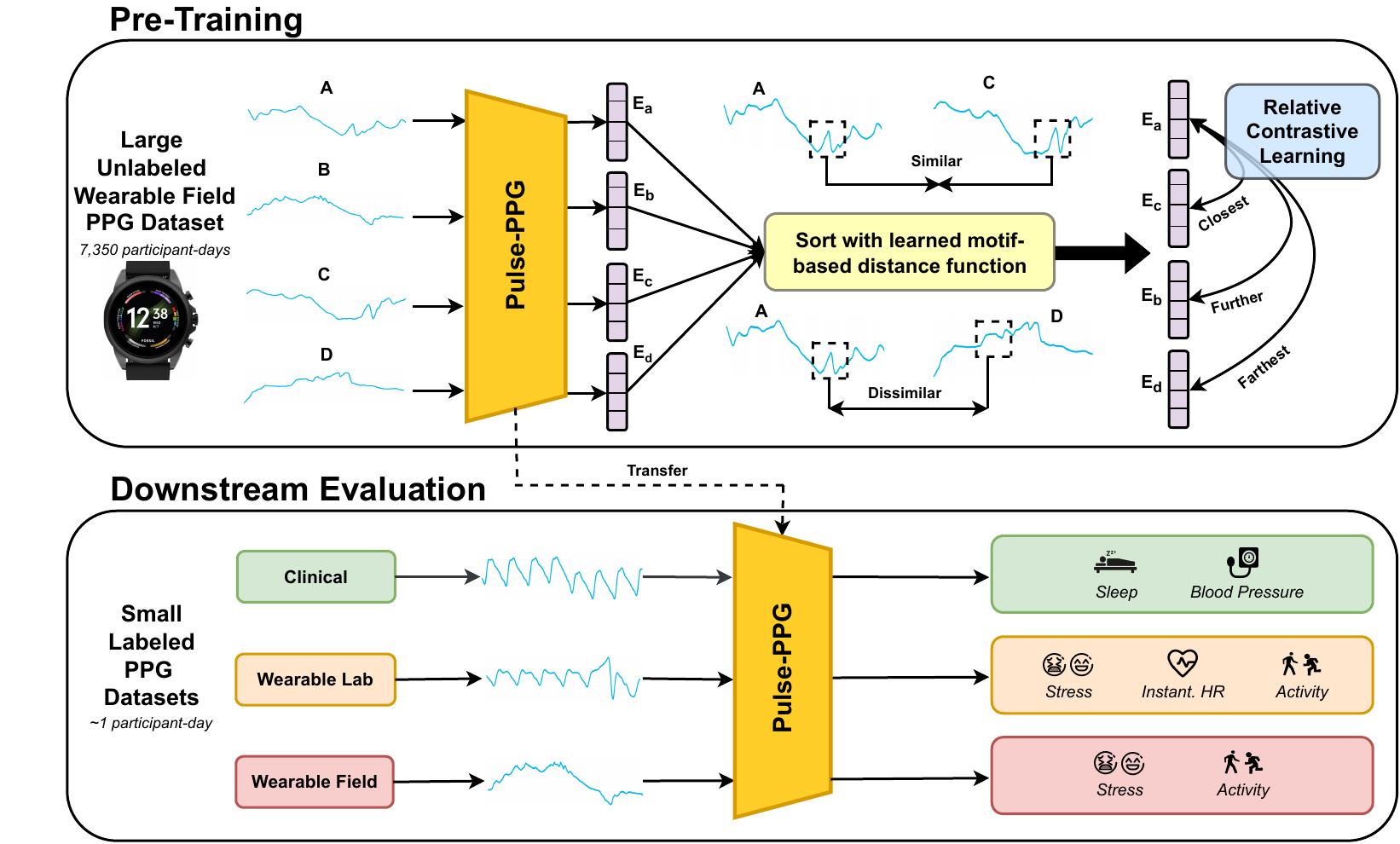}
    \caption{\textbf{Overview of our open-sourced\protect\footnotemark{} Pulse-PPG Foundation Model.} Our model is trained on wearable field PPG with relative contrastive learning, based on the relative distances captured in our learned motif-based distance function. This model demonstrates strong performance on a wide variety of tasks across wearable field, wearable lab, and clinical settings. 
    % (Figure Not finalized yet, still in progress)
    }
    \label{fig:overview}
\end{figure}
\footnotetext{\edit{\small \label{open}Pulse-PPG model weights and code repository are available at \color{black}\url{https://github.com/maxxu05/pulseppg}.}}

The emergence of Foundation Models (FM) offers a new opportunity to address these challenges and accelerate our progress. In other domains such as natural language processing and computer vision, the foundation model paradigm has transformed the development of machine learning solutions to real-world problems~\cite{AppleIntelligence, MetaDINOv2}. The key property of an FM is that it is pre-trained on a large-scale dataset to ensure that the resulting feature representation encompasses all of the complexity of the data domain, which is then validated by demonstrating that the FM can solve multiple downstream tasks without additional representation learning. In computer vision, the field has moved away from collecting individual special-purpose datasets and training task-specific models to leveraging existing foundation model representations, such as DINOv2~\cite{oquab2023dinov2} in solving a variety of perceptual tasks. A key enabling property is that while the training datasets are private and not publicly available, the model weights are released to the research community, enabling everyone to benefit from its powerful representation. This transition in utilizing publicly available FMs has not yet occurred for mHealth and is a crucial next step.

While there have been some exciting recent efforts to develop FMs for PPG~\cite{abbaspourazad2023large,pillai2024papagei}, current approaches suffer from two important limitations: 1) Private models which define the SOTA but whose weights are not available to the research commmunity~\cite{abbaspourazad2023large}, and 2) Models which demonstrate impressive performance on lab-collected data but have not been developed or evaluated for use in the field environment~\cite{pillai2024papagei}. The growing availability of field datasets has created tremendous opportunities for advancing PPG-based research. However, many of these datasets~\cite{truslow2024understanding,van2019protocol,schmidt2019multi} remain inaccessible or the computational resources required to fully exploit their potential are prohibitively expensive for many researchers and practitioners. 

In this paper, we present an open-source\footref{open} foundation model for wearable field PPG signals, \emph{Pulse-PPG}, which can be seen in Figure~\ref{fig:overview}. We pre-train our model on the raw PPG signals from a large-scale wearable field dataset, composed of approximately 200 million seconds of raw PPG data collected from 120 participants. %organized in 822,247 unique 4-minute PPG segments. 
We avoid filtering our signals because such real-world noise patterns offer meaningful contextual cues on a given individual's motion or environment, which we show enables better performance on wearable field, lab, and clinical downstream tasks, compared to training on cleaner clinical data. To address the unique challenges of field-collected PPG data, we adapted a motif-based relative contrastive learning framework \citep{xurebar, xu2024relcon} to flexibly handle the temporal and noise characteristics of the field PPG signals and extract subtle but semantically meaningful patterns. Our empirical evaluation demonstrates that this approach achieves state-of-the-art performance, outperforming a recent PPG foundation model on 10/11 of our downstream evaluation tasks across five different datasets, covering wearable data from the field, wearable data from the lab, and clinical PPG domains.

With the weights publicly available, researchers can build upon our work for their own specialized tasks and bypass the costly and time-consuming process of training from scratch, significantly lowering the barrier to entry for those without access to extensive computational resources. Moreover, our model can serve as an embedding function, enabling researchers to extract meaningful physiological representations from raw PPG data. Beyond these, the availability of our model weights will encourage reproducibility and benchmarking across datasets and settings. Researchers can compare their methods against a common baseline, accelerating research progress.
\\ 
[.15cm]
The key contributions of this work are as follows:
\begin{enumerate}[leftmargin=*]%,noitemsep]
    \item We present Pulse-PPG, the best-performing open-source\footref{open} PPG foundation model trained entirely on raw, uncurated PPG data collected from a 100-day field study involving 120 subjects from diverse backgrounds. %\edit{Pulse-PPG model weights and code repository are available at {\url{https://github.com/maxxu05/pulseppg}}.}
    \item We show that training a motif-based representation learning framework with relative contrastive learning outperforms the current state-of-the-art open-source PPG foundation model \citep{pillai2024papagei} across a wide set of downstream datasets and tasks for wearable field and clinical applications. We also show that our Pulse-PPG model with 28.5 million parameters outperforms substantially more powerful general-purpose time-series models---Chronos with 200 million parameters \citep{ansari2024chronos} and MOMENT with 385 million parameters \citep{goswami2024moment}.
    
    \item We show that training on field data achieves better performance for two different foundation model approaches than a lab-trained model on wearable lab downstream tasks, highlighting the field-to-lab generalizability of the models, with additional potential implications for clinical settings.   

\end{enumerate} 

\section{Related Works}
\subsection{Foundation Models for PPG signals}

There have been some recent work on PPG-specific foundation models, but none of them are open and designed for wearable PPG signals (i.e., collected from a smartwatch). The closest work with open model availability is PaPaGei, an open-source PPG foundation model that improves classification and regression performance across 14 tasks~\cite{pillai2024papagei}, including some wearable lab tasks. However, it is exclusively trained on clean clinical PPG signals, and was not evaluated on field data~\cite{pillai2024papagei}. We show (Section \ref{sec:exppapagei}) that training on this clinical data negatively impacted its generalization performance on wearable tasks involving field-collected data. The most relevant work for wearable PPGs is a foundation model pre-trained on a large-scale wearable field PPG dataset and presented in~\cite{abbaspourazad2023large}. However, the model is closed-source and its training datasets are private, restricting accessibility for the research community. Similarly, SiamQuality is a foundation model pre-trained on clinical data that demonstrates generalizability on wearable lab and clinical datasets, but the model remains private \cite{ding2024siamquality}. 

Additional related works addressed tasks and goals that are adjacent to this paper. REGLE employs autoencoders to extract disentangled embeddings from clinical PPGs, focusing on their applicability to genetic and disease prediction tasks \cite{yun2024unsupervised}. TS2TC introduces a generative self-supervised learning framework for clinical PPG data but does not provide pre-trained weights and has been evaluated exclusively on clinical PPG regression tasks \cite{zhang2024general}. PPG-PT demonstrates the utility of pre-training, but does not meet all of the requirements of a foundation model, as the model was only evaluated on one clinical PPG-specific downstream task \cite{davies2024interpretable}.
 
While these prior works have made significant strides in PPG representation learning, they either lack open-source accessibility, generalizability across tasks, or applicability to field-collected wearable data, highlighting a critical gap that our work aims to address. With the release of our foundation model's weights, ours will be the first open-source\footref{open} PPG model trained exclusively on field-collected data.

\subsection{Foundation Models in General}
Our work is inspired by prior successes in developing foundation models for other data modalities. These works have shown that models pre-trained in large-scale datasets can capture underlying data representations and are highly adaptable for various downstream applications \cite{bommasani2021oppoandriskforfm}. These models have had a substantial impact in fields like Natural Language Processing and Computer Vision. In the language domain, models such as BERT \citep{devlin2018bert} and the Generative Pretrained Transformer (GPT) \citep{radford2018improving} have demonstrated remarkable generalizability across a diverse range of natural language processing tasks, catalyzing the development of large language models (i.e., ChatGPT \cite{zhou2024comprehensive}). The vision domain has seen similar breakthroughs with models like CLIP \citep{radford2021learning} and the Segment Anything Model~\cite{kirillov2023segment} that are trained on large-scale data and exhibit strong performance on a broad set of downstream tasks. These works have demonstrated that pre-training on large-scale data can yield effective feature representations and serve as the motivation for this work.

\section{Pulse-PPG Methodology}\label{sec:methods}
Our goal is to develop a foundation model capable of learning a generalizable embedding space that can be effectively applied to a diverse set of datasets (ranging from clinical PPG to mHealth field PPG) and downstream tasks (ranging from blood pressure prediction to stress detection). In Section \ref{sec:whyfield}, we detail our rationale for pre-training the foundation model on noisy field PPG data, as opposed to the traditional approach of pre-training on clean, lab-study-based PPG data. Then, in Section \ref{sec:designpretrain}, we outline the key challenges involved in learning a robust foundation model in this setting and describe how our methodology addresses each of these challenges.

\subsection{Motivation for Pre-training on Field PPG data} \label{sec:whyfield}
Traditionally, prior work has focused on training models using clean, lab-collected PPG data and evaluating them on mHealth datasets~\cite{pillai2024papagei}. This approach is largely due to the significant challenges associated with utilizing real-world field data, which is inherently noisy and variable. However, we hypothesize that by designing a methodology capable of effectively learning patterns from field data, we can develop more robust representations, generalize better, and achieve superior performance on downstream mHealth tasks, such as instantaneous heart rate prediction or activity recognition. In this way, the noise present from the wearable field sensor becomes a benefit, as it could introduce additional contextual information of a person's current state and environment. Importantly, pre-training on mHealth data and evaluating on mHealth data reduces domain shift, as the model is exposed to the same types of noise, variability, and environmental factors during both training and evaluation. 

\begin{figure}
    \centering
        \vspace{-2mm}
    \includegraphics[width=0.9\linewidth]{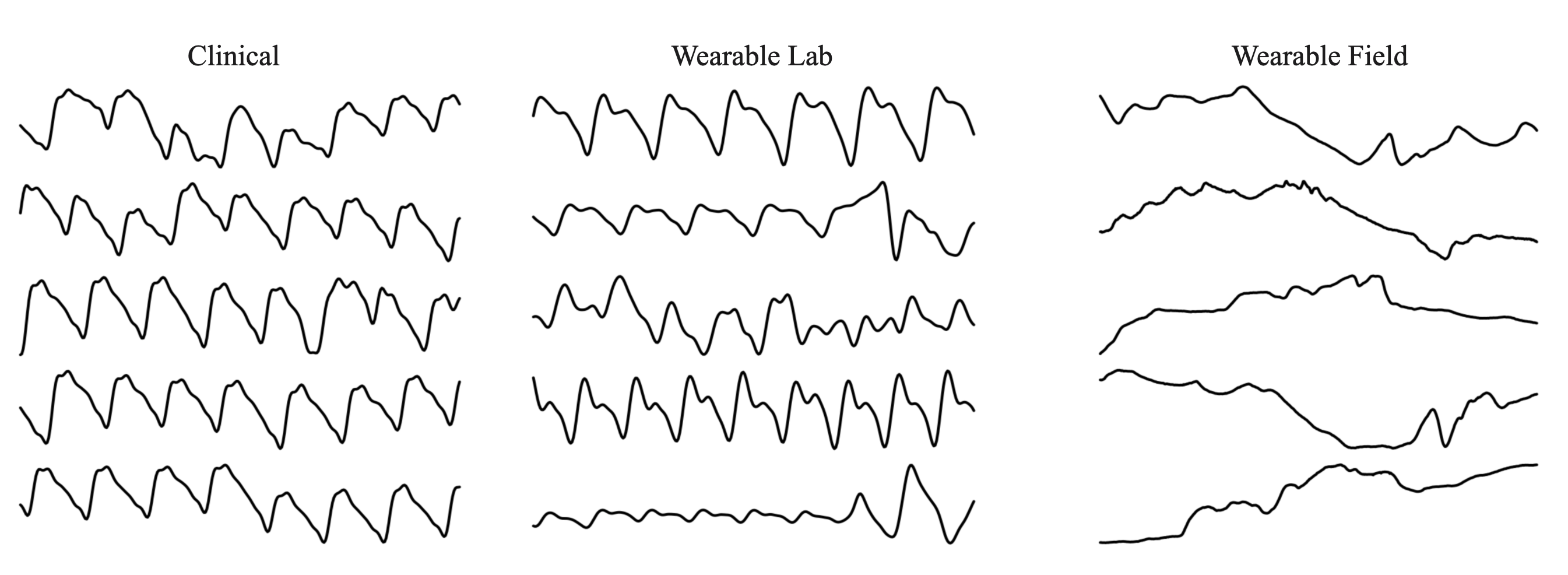}
    \vspace{-.3cm}
    \caption{\textbf{Random 5s Snippets of PPG signals from Clinical, Wearable Lab, and Wearable Field Settings}. Signal quality declines gradually but manageably when moving from clinical-grade PPG in hospital settings to wearable PPG in controlled lab conditions. However, a marked deterioration occurs in the uncontrolled real-world environment of the wearable field setting, driven by daily wear factors, such as motion artifacts~\cite{pollreisz2022detection}, ambient light \citep{cubas2023design}, and poor sensor contact \citep{lo2024advanced}.
    }    
    \label{fig:fieldhard}     \vspace{-2mm}
\end{figure}

Clinical datasets, while less noisy and more consistent due to their controlled collection processes administered by professionals, lack the dynamic variability and environmental factors present in real-world settings. For example, clinical datasets often involve participants with specific health conditions and are collected under static, controlled conditions, which fail to capture the wide range of physiological and environmental variability encountered in the wild. In contrast, field data includes participants with and without specific conditions, as well as a diverse array of physiological patterns influenced by real-world factors, such as poor sensor contact from loose bands and daily motion. This richness and diversity make field data a more comprehensive source for pre-training a foundation model. 

By pre-training on field data, we aim to create a foundation model that better reflects the complexity of real-world scenarios and generalizes more effectively across diverse applications.

\subsection{Learning Robust PPG Representations from Field Data} \label{sec:designpretrain}
To build a robust foundation model, we must design a pre-training task that not only captures high-level, obvious semantic features, such as heart rate changes, but also uncovers subtle, latent patterns in the data, such as variations in pulse waveform morphology or transient physiological responses. PPG signals, which measure blood volume changes, encode a wealth of physiological information. However, this information is often obscured by noise, particularly in the field. Traditional approaches address noise by normalizing or aggregating data into statistical features and training models on these features. While this reduces the impact of noise, it often discards fine-grained details that are critical for accurate inference or discovering hidden clusters and trends in the data.

To overcome these challenges, we introduce Pulse-PPG, a method specifically designed to address three key issues: 1) the development of a domain-specific pre-training task that leverages PPG-specific domain knowledge; 2) exploiting the inherent noise of real-world PPG data; and 3) the ability to capture subtle, yet meaningful, patterns in the data that are essential for ensuring generalizability across diverse and dynamic real-world settings.

\subsubsection{PPG-Specific Pre-training Task} \hfill\\
A classic self-supervised objective for time-series foundation models is contrastive learning, where augmentations (e.g., cropping, scaling, jittering, shuffling) create distinct but semantically similar positive pairs \citep{meng2023unsupervised, yue2022ts2vec, zhang2022tfc, eldele2023tstcc, woo2022cost, yang2022btsf, yang2022timeclr, ozyurt2022cluda, lee2022SSLAPP}. However, PPG signals face unique challenges: (1) time-series lack image-like invariances (e.g., rotation), and (2) augmentations can distort PPG semantics. For example, scaling alters amplitude (critical for blood volume inference) and shuffling disrupts temporal dependencies (key for heart rate variability). Alternatively, subject-aware contrastive learning avoids augmentations by leveraging long-term recordings from individual subjects~\cite{abbaspourazad2023large}. While especially effective for biosignals \citep{jeong2024finding}, existing methods rely on naive temporal sampling \citep{tonekaboni2021tnc} or supplemental context (e.g. event labels) \citep{jeong2023event}. To address this, we propose directly exploiting raw PPG semantics in our large-scale unlabeled dataset, eliminating reliance on augmentations or auxiliary data. To unlock the potential of our large-scale, unlabeled wearable field PPG dataset, we need a methodology that utilizes semantic information from only the raw PPG signals. 

A critical characteristic of raw PPG signals lies in their origins in the cardiopulmonary system. PPG signals are pulsative: quasiperiodic and composed of repeating motifs that correspond to individual cardiac cycles~\cite{allen2007photoplethysmography, elgendi2012analysis}. We define motifs abstractly as the short temporal shapes within the overall PPG morphology that reflect specific semantic information, such as the first upward slope in a specific PPG beat that represents systolic rise time. These motifs encode critical physiological information, such as heart rate variability, blood pressure changes, and vascular dynamics. 

Therefore, we propose to utilize our pre-training task to learn differences between PPG instances, based upon the differences in their motifs, drawing from~\citet{xu2024rebar}. In this way, the distance function, $d(\textbf{X}_{\text{anchor}}, \textbf{X}_{\text{cand}})$ can compare two PPG sequences, $\textbf{X}_{\text{anchor}} \in \mathbb{R}^{T}$ and $\textbf{X}_{\text{cand}} \in \mathbb{R}^{T}$ by finding the motif in $\textbf{X}_{\text{cand}}$ that best matches a given motif $\textbf{X}_{\text{anchor}}$. This idea can be seen below in Equation \ref{eq:distance}, with the "matching" function given by a distance function $d_m$.
\begin{align}
    d(\textbf{X}_{\text{anchor}}, \textbf{X}_{\text{cand}}) \approx \sum_{i=0}^T \underset{j \in [0,\cdots, T]} {\text{argmin }} \Big\{ d_m\Big(\textrm{Motif}\big(\textbf{X}_{\text{anchor}}[i]\big), \textrm{Motif}\big(\textbf{X}_{\text{cand}}[j]\big)\Big)\Big\} \label{eq:distance}
\end{align}

This distance function would enable semantically aware comparisons between PPG sequences, learning an embedding space that captures physiologically meaningful relationships.

\subsubsection{Exploiting the Noise of Real-World, mHealth PPG data} \label{sec:robustness} \hfill\\

Many prior works that utilize morphology information from raw PPG signals for their task-specific application, such as for blood pressure prediction \citep{salah2022beat, shin2017feasibility} or stress detection \citep{ahmed2022ppg, rinkevivcius2019photoplethysmogram}, will first utilize a peak detection algorithm to segment individual PPG beats that correspond to as single cardiac cycle  \citep{haddad2020beat, kavsaouglu2016innovative, fischer2016algorithm}. After segmentation, specific temporal features within the beat can be investigated, such as the dicrotic notch or the systolic rise time. However, these methods are evaluated exclusively on clinical or lab-collected PPG data, where signal quality is controlled and high. 

Exploiting PPG morphological information is significantly more challenging in real-world wearable field PPG data, where persistent noise and motion artifacts make beat-to-beat segmentation and morphological feature extraction infeasible. As shown in Figure \ref{fig:fieldhard}, the sharp decline in PPG signal quality, ubiquitous in mobile health settings, severely undermines the reliability of traditional segmentation and feature isolation methods.

To address this challenge, we utilize a learnable motif-based distance function (Equation \ref{eq:rebardistance}) trained directly on noisy wearable field PPG data. Unlike traditional approaches that depend on predefined morphological features or clean beat segmentation, our method learns to identify motifs from raw, uncurated signals. By training under real-world noise conditions, we leverage the inherent diversity of field data motifs. \textit{Crucially, noise patterns are not discarded but instead leveraged as meaningful contextual cues: non-random artifacts (e.g., from motion or environment) correlate with specific activities, states, or conditions, enhancing motif uniqueness.} This enables the model to identify semantic clusters and learn robust representations generalizable to diverse real-world scenarios.

\begin{align}
    d(\textbf{X}_{\textrm{anc}},\textbf{X}_{\textrm{cand}}) &\coloneqq \sum_{\textbf{x}_\textrm{anc} \in_s {\textbf{X}_\textrm{anc}}} \Bigg( \bigg(
    \sum_{\textbf{x}_\textrm{cand}  \in_s {\textbf{X}_\textrm{cand}}}  \underset{{\textbf{x}_\textrm{cand} \in_s {\textbf{X}_\textrm{cand}}}}{\textrm{softmax}} \Big(  \textrm{sim}\big( f_q( {\textbf{x}}_{\textrm{anc}} ), f_k( \textbf{x}_{\textrm{cand}}) \big) \Big) f_v( \textbf{x}_{\textrm{cand}}) \bigg) -\textbf{x}_{\textrm{anc}} \Bigg)^2 \label{eq:rebardistance} \\
    f_{\{q/k/v\}}( {\textbf{X}}) &= \textrm{DilatedConvNet}_{\{q/k/v\}} \big( {{\textbf{X}} } \big) \label{eq:rebardilconv} 
\end{align}
where $\smash{\textbf{X} \in \mathbb{R}^{T \times D}}$ and $\smash{\textbf{x}\in \mathbb{R}^{D}}$ with $T$ as the time length, $\in_S$ as a $\in$ but with a subsampling of stride $s$, and $sim()$ is a dot product. This function is implemented with a standard cross-attention block \citep{vaswani2017attnisallyouneedtransformer} with projection and normalization layers being omitted for brevity. 
\\

This trainable function closely matches our idea presented in Equation \ref{eq:distance}. $d_m$ is modeled as a reconstruction error, where the motifs of $\textbf{X}_{\text{cand}}$ that best matches a given motif $\textbf{x}_{\text{anchor}}$ are used to reconstruct $\textbf{x}_{\text{anchor}}$, in a kernel regression. Motifs are captured with the dilated convolutions of $f_{\{q/k/v\}}$, as seen in Equation \ref{eq:rebardilconv}. The weights in the kernel regression found in $\smash{(\sum_{{\textbf{X}_\textrm{cand}}}  {\textrm{softmax}})}$ serve the purpose of the original argmin, where the function is learned to find the weights that minimize the reconstruction error. 

This work builds on \citet{xurebar}, which proposed a proof-of-concept motif-based distance for physiological signals in controlled, small-scale settings. However, their approach was limited to pre-training and evaluation on the same small labeled dataset, lacking the scalability and generalizability required for real-world foundation models. For example, the original work only evaluated on sequences with a maximum length of 3,840 time-points, much shorter than our 4-minute-long PPG input lengths of 12,000. This is significant because the backbone of the training function utilizes a cross-attention function, with a computational complexity of $O(T^2)$, but we were able to address this with our $\in_s$ function with $s=10$. This allows for the reconstruction error to be calculated from subsample of the original sequence, effectively reducing computational complexity to  $O((T/s)^2)$, but without degrading the quality of motif similarity comparisons by allowing $f_{\{q/k/v\}}$ to capture the motifs found from the full sequence. Additionally, the prior work \citep{xurebar} utilized a simple, untuned dilated convolution encoder to act as a control to fairly compare self-supervised learning methods. In our work, we utilize a much more complex ResNet encoder backbone to develop a powerful foundation model that can capture and handle the noise and diversity in real-world PPG data. To this end, it necessitated an extensive tuning of the methodology (see Section~\ref{sec:model_adaptation}) to work in our setting, but due to this, we have been able to demonstrate robustness in the dynamic field environment across diverse downstream tasks and datasets.

\subsubsection{Capturing Subtle Yet Meaningful Patterns} \label{sec:meaningful-patterns} \hfill\\

Given our motif-based distance function, we can identify positive and negative pairs to draw together and push apart in the embedding space, respectively. Traditionally, contrastive approaches will utilize the Normalized-Temperature Cross-Entropy Loss (NT-Xent) function (Equation \ref{eq:ntxent}), in which the positive pair is pulled towards the anchor, and all negatives are pushed away equally. 
\begin{align}
     \ell(\textbf{X}_{\textrm{anc}}, \textbf{X}_{\textrm{pos}}, \mathcal{S}_{\textrm{neg}})  = - \log \frac{\exp(\textrm{sim}(\textbf{X}_{\textrm{anc}},  \textbf{X}_{\textrm{pos}}) / \tau)}{\sum_{\textbf{X}_{\textrm{neg}} \in \mathcal{S}_{\textrm{neg}}} \exp(\textrm{sim}(\textbf{X}_{\textrm{anc}},  \textbf{X}_{\textrm{neg}}) / \tau) + \exp(\textrm{sim}(\textbf{X}_{\textrm{anc}},  \textbf{X}_{\textrm{pos}}) / \tau)} \label{eq:ntxent}
\end{align}
The standard NT-Xent approach’s reliance on rigid positive/negative assignments makes it prone to false positives (misclassifying dissimilar instances as positives) and false negatives (overlooking valid positives in the candidate pool). For example, in stress assessment, being stressed due to bad traffic differs from the experience of being tailgated, but the two are more similar compared to being engaged in a verbal conflict. Treating all negatives equally ignores such nuances, resulting in overly coarse clusters that fail to capture subtle semantic relationships. 

An alternative idea to address this is utilizing a metric learning loss function \citep{kim2019metricbeyondlogratio, hoffmann2022metricrince, kan2021metricroc}, in which the encoder learns to embed the PPG instances exactly as described by our motif-based distance function.  However, although our learned motif-based distance function will be able to capture semantic information, it was trained without labels and is not expected to always perfectly identify the correct semantic relationships.

A "goldilocks" approach between these two ideas would be using a relative contrastive learning framework that models the relative positions of the distances. As such, we utilize the concept of relative dissimilarity via multiple negative sets \citep{xu2024relcon}. For each positive pairing, a negative set is created by selecting samples whose distance from the anchor exceeds the distance of the positive pair. Each candidate in the dataset is iteratively used to form a positive pair, while the remaining candidates contribute to the negative set:
\[
f_{\textrm{neg}}({X}_{\textrm{anc}}, {X}_{\textrm{pos}}, \mathcal{S}) = \{{X} \in \mathcal{S} \mid d({X}_{\textrm{anc}}, {X}) > d({X}_{\textrm{anc}}, {X}_{\textrm{pos}}) \}
\]
This concept is incorporated into the Relative Contrastive Loss function, which builds representation spaces that are aware of relative distances and semantic nuances:
\begin{align}
    \mathcal{L}_{\textrm{RelCon}} = \sum_{{X}_{\textrm{i}} \in \mathcal{S}_{\textrm{cand}}} \ell({X}_{\textrm{anc}},\ {X}_{\textrm{pos}} \coloneqq {X}_{\textrm{i}},\ \mathcal{S}_{\textrm{neg}} \coloneqq f_{\textrm{neg}}({X}_{\textrm{anc}}, {X}_{\textrm{i}}, \mathcal{S}_{\textrm{cand}})) \label{eq:relconloss}
\end{align}
By integrating relative dissimilarity into contrastive learning, the model forms tight clusters in a representation space reflecting fine-grained relationships. This improves generalization to unseen users by capturing within- and between-class nuances, ensuring robustness to intra-user variability and adaptability to new users’ physiological patterns. The resulting representations reliably handle field data variability for users unseen during training.

This work builds on \citet{xu2024relcon}, which proposed relative contrastive learning to learn a foundation model for accelerometry sensor data. However, accelerometry sensor data varies significantly from PPG sensor data. For example, accelerometry has a well-established set of invariances (i.e., 3D rotation) \citep{tang2020exploring}, due to how inertial its measurements are fundamentally governed by physics—the magnitude and temporal patterns of acceleration reflect biomechanical forces generated by movement, regardless of sensor orientation or placement. PPG signals, by contrast, exhibit no such geometric invariances—their morphology is highly sensitive to sensor contact, motion artifacts, and physiological variability. Rather than presupposing geometric symmetries, for Pulse-PPG, we allow the motif-based distance function to implicitly learn noise-robust representations directly from the raw PPG waveform’s temporal structure. Additionally, their work, as well as other accelerometry machine learning work, demonstrates that short time windows can capture semantically meaningful motion information, and as such, they only encode accelerometry segments spanning 2.56 seconds, or 256 time points. However, for Pulse-PPG, we seek to model 4-minute (12,000 time point) PPG segments, to align with prior stress models \citep{lamichhane2017towards,wilson2018couples,bari2020automated}. Therefore, this also results in an extensive tuning of the methodology (see Section~\ref{sec:model_adaptation}) to work in our setting. For example, we reduce the same-subject candidates from 20 to 1 and constrain them to be drawn from the same hour as the anchor. Due to this tuning, we have been able to develop Pulse-PPG, a powerful foundation model that generalizes across many different PPG settings and tasks.

\section{Pre-training Procedure for Pulse-PPG} 

In this section, we provide the technical details of our foundation model's implementation and pre-training. Section \ref{sec:modelpretraining} outlines how we employ our learnable motif-based distance function and relative contrastive loss to train the Pulse-PPG foundation model. Section \ref{sec:pretraining_data} describes the composition and characteristics of our pre-training dataset, detailing how we process the wearable field PPG data to align with our pre-training task, as well as highlighting the dataset's inherent properties that make it well-suited for learning robust representations in noisy, real-world conditions. Finally, Section \ref{sec:implementation} describes specific implementation details, such as specific architecture and total GPU compute used. Please refer to Appendix \label{sec:hyperparams} for specific hyperparameter details. 
% or our code repository at \url{https://github.com/maxxu05/Pulse-PPG}.

\subsection{Model Pre-training Details} \label{sec:modelpretraining}

\subsubsection{Training the Learnable Motif-based Distance Function} \label{sec:trainrebar} As mentioned previously in Section \ref{sec:robustness}, making our distance function learnable is a key strength of our approach, as it enables motif-comparisons between PPG sequences, even when such PPGs are afflicted with noise. Therefore, before utilizing our relative contrastive loss, we learn our distance function in an unsupervised fashion.

In our initial label-free pre-training setting, we do not know if a random pair of PPG sequences shares class labels, and thus, we do not know if reconstruction error should be minimized to learn a motif-matching similarity function between them. Therefore, during training, we set ${\textbf{X}}_q$ and $ {\textbf{X}}_k$ to be the same value and apply a missingness mask on ${\textbf{X}}_q$. This mask has a contiguous segment of missingness that stretches 2 seconds. In this way, our distance function will learn to retrieve the regions from the key $\textbf{X}_k$ that match the missing 2-second motif from the query, $\bar{\textbf{X}}_q$, for reconstruction. Then, after training, we can utilize this distance as a static function for the RelCon loss to identify relative distances between ${\textbf{X}}_q $ and $ {\textbf{X}}_k$, where ${\textbf{X}}_q $ and $ {\textbf{X}}_k$ are different instances with different values.

\subsubsection{Training the Pulse-PPG Foundation Model} Now, once we have our trained motif-based distance function, we can use it to identify relative distances of a candidate set of PPG sequences from an anchor PPG sequence.  This candidate set of PPG sequences is sampled from two sources: within-subject and between-subject. 
\begin{itemize}[leftmargin=.7cm,  noitemsep]
    \item Sampling within-subject, across-time, draws candidate PPG sequences from the same subject, but at a different time window from the anchor. This allows us to model how an individual's behavioral patterns throughout their day affect their PPG signal, for example, via exercise. 
    \item Sampling between-subject draws candidate PPG sequences from other persons within a batch. This allows us to model similarities and differences \edit{in physiology among different individuals.}  
\end{itemize}

Given this candidate set, we can sort the candidates with our motif-based distance function found in Equation \ref{eq:rebardistance} and then apply our RelCon loss function in Equation \ref{eq:relconloss} based on their relative distances from the anchor. As such, our Pulse-PPG foundation model will be able to capture both within-subject and between-subject semantic information within the embedding space.

\subsection{Pre-training on a Field Wearable PPG Dataset} \label{sec:pretraining_data}
\subsubsection{Dataset Description:} For our pre-training, we obtained access to the Mobile Open Observation of Daily Stressors (MOODS) study dataset~\cite{neupane2024momentarymoods} that is composed of 822,247 unique 4-minute 50 Hz PPG segments, with 122 participants who wore a smartwatch every day for up to 100 days. Participants included 39 men, 77 women, and 6 non-binary individuals, with an average age of 38 years (\( SD = 13 \)). Seventy-one were working professionals across fields such as academia, information technology, administration, public safety, health, law, sales, and construction; fifty were students pursuing higher education, and one participant did not disclose their occupation. Participants wore a study-provided Fossil Sport (Version 4) smartwatch, and approximately 55K hours of raw PPG data, sampled at 100 HZ, were collected during the study. The study was conducted using a WearOS app for smartwatches, a cross-platform smartphone app, and cloud services. Detailed descriptions of the system and study design of the MOODS study are available in~\cite{neupane2024momentarymoods}.

\subsubsection{Subject-wise Train/Val/Test Splits:}\label{sec:train_val_test} To pre-train a model capable of handling substantial within- and between-subject variability, we randomly shuffled participants and created a train/val/test split of 84/18/18. During pre-training, the test set is never used, however, we establish this split so that the downstream \edit{tasks} that utilize this same dataset (i.e., Wearable Field Stress and Activity) have a held-out test set \edit{never exposed during pre-training}. We deliberately adopted a subject-based splitting strategy to pose a more challenging task for the model: making inferences on data from entirely unseen individuals, a common scenario in real-world applications.

\subsubsection{Preserving Noisiness:} To pre-train the physiological model, we utilize raw photoplethysmography (PPG) data collected from the participants. Out of 122 participants, the dataset recorded PPG data from 120 participants, which we use for pre-training. PPG data collected in the field is noisy; however, avoiding signal-specific filtering can provide certain advantages for generalized, robust representations, e.g., adapting to dataset diversity, preserving subtle meaningful variations, and better transferability. Hence, we refrain from applying any particular filtering technique, but focus on addressing between-person variability through a global person-specific $z$-normalization. \edit{Please see Appendix~\ref{sec:ablation} for our ablation study, which shows that normalization consistently enhances performance.}

\subsubsection{PPG Sequence Inputs for Pre-Training Model}
In our modeling approach, we use 4-minute long windows as sequence inputs for the Pulse-PPG Foundation Model. This choice is grounded in research highlighting the cyclical nature of physiological responses to stress. Studies have shown that stress-induced patterns in heart rate and heart rate variability (HRV) follow cyclical behaviors, with durations typically ranging from 3.5 to 4.2 minutes \citep{lamichhane2017towards,wilson2018couples,bari2020automated}. For example, heart rate increases during stress and remains elevated until the stressor ends, after which it returns to baseline, forming a cyclical pattern. Field studies involving stressful conversations or work-related stress, have further validated these patterns, with median cycle durations around 4 minutes \citep{bari2020automated}.

By segmenting PPG signals into non-overlapping 4-minute windows, we aim to capture both micro-level perturbations (e.g., transient changes in pulse waveform morphology) and macro-level cyclical patterns (e.g., sustained heart rate elevation during stress). This approach allows our model to learn representations that are sensitive to the temporal dynamics of stress, which are critical for accurate stress detection. The 4-minute window size strikes a balance between capturing meaningful cyclical patterns and retaining fine-grained temporal details. \edit{Please refer to Appendix \ref{sec:ablation} for the ablation study of evaluating the effects of varying input window sizes. Overall, the 4-minute window yields the best performance, consistently outperforming the 1- and 2-minute models across most metrics and ranking within the top two.}

It is important to note that, although our initial pre-training inputs are 4 minutes long, our model incorporates a temporal pooling mechanism that enables it to generalize to inputs of varying lengths. This flexibility allows our model to be utilized on our diverse downstream tasks, in which the input windows are variable in length. 

\subsection{Pulse-PPG Model Implementation Details}  \label{sec:implementation} 

\subsubsection{Model Adaptation and Tuning}\label{sec:model_adaptation}
As described in Sections~\ref{sec:robustness} and~\ref{sec:meaningful-patterns}, we adapt the recent motif-similarity-based reconstruction and relative contrastive loss-based self-supervised modeling approach~\cite{xu2024rebar,xu2024relcon} so that it can capture the intricacies of 4-minute segments of PPG data.

We began with a parameter search for the distance function model, which was responsible for motif-based reconstruction of masked PPG segments. In this phase, we experimented with different kernel sizes (5, 11, 15, 21, 60, 120), embedding dimensions (32, 64, 128), double receptive field values (to control how many layers and how much temporal context the network could attend to), and learning rates (0.001, 0.0001). Smaller kernel sizes and shallower receptive fields failed to capture sufficient temporal context for motif alignment, while overly large fields led to over-smoothing and poor reconstruction on noisy segments. The selected configuration provided the best trade-off between learning localized motifs and capturing broader temporal dependencies in field data. 

Once the distance model was trained, we used it to train a 1D ResNet self-supervised encoder. In our parameter exploration for this ResNet encoder, we evaluated different kernel sizes (9, 11, 15), initial base filter sizes (32, 64, 128), stride sizes (1, 2, 5, 10, 15, 20), number of ResNet blocks (4, 8, 12, 16), number of within-user same-hour candidates (1, 5, 10, 15), final pooling strategies (mean, max), and learning rates (0.001, 0.0001). We found that while deeper architectures marginally improved performance on training subsets, they generalized poorly on held-out field data, likely due to overfitting on participant-specific noise. Pooling strategies also played a significant role: max pooling consistently produced more discriminative embeddings for downstream tasks, especially when separating subtle variations in physiological activity. Once we identified the best combination that gave us reasonable performance on the 10-day dataset, we initiated the same process for the full 100-day dataset, using a more informed combination of parameters rather than a random search. In this stage, we repeated the parameter search process for both learning stages and found the best hyperparameter combinations to be:
\begin{itemize}[leftmargin=*]
    \item Distance learning: kernel size = 15, embedding dim = 64, double receptive field = 5, epochs = 20, learning rate = 0.001, batch size = 16, and stride = 10.
    
\item Contrastive learning: base filters = 128, kernel size = 11, stride = 2, groups = 1, nblocks = 12, final pool = max, epochs = 20, learning rate = 0.0001, and batch size = 64.
\end{itemize}

\subsubsection{Encoder Details:}\label{sec:resnet}
We utilize a 1D ResNet-26 network from the \url{https://github.com/hsd1503/resnet1d} code repository as our encoder backbone, with a total of 28.5m trainable parameters. There is instance normalization placed at the start of the ResNet to account for PPG distribution shifts across users and datasets. Our ResNet implementation utilizes an initial filter size of 128, a kernel size of 11, and a stride of 2, with 12 residual blocks that linearly increase the filter size every 4 blocks. There is a global average temporal pooling at the end of the network for our model to generalize across different time scales. As a result of this, our encoder embeds variable-length PPG signals into a single 1D 512-dimensional embedding vector.

\subsubsection{Distance Function Details:}
We utilize the code from the \url{https://github.com/maxxu05/rebar} code repository to implement the distance function. This is a lightweight model, used as a static function to identify relative distances for the RelCon loss function, and only has 127k trainable parameters. The dilated convolution network is a series of dilated convolution blocks, made up of a dilated convolution layer with a residual connection skip, followed by a ReLU and instance norm. There are 5 blocks, with an initial dilation of 1 that doubles after each layer. The filter size is set to 64, the kernel size is set to 15, and the stride is set to 10. The input convolution layer to the DilatedConvNet is set to be a partial convolution \citep{liu2018partialconv} to handle missingness in our pre-training and application procedures. The reconstruction error used as our distance function is a standard MSE loss.

\subsubsection{Training Details:}\label{sec:training}
Our Pulse-PPG foundation model was trained for 5 days for 6 epochs on an NVIDIA L40S GPU. Each epoch is composed of 606,833 unique 4-minute PPG segments or about 5,424 unique participant days, and due to this large data size, each epoch took 20 hours to finish. We chose this early stopping point due to convergence of the pre-training loss, (2) diminishing returns on downstream task performance, and (3) computational efficiency—extending further would yield negligible benefits while significantly increasing the carbon footprint. \edit{This can be seen in the training and validation loss curves visualized in Figure \ref{fig:losscurvel}}. \edit{Both training and validation loss are calculated for each batch, but aggregated across the entire epoch for reporting. The training loss is backpropagated and weights are updated after each step for each batch (accelerating convergence and lowering computational cost), aligning with deep learning standards \citep{amari1993backpropagation}.} An Adam optimizer was used with a learning rate set to 0.0001, betas set to (0.9, 0.999), no weight decay, and a batch size of 64.

The distance function was trained over 10 hours for 20 epochs with an Adam optimizer with a learning rate of 0.001, betas set to (0.9, 0.999), no weight decay, and a batch size of 16. Models were constructed with the PyTorch Python package, and full reproducibility details can be found in our final released codebase.

\section{\protect\edit{Evaluation Set-up}}
The power of a foundation model lies in its ability to generalize across a broad range of tasks and datasets. This makes them especially valuable to the research community, providing pre-trained, general-purpose representations that can be easily adapted to new tasks, streamlining research efforts without the need for extensive task-specific data collection or manual feature engineering. Therefore, we design our experiments to validate this by evaluating our Pulse-PPG Foundation Model on 11 different downstream tasks, across 5 datasets, spanning typical wearable PPG tasks (i.e., stress classification) and typical hospital PPG tasks (i.e., systolic blood pressure regression). \edit{In the following sections, we will describe our evaluation set-up and experimental results. We describe our downstream datasets and tasks in Section~\ref{sec:downstream}, evaluation methodologies in Section~\ref{sec:evalmethods}, evaluation metrics in Section~\ref{sec:metrics} and experimental results in Section~\ref{sec:experiment_results}.
}

\subsection{Downstream Datasets and Tasks} \label{sec:downstream}
In order to evaluate the generalizability of our Pulse-PPG foundation model, we evaluate our 5 downstream datasets and 11 tasks, which can be seen in Table \ref{tab:ppg_comparison}. Our datasets have a range of different "PPG type" settings:
\vspace{-1mm}
\begin{itemize}[leftmargin=*,  noitemsep]
    \item Wearable Field PPGs reflect the real-world setting for wearable applications with low signal quality.
    \item Wearable Lab PPGs originate from controlled lab studies, in which subjects are given a wearable PPG sensor and asked to do a set of scripted activities under supervision. Generally, signal quality is higher.
    \item Clinical PPGs originate from a hospital or another heavily monitored environment, in which PPG signals are collected from a finger sensor on stationary patients. Generally, signal quality is high.
\end{itemize}  

\begin{table}[htbp] % [!htbp]
\centering     \vspace{-1mm}
\renewcommand{\arraystretch}{1.2} % Adjust row height for better spacing
\resizebox{.95\textwidth}{!}{%
\begin{tabular}{l ccccc}
\toprule
\textbf{}                        & \textbf{MOODS} \citep{neupane2024momentarymoods} & \textbf{PPG-DaLiA} \citep{reiss2019ppgdalia} & \textbf{WESAD} \citep{schmidt2018wesad} & \textbf{SDB} \citep{garde2014development} & \textbf{PPG-BP} \citep{elgendi2022datasetppgbp}\\ 
\midrule
\textbf{PPG Type}                & Wearable        & Wearable            & Wearable        & Clinical     & Clinical         \\ 
\textbf{Quality Type}         & Field          & Clean              & Clean          & Clean        & Clean            \\ 
\textbf{Num of Subjects} & 120            & 15                 & 15             & 146          & 219              \\ 
\textbf{Time per Subject}  & 100 Days       & 103 Minutes        & 60 Minutes    & 360 Minutes           & 10 Seconds               \\ 
\midrule
\textbf{Tasks Available}         & \parbox{2.5cm}{\centering Stress (2) \\ Activity (2)}  
                                 & \parbox{2.5cm}{\centering Instant HR (R) \\ Activity (9)}  
                                 & \parbox{2.5cm}{\centering Stress (4) \\ Stress (2)}  
                                 & \parbox{2.5cm}{\centering Sleep \\ Disturbance (2)}  
                                 & \parbox{2.5cm}{\centering Systolic BP (R) \\ Diastolic BP (R) \\ Avg. HR (R) \\ Hypertension (2)} \\ 
\bottomrule
\end{tabular}%
}
\caption{\textbf{Downstream Datasets and Tasks.} $(\boldsymbol{\cdot})$ designates either a regression task or \# of classes for a classification task.}
\label{tab:ppg_comparison} \vspace{-5mm}
\end{table}

Additionally, with each of the PPG types, there is a set of canonical downstream tasks associated with them. For wearables, that includes stress classification~\cite{schmidt2018wesad,neupane2024momentarymoods}, activity classification~\cite{neupane2024momentarymoods,reiss2019ppgdalia}, and instantaneous HR regression~\cite{reiss2019ppgdalia}. For clinical settings, this includes sleep disturbance classification~\cite{garde2014development}, blood pressure regression~\cite{elgendi2022datasetppgbp}, and hypertension ~\cite{elgendi2022datasetppgbp}. While activity classification is traditionally performed using IMU sensors or motion signals, we include it as a downstream task to showcase the versatility of our model in learning meaningful representations solely from PPG data.
Each of these canonical tasks is captured in our downstream datasets. Please see Appendix \ref{sec:evaldata} for further dataset details.

\subsection{Evaluation Implementation Details} \label{sec:evalmethods}

All PPG signals were resampled to match the 50 Hz sampling frequency present in our pre-training MOODS dataset \citep{neupane2024momentarymoods}. For our evaluations using PaPaGei, we made sure to resample the signals to 125 Hz to match the sampling frequency that they trained on \citep{pillai2024papagei}. 

To train and evaluate our linear probes, we combined the train and validation splits and used a cross-validation hyperparameter grid search with the Scikit Learn package. For the regression tasks, we utilized a linear regression with an L2 regularization and conducted a grid search over the following parameters: \url{ 'alpha': [0.1, 1.0, 10.0, 100.0], 'solver': ['auto', 'cholesky', 'sparse_cg']} with the scoring function as negative MSE. For the classification tasks, we utilized a logistic regression with an L2 regularization and lbfgs solver and conducted a grid search over the following parameters \url{'C': [0.01, 0.1, 1, 10, 100], 'solver': ['lbfgs'], 'max_iter': [1000, 10_000]} with the scoring function as macro F1. Inputs into the probes were normalized with the Scikit Learn StandardScaler module. 

In order to train and evaluate our fine-tuning results, we fit a classification or regression head on top of the existing network, then train the entire network end-to-end. For classification, the head was made up of a layer normalization layer, followed by a linear layer, with the loss function being a linear combination of Cross Entropy loss, a soft F1 loss \citep{benedict2021sigmoidf1}, and Dice loss \citep{li2019dice}. For regression, the head was an MLP layer, with an initial layer with an output dimensionality of 128, followed by the GeLU activation, followed by a final linear layer. The loss function used was the MSE loss function. For both evaluations, we utilize an Adam optimizer. The checkpoint with the highest macro F1 score and the lowest MAE on the validation split was selected for inference on the test set. 
% For more details, please check our public\footref{open} codebase that will be released upon acceptance.
\edit{For exact implementation details, please refer to our public codebase: \url{https://github.com/maxxu05/pulseppg}}.

\subsection{Evaluation Metrics} \label{sec:metrics}
For evaluating our models across diverse downstream tasks, we utilize task-specific metrics for both regression and classification problems. For regression tasks, we report standard Mean Absolute Error (MAE) and Mean Squared Error (MSE). Additionally, we use Mean Absolute Percentage Error (MAPE) to assess relative errors across different scales.
For classification tasks, we report macro F1 score, Accuracy, Precision, and Recall to evaluate overall model performance and the trade-offs between false positives and false negatives. To provide a more comprehensive assessment, particularly for imbalanced datasets, we compute Area Under the Precision-Recall Curve (AUPRC) and Area Under the Receiver Operating Characteristic Curve (AUROC), which measure the model’s ability to distinguish between classes.

\section{\protect\edit{Experimental} Results} \label{sec:experiment_results}
\edit{
Our primary claims are that (1) Pulse-PPG is able to achieve state-of-the-art performance across a wide set of downstream PPG tasks for wearable and clinical applications, (2) pre-training on wearable field data achieves surprisingly strong performance compared to cleaner clinical data, and that (3) researchers will be able to utilize our open-sourced model in a variety of use-cases to expedite their own research efforts. 

We evaluate these claims through a series of experiments: Pulse-PPG vs. PaPaGei, a prior PPG FM (Section~\ref{sec:exppapagei}); Pulse-PPG vs. General Time-series FMs and Supervised Baselines (Section~\ref{sec:expbaselinegeneralfoundation}); Assessing Field-to-Lab Generalizability (Section \ref{sec:expfieldtolab}); and finally  the Evaluation of Pulse-PPG use-cases (Section~\ref{sec:finetuneeval}). }

\subsection{Experiment 1: Pulse-PPG vs. PaPaGei, A Prior PPG foundation model} \label{sec:exppapagei}
\subsubsection{Background} \label{sec:exppapagei-background}
To evaluate the effectiveness of our Pulse-PPG foundation model, we benchmark against the current state-of-the-art in open PPG foundation model, PaPaGei \cite{pillai2024papagei}. Their PaPaGei-S model uses a 1D ResNet backbone and leverages a novel self-supervised objective with PPG-morphology specific metrics, like the Inflection Point Area ratio, achieving top performance on clinically-focused downstream tasks. Crucially, they release model weights and code (\url{https://zenodo.org/records/13983110}), which we utilize for replication.

However, their pre-training dataset was composed solely of pre-processed, clean clinical PPG signals. As previously discussed in Section \ref{sec:robustness}, we hypothesize that their PPG-morphology specific metrics will be unable to generalize well in the noisier field setting. This is unlike our learned distance metric, which can potentially learn to do motif-similarity comparisons, even in noisy settings.

% We also note that while PaPaGei demonstrated strong performance across various datasets in their work, their evaluation methodology involved using a different version of the foundation model for \emph{each task}. Thus, while PaPaGei sets a critical stepping stone for PPG foundation modeling, our goal is to build a single open-source\footref{open} PPG foundation model that can generalize across many tasks, without re-training the network from scratch.

We also note that while PaPaGei demonstrated strong performance across various datasets in their work, their evaluation methodology involved using multiple different hyperparameter-tweaked versions of the foundation model and choosing the best model per task \citep{pillai2024papagei}. We report results on one set of pre-trained weights per model (for both PaPaGei and Pulse-PPG) throughout our entire paper, which explains any discrepancies in our reported numbers versus their original work. While PaPaGei sets a critical stepping stone for PPG foundation modeling, our goal is to build a \textit{single} open-source\footref{open} PPG foundation model that can generalize across many tasks.

\begin{figure}
    \centering
        \vspace{-3mm}
    \includegraphics[width=.9\linewidth]{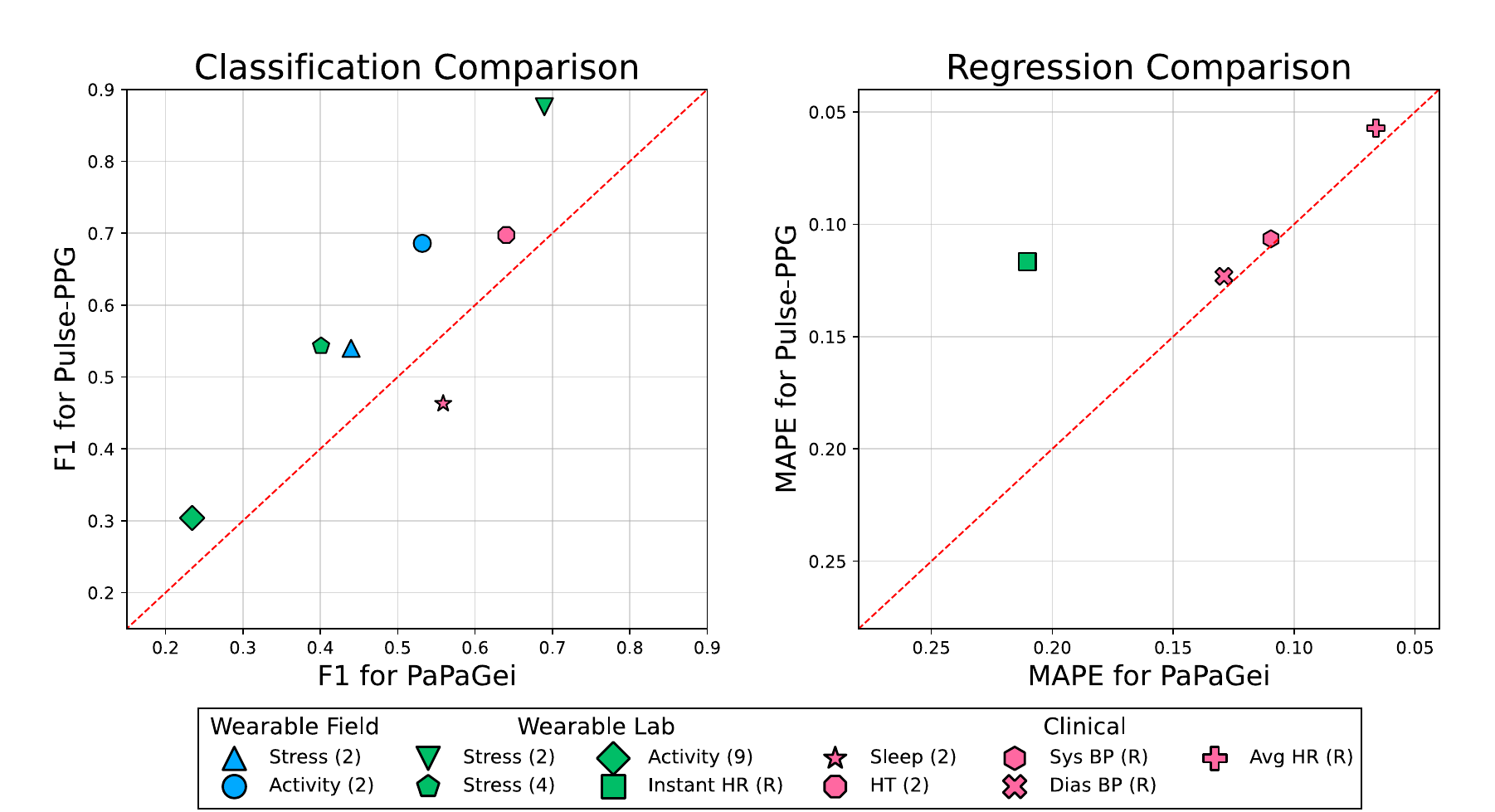}
    \Description{This figure compares the Pulse-PPG model with the PaPaGei foundation model, showcasing performance in classification (F1 score) and regression (MAPE). Pulse-PPG consistently outperforms PaPaGei, especially in Wearable Field and Lab PPG tasks.}
        \vspace{-2mm}
    \caption{\textbf{Comparison of Pulse-PPG vs. PaPaGei~\citep{pillai2024papagei}, a prior PPG foundation model}. The two plots show the relative performance of a linear probe evaluation for each model, with the left for classification via F1 score and the right for regression via Mean Average Percentage Error. 10/11 of the task data points reside above the slope, demonstrating how Pulse-PPG consistently outperforms PaPaGei, with particularly substantial improvements for the Wearable Field and Lab PPG tasks. See further discussion in Section \ref{sec:badonsdb}.    
    }     \vspace{-2mm}
    \label{fig:relconvspapagei}
\end{figure}

\subsubsection{Experimental Design}
To ensure a fair comparison, we utilize a linear probing evaluation for each downstream task for both Pulse-PPG and PaPaGei-S models. Linear probing is a widely accepted evaluation method for foundation models because it isolates the quality of each foundation model's learned representations by freezing the model's embeddings and training only a simple logistic classifier or linear regression. Both PaPaGei and Pulse-PPG embed PPG sequences into a single 1D 512-dimensional vector, making their linear probe results directly comparable. By using linear probing—training a simple classifier on frozen embeddings, we isolate the quality of the learned representations, avoiding biases from architectural differences. 

To re-emphasize, this linear probing evaluation is \emph{not intended to achieve maximum task-specific performance}, instead serving as a controlled experiment to enable a fair comparison between the foundation models. 

\subsubsection{Results} \label{sec:badonsdb} Figure \ref{fig:relconvspapagei} illustrates the difference in performance between our Pulse-PPG foundation model and the publicly released pre-trained PaPaGei. Full results can be found in Table \ref{tab:exp23} in the Appendix. Pulse-PPG outperforms PaPaGei on 10/11 of our downstream tasks, consistently across many of the reported metrics. 
% \edit{As noted earlier in Section~\ref{sec:exppapagei-background}), discrepancies with the original PaPaGei paper stem from their use of a different model version per task, which may not be indicative of true generalization performance.}

On the Wearable PPG tasks, our Pulse-PPG achieves a consistently large \edit{$\ge.\textcolor{black}{10}$} increase in F1 scores in the classification tasks, \edit{except one.} 
%(Stress - MOODS: 0.10 / Activity - MOODS: 0.15 / Stress (2) - WESAD: 0.15 / Stress (4) - WESAD: 0.14 / Activity - PPG-DaLiA: \edit{0.06}). 
Then, on the instantaneous HR regression task, our Pulse-PPG model has a large \edit{6.28} improvement in MAE. These heart rate estimates are conducted on unfiltered wearable PPG data, which is significant due to the widely recognized challenge of noise-induced degradation in heart rate estimation. On the Clinical PPG tasks, our Pulse-PPG model continues to outperform PaPaGei, with $\sim0.5$ improvement in MAE for Systolic BP, Diastolic BP, and average HR regression, as well as a .05 improvement in F1 score on hypertension classification.  While clinical evaluation tasks show smaller improvements compared to wearable ones, the consistent gains highlight the model’s cross-domain generalizability.

The only task on which Pulse-PPG underperforms is the Sleep Disturbance task. This can be attributed to the nature of Pulse-PPG's pre-training dataset, MOODS, where participants were instructed to wear the smartwatch only during waking hours \citep{neupane2024momentarymoods}. Consequently, Pulse-PPG was not exposed to PPG signals during sleep in its pre-training phase, and human physiology during sleep is significantly different from that of waking hours due to natural biological processes~\cite{smets2018large}. In contrast, PaPaGei's pre-training included a sleep study dataset \citep{chen2015mesa}.

\subsection{Experiment 2: \edit{Pulse-PPG vs. General Time-series FMs and Supervised Baselines}} \label{sec:expbaselinegeneralfoundation}

\begin{table*}[!t]
    \centering
    \LARGE
    \color{black}
    \captionsetup{labelfont={color=black}, textfont={color=black}, font={small}}
    \caption{\textbf{Pulse-PPG vs. PPG FM, General Time-series FM, and Supervised Baselines.} \edit{Mean} Performance across \edit{Classification and Regression Tasks. The best is bolded, and the second is underlined.} Full results can be found in Table\edit{~\ref{tab:baseline_general_TS}}. \edit{Pulse-PPG achieves stronger performance across all metrics than the prior PPG foundation model, PaPaGei. It always maintains Top-2 performance, even among general time-series foundation models, which have significantly higher parameter count (Pulse-PPG with 28.5m vs. Chronos with 200m and MOMENT with 385m) and among fully-supervised models.}    
    }     \vspace{-2mm}
    \label{tab:baseline_general_mean}
    \resizebox{\textwidth}{!}{%
    \begin{tabular}{@{}llccccccccc@{}}
        \toprule
        & & \multicolumn{6}{c}{\textbf{Avg. Classification Performance}} 
        & \multicolumn{3}{c}{\textbf{Avg. Regression Performance}} \\
        \cmidrule(lr){3-8} \cmidrule(lr){9-11}
        {} & \textbf{Model} & {F1 Score} & {Accuracy} & {Precision} & {Recall} & {AUPRC} & {AUROC} & {MAE} & {MSE} & {MAPE} \\
        \midrule
        \multirow{2}{*}{{\edit{PPG FM}}} 
        & Pulse-PPG  & \edit{$\boldsymbol{\textcolor{black}{0.5870} \pm 0.17}$} & \edit{$\boldsymbol{\textcolor{black}{0.6840} \pm 0.17}$} & \edit{\underline{$\textcolor{black}{0.6237} \pm 0.16$}} & \edit{$\boldsymbol{\textcolor{black}{0.5894} \pm 0.17}$} & \edit{\underline{$\textcolor{black}{0.5149} \pm 0.20$}} & \edit{$\boldsymbol{\textcolor{black}{0.7481} \pm 0.17}$} & \edit{\underline{$\textcolor{black}{8.8593} \pm 6.25$}} & \edit{\underline{$\textcolor{black}{145.255} \pm 171.08$}} & \edit{\underline{$\textcolor{black}{0.1009} \pm 0.05$}} \\
        
        & PaPaGei\citep{pillai2024papagei}    & \edit{$0.4994 \pm 0.14$} & \edit{$0.6361 \pm 0.15$} & \edit{$0.5530 \pm 0.14$} & \edit{$0.5210 \pm 0.14$} & \edit{$0.4164 \pm 0.12$} & \edit{$0.6840 \pm 0.10$} & \edit{$10.7488 \pm 7.75$} & \edit{$204.6825 \pm 237.78$} & \edit{$0.1288 \pm 0.10$} \\

        \midrule
        \multirow{2}{*}{{\edit{General FM}}}
        & Chronos\citep{ansari2024chronos}      &\edit{\underline{$\textcolor{black}{0.5845} \pm 0.12$}} & \edit{\underline{$\textcolor{black}{0.6755} \pm 0.14$}} & \edit{$\boldsymbol{\textcolor{black}{0.6386} \pm 0.14}$} & \edit{\underline{$\textcolor{black}{0.5875} \pm 0.13$}} & \edit{$\boldsymbol{\textcolor{black}{0.5189} \pm 0.16}$} & \edit{\underline{$\textcolor{black}{0.7329} \pm 0.15$}} & \edit{$\textcolor{black}{9.1902} \pm 6.40$} & \edit{$\textcolor{black}{162.1719} \pm 211.33$} & \edit{$\textcolor{black}{0.1054} \pm 0.04$} \\
        
        & MOMENT\citep{goswami2024moment}       & \edit{$\textcolor{black}{0.5456} \pm 0.15$} & \edit{$\textcolor{black}{0.6691} \pm 0.15$} & \edit{$\textcolor{black}{0.6059} \pm 0.15$} & \edit{$\textcolor{black}{0.5603} \pm 0.15$} & \edit{$\textcolor{black}{0.4902} \pm 0.19$} & \edit{$\textcolor{black}{0.7241} \pm 0.16$} & \edit{$\boldsymbol{\textcolor{black}{8.5742} \pm 6.09}$} & \edit{$\boldsymbol{\textcolor{black}{139.6036} \pm 168.57}$} & \edit{$\boldsymbol{\textcolor{black}{0.0988} \pm 0.05}$} \\
        \midrule
        \multirow{2}{*}{{\edit{Supervised}}}
        & 1D ResNet-26     & \edit{$\textcolor{black}{0.4623} \pm 0.16$} & \edit{$\textcolor{black}{0.6212} \pm 0.17$} & \edit{$\textcolor{black}{0.4684} \pm 0.18$} & \edit{$\textcolor{black}{0.4944} \pm 0.14$} & \edit{$\textcolor{black}{0.4025} \pm 0.15$} & \edit{$\textcolor{black}{0.6224} \pm 0.18$} & \edit{$\textcolor{black}{17.2733} \pm 18.49$} & \edit{$\textcolor{black}{580.3867} \pm 961.79$} & \edit{$\textcolor{black}{0.1803} \pm 0.15$} \\
        
        & Rand Forest   & \edit{$\textcolor{black}{0.4152} \pm 0.10$} & \edit{$\textcolor{black}{0.5911} \pm 0.15$} & \edit{$\textcolor{black}{0.4671} \pm 0.12$} & \edit{$\textcolor{black}{0.4337} \pm 0.10$} & \edit{$\textcolor{black}{0.3328} \pm 0.08$} & \edit{$\textcolor{black}{0.5977} \pm 0.09$} & \edit{$\textcolor{black}{12.5655} \pm 7.70$} & \edit{$\textcolor{black}{268.2220} \pm 283.46$} & \edit{$\textcolor{black}{0.1495} \pm 0.08$} \\
        \bottomrule
    \end{tabular}
    }
\end{table*}

\subsubsection{Background} To evaluate the effectiveness of Pulse-PPG, we compare it against additional baseline models. We include general-purpose time series foundation models, which are widely adopted for benchmarking due to their large-scale and diverse pretraining datasets, in this comparison. Specifically, we benchmark against Chronos~\cite{ansari2024chronos} and MOMENT~\cite{goswami2024moment}, two well-known general-purpose time series foundation models in this space. Chronos trains language models with suitable adaptations to time series data from domains that include energy, finance, economics, nature, retail, and transportation. MOMENT uses reconstruction of masked embeddings as a pre-training task, on datasets covering healthcare, human physiology, nature, power, economics, traffic, and weather. For supervised baselines, from classical machine learning, we use Random Forest, while from deep learning, we employ 1D ResNet-26, the same backbone used in the original Pulse-PPG model.

\subsubsection{Experimental Design:} 
We train a Random Forest model with statistical features—mean, median, maximum, minimum, 25\textsuperscript{th} percentile, and 75\textsuperscript{th} percentile, computed from raw PPG windows corresponding to ground truth from respective downstream datasets. To train the ResNet model, we append a dense layer with output units equal to the number of classes, followed by an activation layer. For regression tasks, we use a dense layer with a single output unit without activation. Chronos (200 million) and MOMENT (385 million) produce embeddings of size 768 and 1024, respectively \edit{(compared to Pulse-PPG's 28.5 million parameters with an embedding size of 512)}. We extract embeddings from these models for respective input windows from each of the downstream datasets and apply the same linear probe pipeline for classification and regression tasks.   

\subsubsection{Results:} We present downstream task performance of Pulse-PPG against supervised baselines and general-purpose time series models in Table~\ref{tab:baseline_general_TS} in the Appendix. We summarize the results with their average task-relevant (classification/regression) performance metrics in Table~\ref{tab:baseline_general_mean}. Pulse-PPG was trained on a single dataset in a self-supervised fashion, yet it demonstrates balanced generalizability across diverse downstream tasks with a simple linear probe evaluation. Whereas, 1D ResNet-26 and Random Forest were trained separately on each downstream dataset. While ResNet-26 achieves exceptional performance for certain tasks, e.g., an F1-score of 0.7903 for PPG-DaLiA multiclass activity detection, its performance in multiclass stress classification for WESAD with an F1-score of 0.1998 is much worse. Pulse-PPG outperforms the Random Forest model on the majority of tasks. 
     
General-purpose time series models benefit from exposure to diverse datasets, both in terms of nature and scale. This diversity enables them to learn periodicities and trends across contexts, helping them capture complex temporal patterns. Hence, although general-purpose time series models may outperform modality-specific foundation models like Pulse-PPG and PaPaGei in certain tasks, e.g., multi-class activity detection, the value of pretraining on noisy, real-world field PPG data is evident. As shown in Table~\ref{tab:baseline_general_mean}, Pulse-PPG consistently ranks among the top two across a wide range of tasks focused on diverse applications in the physiological domain.

\subsection{Experiment \edit{3}: Assessing Field-to-Lab Generalizability} \label{sec:expfieldtolab}

\subsubsection{Background:} Traditionally, mHealth models were built and tested in controlled lab settings due to a lack of diverse field datasets. As more real-world data became available, researchers began applying these lab-trained models to field data. However, we hypothesize that pre-training foundation models directly on diverse field data will yield better generalizability and performance.

\subsubsection{Experimental Design} 
In these experiments, we assess how pre-training on wearable field PPG compares to pre-training on clean PPG. Similar to the prior experiments in Section \ref{sec:exppapagei}, we utilize a linear probe for evaluation to ensure fair comparisons. From the prior experiments, we already have our Pulse-PPG pre-trained on field PPGs and PaPaGei pre-trained on the clinical PPGs.

In order to construct a large clean clinical PPG pre-training dataset for re-training Pulse-PPG, we utilize the curated MIMIC-III PPG waveform dataset from \citep{xu2022pulseimpute}. This dataset is composed of 151,738 100 Hz 5-minute-long clean clinical PPGs from 18,210 patients.\footnote[4]{Unfortunately, PaPaGei did not release their full curation code for constructing their clinical pre-training dataset, making it impossible for us to train on the same clinical data. However, the MIMIC-III waveform did form a significant portion of their dataset.} We then use this clean clinical PPG dataset to pre-train our Pulse-PPG model, following the same training procedures and hyperparameters that we used when we were pre-training on the MOODS field PPG dataset.

Next, we pre-train PaPaGei from scratch with the MOODS field dataset, using its original training procedure from their codebase located here: \url{https://github.com/Nokia-Bell-Labs/PaPaGei-foundation-model}. 

\begin{figure}
    \centering
    \vspace{-3mm}
    \includegraphics[width=.9\linewidth]{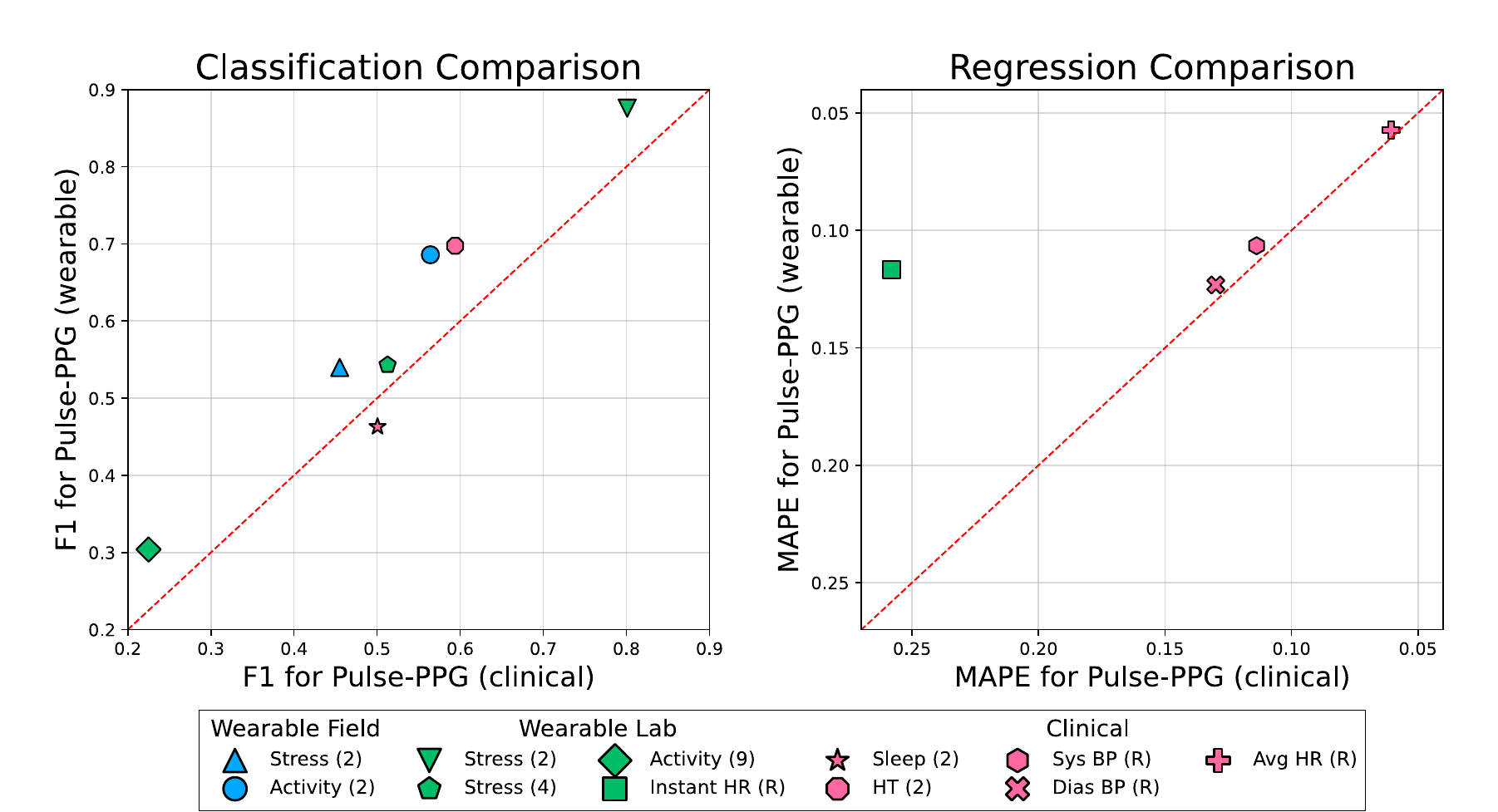}
    \Description{This figure compares the generalizability of models pre-trained on wearable field data versus clinical data. It shows classification performance via F1 score and regression accuracy via MAPE, highlighting better field-to-lab generalizability from wearable data.} \vspace{-2mm}
    \caption{\textbf{Comparison of Pre-training on Wearable Field data vs. Clinical Data to Assess Field-to-Lab Generalizability}. The two plots show the relative performance of a linear probe evaluation for each model, with the left for classification via F1 score and the right for regression via Mean Average Percentage Error. 10/11 of the task data points reside above the slope, demonstrating how pre-training on wearable field data is highly generalizable, demonstrating strong field-to-lab generalizability and even field-to-clinical generalizability. This is contrary to the current standard line of research, which tends to focus on lab-to-field methods.  See further discussion in Section \ref{sec:pretrainresults}.}  
    \label{fig:pretrainwearablevsclinical}
    \vspace{-2mm}
\end{figure}

\subsubsection{Results} \label{sec:pretrainresults} Figure \ref{fig:pretrainwearablevsclinical} visualizes the difference in performance between Pulse-PPG pre-trained on the wearable field vs.\ clinical PPGs. Full results, including comparisons of PaPaGei pre-training with wearable field vs. clinical data, can be found in Table \ref{tab:exp23} in the Appendix, while referencing the second row named "Pre-Train Data". 

Pulse-PPG consistently exhibits stronger performance on 10 out of 11 tasks when pre-trained on the Field Wearable PPG data versus when pre-trained on the clean Clinical PPG.  For all other tasks except for sleep disturbance, the field wearable pre-training yielded consistently stronger results, sometimes substantially so. For example, for Field Wearable datasets, the F1 score for stress improved by .0848, and the F1 score for activity improved by .1218. Even for the clinical tasks, we see that pre-training on the wearable PPGs achieves much stronger performance (i.e., .10 F1 improvement in hypertension prediction and $\sim1$ MAE improvement in systolic and diastolic BP regression). In some cases, such as in the WESAD evaluation datasets, one metric was worse, but all of the remaining metrics were stronger. 

Pre-training Pulse-PPG on field wearable data appears to yield a more robust and generalizable representation, outperforming clinical pre-training, even on clinical tasks. Although performance gains in clinical tasks are modest, the consistency demonstrates that pre-training on field wearable data performs as well as, and sometimes better than, pre-training on clinical data. As discussed earlier in Section \ref{sec:badonsdb}, relatively poor results on sleep can be explained by the lack of sleep data present in the MOODS dataset.

This trend of stronger performance when trained on wearable data is also observed in PaPaGei, though to a lesser extent. When pre-trained on wearable data, PaPaGei outperforms clinical pre-training in 6 out of 11 tasks, including five wearable PPG evaluation datasets (Activity - MOODS, Instant HR - PPG-DaLiA, Activity - PPG-DaLiA, Stress (2) - WESAD, and Stress (4) - WESAD). This aligns with the idea that pre-training and evaluating on wearable data introduces less domain shift. The difference in performance across pre-training settings between Pulse-PPG and PaPaGei can be attributed to their pre-training strategies. PaPaGei relies on PPG-specific morphologies, which may not generalize well across domains with varying PPG patterns (e.g., wearable vs. clinical). In contrast, Pulse-PPG leverages a learnable motif-similarity distance, enabling it to capture more general PPG patterns, which are applicable across different settings.

% \newpage
\subsection{Experiment \edit{4}: Evaluation of Pulse-PPG Use-cases}\label{sec:finetuneeval}

\subsubsection{Background} In this section, we quantitatively measure the effectiveness of Pulse-PPG for 11 downstream tasks, to provide further insight into its utility to the research community.

\subsubsection{Experimental Design} We evaluate downstream performance using two standard evaluation methodologies:
\begin{itemize}[leftmargin=.7cm,  noitemsep]
    \item Linear Probe: The model’s embeddings are frozen, and a logistic classifier or linear regression is trained on the embedding vectors, consistent with prior foundation model work \citep{abbaspourazad2023large, devlin2018bert, oquab2023dinov2}. This is a simple evaluation designed to evaluate the quality of the learned representations to generalize to new tasks.  By requiring minimal computational resources, it also enables a broader range of researchers to utilize the model’s learned representations. Introducing a more sophisticated traditional-ML-based probe would confound the results, with it being more difficult to isolate representation quality from the strength of the classifier.
    \item Fine Tune: This evaluation is designed to boost task-specific performance by training the entire network end-to-end to adapt to a specific task. This highlights the adaptability of the foundation model, showing its potential to serve as a strong starting point for specialized tasks. While computationally intensive, it is essential for applications requiring peak performance.
\end{itemize}

\begin{wrapfigure}{r}{0.41\textwidth}
\centering
    \vspace{-1mm}
    \captionof{table}{\edit{\textbf{Pulse-PPG performance across wearable and clinical PPG tasks.} There are three evals: Naive, predicting majority class or mean; Linear Probing, which trains a simple head on the Pulse-PPG's frozen embeddings; and Fine-Tuning, which trains Pulse-PPG's network end-to-end.}
    }
    \vspace{-.2cm} \label{tab:exp1}
    \resizebox{.41\textwidth}{!}{%
    \begin{tabular}{>{\raggedright}p{.2cm}>{\centering}p{2.8cm}>{\raggedright}p{1.1cm}>{\raggedleft}p{.01cm}ccc} 
    \toprule
     &  &  & & & \multicolumn{2}{c}{\textcolor{black}{\textbf{Pulse-PPG}}}\\
    \midrule
    & \multicolumn{3}{r}{\textbf{Eval Method:} } & Naive & LP & FT \\
    \midrule
    \multirow{21}{*}{\textbf{\rotatebox{90}{\centering Wearable Lab}}}
    & \multirow{3}{*}{\centering\parbox{2.3cm}{\centering \textbf{Instant HR (R)} \\ PPG-DaLiA}}
    & MAE  & \multirow{3}{*}{\rotatebox[origin=c]{-90}{\parbox{.3cm}{\rightarrowfill}}} & 19.34 & \edit{\underline{\textcolor{black}{8.936}}} & \textbf{3.705} \\
    & & MSE  &  & 522.7 & \edit{\underline{\textcolor{black}{149.3}}} & \textbf{59.81} \\
    & & MAPE  & & 0.2742 & \edit{\underline{\textcolor{black}{0.1166}}} & \textbf{0.0484} \\
    \addlinespace[-1pt]
    \cmidrule(lr){2-7}
    \addlinespace[-3pt]
    & \multirow{6}{*}{\centering \parbox{2cm}{\centering \textbf{Activity (9)} \\ PPG-DaLiA}}
    & F1  & \multirow{6}{*}{\rotatebox[origin=c]{90}{\parbox{.5cm}{\rightarrowfill}}} & 0.0455 & \edit{\underline{\textcolor{black}{0.3039}}} & \textbf{0.3648} \\
    & & Accuracy  &  & 0.2572 & \edit{\underline{\textcolor{black}{0.4104}}} & \textbf{0.4234} \\
    & & Precision & & 0.0286 & \edit{\underline{\textcolor{black}{0.3882}}} & \textbf{0.4398} \\
    & & Recall  & & 0.1111 & \edit{\underline{\textcolor{black}{0.3107}}} & \textbf{0.3663} \\
    & & AUPRC  & & 0.0728 & \edit{\underline{\textcolor{black}{0.3504}}} & \textbf{0.4056} \\
    & & AUROC  & & 0.5000 & \edit{\underline{\textcolor{black}{0.8051}}} & \textbf{0.8246} \\
    \addlinespace[-1pt]
    \cmidrule(lr){2-7}
    \addlinespace[-3pt]
    & \multirow{6}{*}{\centering \parbox{2cm}{\centering \textbf{Stress (2)} \\ WESAD}}
    & F1  & \multirow{6}{*}{\rotatebox[origin=c]{90}{\parbox{.5cm}{\rightarrowfill}}} & 0.4065 & \edit{\underline{\textcolor{black}{0.8759}}} & \textbf{0.9392} \\
    & & Accuracy  & & 0.6849 & \edit{\underline{\textcolor{black}{0.8904}}} & \textbf{0.9452} \\
    & & Precision & & 0.3425 & \edit{\underline{\textcolor{black}{0.8688}}} & \textbf{0.9259} \\
    & & Recall  & & 0.5000 & \edit{\underline{\textcolor{black}{0.8848}}} & \textbf{0.9600} \\
    & & AUPRC  & & 0.3151 & \edit{\underline{\textcolor{black}{0.9410}}} & \textbf{0.9790} \\
    & & AUROC  & & 0.5000 & \edit{\underline{\textcolor{black}{0.9687}}} & \textbf{0.9852} \\
    \addlinespace[-1pt]
    \cmidrule(lr){2-7}
    \addlinespace[-3pt]
    & \multirow{6}{*}{\centering \parbox{2cm}{\centering \textbf{Stress (4)} \\ WESAD}}
    & F1  & \multirow{6}{*}{\rotatebox[origin=c]{90}{\parbox{.5cm}{\rightarrowfill}}} & 0.1496 & \edit{\underline{\textcolor{black}{0.5431}}} & \textbf{0.5582} \\
    & & Accuracy  & & 0.4270 & \edit{\underline{\textcolor{black}{0.6517}}} & \textbf{0.7079} \\
    & & Precision & & 0.1067 & \textbf{0.6003} & \edit{\underline{\textcolor{black}{0.5515}}} \\
    & & Recall  & & 0.2500 & \edit{\underline{\textcolor{black}{0.5716}}} & \textbf{0.5812} \\
    & & AUPRC  & & 0.2584 & \edit{\underline{\textcolor{black}{0.5671}}} & \textbf{0.5898} \\
    & & AUROC  & & 0.5000 & \textbf{0.7807} & \edit{\underline{\textcolor{black}{0.7630}}} \\
    
    \midrule
    \addlinespace[-0.5pt]
    \multirow{12}{*}{\textbf{\rotatebox{90}{\centering Wearable Field}}}
    & \multirow{6}{*}{\centering \parbox{2cm}{\centering \textbf{Stress (2)} \\ MOODS}}
    & F1  & \multirow{6}{*}{\rotatebox[origin=c]{90}{\parbox{.5cm}{\rightarrowfill}}} & 0.3859 & \edit{\underline{\textcolor{black}{0.5398}}} & \textbf{0.5872} \\
    & & Accuracy  & & \textbf{0.6285} & 0.6156 & \edit{\underline{\textcolor{black}{0.6247}}} \\
    & & Precision &  & 0.3143 & \edit{\underline{\textcolor{black}{0.5606}}} & \textbf{0.5911} \\
    & & Recall  & & 0.5000 & \edit{\underline{\textcolor{black}{0.5460}}} & \textbf{0.5859} \\
    & & AUPRC  & & 0.3715 & \edit{\underline{\textcolor{black}{0.4275}}} & \textbf{0.5729} \\
    & & AUROC  & & 0.5000 & \edit{\underline{\textcolor{black}{0.5691}}} & \textbf{0.5909} \\    
    \addlinespace[-1pt]
    \cmidrule(lr){2-7}
    \addlinespace[-3pt]
    & \multirow{6}{*}{\centering\parbox{2cm}{\centering \textbf{Activity (2)} \\ MOODS}}
    & F1  & \multirow{6}{*}{\rotatebox[origin=c]{90}{\parbox{.5cm}{\rightarrowfill}}} & 0.4574 & \edit{\underline{\textcolor{black}{0.6859}}} & \textbf{0.7992} \\
    & & Accuracy  & & 0.8430 & \edit{\underline{\textcolor{black}{0.8705}}} & \textbf{0.8930} \\
    & & Precision &  & 0.4215 & \edit{\underline{\textcolor{black}{0.7835}}} & \textbf{0.7974} \\
    & & Recall  & & 0.5000 & \edit{\underline{\textcolor{black}{0.6509}}} & \textbf{0.8010} \\
    & & AUPRC  & & 0.1570 & \edit{\underline{\textcolor{black}{0.5835}}} & \textbf{0.8596} \\
    & & AUROC  & & 0.5000 & \edit{\underline{\textcolor{black}{0.8722}}} & \textbf{0.9196} \\
    
    \midrule
    \addlinespace[-0.5pt]
    \multirow{21}{*}{\textbf{\rotatebox{90}{\centering Clinical}}} 
    & \multirow{6}{*}{\centering\parbox{2.4cm}{\centering \textbf{Sleep Disturbance (2)} \\ SDB}} 
    & F1  & \multirow{6}{*}{\rotatebox[origin=c]{90}{\parbox{.5cm}{\rightarrowfill}}} & 0.3945 & \edit{\underline{\textcolor{black}{0.4630}}} & \textbf{0.5782} \\
    & & Accuracy  & & \textbf{0.6515} & 0.5364 & \edit{\underline{\textcolor{black}{0.6428}}} \\
    & & Precision & & 0.3258 & \edit{\underline{\textcolor{black}{0.4633}}} & \textbf{0.5911} \\
    & & Recall  & & \edit{\underline{\textcolor{black}{0.5000}}} & 0.4673 & \textbf{0.5772} \\
    & & AUPRC  & & \edit{\underline{\textcolor{black}{0.3485}}} & 0.3085 & \textbf{0.5904} \\
    & & AUROC  & & \edit{\underline{\textcolor{black}{0.5000}}} & 0.4437 & \textbf{0.6243} \\
    \addlinespace[-1pt]
    \cmidrule(lr){2-7}
    \addlinespace[-3pt]
    & \multirow{3}{*}{\centering\parbox{2.4cm}{\centering \textbf{Systolic BP (R)} \\ PPG-BP}} 
    & MAE  & \multirow{3}{*}{\rotatebox[origin=c]{-90}{\parbox{.3cm}{\rightarrowfill}}} & 17.21 & \edit{\underline{\textcolor{black}{13.62}}} & \textbf{12.33} \\
    & & MSE  & & 455.9 & \edit{\underline{\textcolor{black}{286.5}}} & \textbf{280.7} \\
    & & MAPE  & & 0.1345 & \edit{\underline{\textcolor{black}{0.1065}}} & \textbf{0.0939} \\
    \addlinespace[-1pt]
    \cmidrule(lr){2-7}
    \addlinespace[-3pt]
    & \multirow{3}{*}{\centering\parbox{2.5cm}{\centering \textbf{Diastolic BP (R)} \\ PPG-BP}} 
    & MAE  & \multirow{3}{*}{\rotatebox[origin=c]{-90}{\parbox{.3cm}{\rightarrowfill}}} & 10.12 & \edit{\underline{\textcolor{black}{8.878}}} & \textbf{8.695} \\
    & & MSE  & & 147.9 & \textbf{118.2} & \edit{\underline{\textcolor{black}{119.8}}} \\
    & & MAPE  & & 0.1407 & \edit{\underline{\textcolor{black}{0.1232}}} & \textbf{0.1209} \\
    \addlinespace[-1pt]
    \cmidrule(lr){2-7}
    \addlinespace[-3pt]
    & \multirow{3}{*}{\centering\parbox{2.4cm}{\centering \textbf{Average HR (R)} \\ PPG-BP}} 
    & MAE  & \multirow{3}{*}{\rotatebox[origin=c]{-90}{\parbox{.3cm}{\rightarrowfill}}} & 7.965 & \edit{\underline{\textcolor{black}{4.003}}} & \textbf{3.697} \\
    & & MSE  & & 98.08 & \edit{\underline{\textcolor{black}{27.02}}} & \textbf{23.26} \\
    & & MAPE  & & 0.1176 & \edit{\underline{\textcolor{black}{0.0573}}} & \textbf{0.0531} \\
    \addlinespace[-1pt]
    \cmidrule(lr){2-7}
    \addlinespace[-3pt]
    & \multirow{6}{*}{\centering\parbox{2.8cm}{\centering \textbf{Hypertension (2)} \\ PPG-BP}} 
    & F1  & \multirow{6}{*}{\rotatebox[origin=c]{90}{\parbox{.5cm}{\rightarrowfill}}} & 0.4459 & \edit{\underline{\textcolor{black}{0.6975}}} & \textbf{0.7213} \\
    & & Accuracy  & & \edit{\underline{\textcolor{black}{0.8049}}} & \textbf{0.8130} & 0.7967 \\
    & & Precision & & 0.4024 & \edit{\underline{\textcolor{black}{0.7009}}} & \textbf{0.7031} \\
    & & Recall  & & 0.5000 & \edit{\underline{\textcolor{black}{0.6944}}} & \textbf{0.7633} \\
    & & AUPRC  & & 0.1951 & \edit{\underline{\textcolor{black}{0.4266}}} & \textbf{0.7408} \\
    & & AUROC  & & 0.5000 & \edit{\underline{\textcolor{black}{0.7971}}} & \textbf{0.8258} \\
    \bottomrule    
    \end{tabular}%
    } \vspace{-1.47cm}
\end{wrapfigure}

Implementation details can be found in Section~\ref{sec:evalmethods}. We also include a naive baseline for clear comparison. The naive classifier always predicts the majority class, and the naive regressor predicts the training mean. Though simple, this benchmark contextualizes our model’s performance, highlighting the value of learned representations and ensuring transparent evaluation.

\subsubsection{Results} \label{sec:resultsfinetune}

Table \ref{tab:exp1} shows that Pulse-PPG always outperforms the naive baseline across all tasks. When evaluated with a linear probe, the model’s frozen embeddings already yield competitive performance, highlighting the strength of the learned representations. For example, in the wearable field stress detection task, a traditionally difficult task, especially in the field setting, where noise and variability are inherent, a simple linear probe can achieve an F1 score of 0.5398, compared to the naive method achieving an F1 of 0.3859. Additionally, in the binary stress classification for the WESAD dataset, the simple logistic regression on our learned feature can achieve an F1 of 0.8759, more than doubling the naive model's F1 score of 0.4065. % We also summarize task-relevant (classification/regression) mean performance metrics results in Table~\ref{tab:linearprobe_vs_finetuning_vs_naive}.

Fine-tuning further enhances performance by adapting the entire network to the specific task. This is evident across all tasks, where fine-tuning yields the best results. Although occasionally, linear probing may outperform fine-tuning for one metric, for the remaining metrics, the fine-tuned results are stronger. % In fact, the much stronger performance becomes evident from Table~\ref{tab:linearprobe_vs_finetuning_vs_naive}, where Pulse-PPG, when fine-tuned, appears as the winner across all metrics for both classification and regression tasks. 
It can be observed that in some cases, fine-tuning results in significantly higher performance, such as in PPG-DaLiA instantaneous HR regression, from 8.936 $\rightarrow$ 3.705, compared to more marginal PPG-BP systolic BP regression, from 13.62 $\rightarrow$ 12.33. This makes sense as Table \ref{tab:ppg_comparison} shows that PPG-DaLiA contains much more data, with 1545 subject-minutes, compared to PPG-BP's 36.5 subject-minutes. These fine-tuning results demonstrate the model’s adaptability and its potential to serve as a robust starting point for researchers to adapt for their specialized applications. 
 
The linear probe evaluation offers a computationally efficient means to leverage powerful representations, while fine-tuning maximizes task-specific performance. Together, they highlight the versatility and practical utility of foundation models in both exploratory and application-focused research settings.

\section{Discussions}\label{sec:discussions}

\subsection{Shifting the Paradigm: From Task-Specific Models to Foundation Models for PPG}\label{sec:task_to_foundation}

Early mHealth research was limited by the lack of large-scale field datasets, prompting researchers to collect data in controlled lab environments. A common approach involved recruiting a group of participants, collecting physiological data in a lab setting, training a model on this clean data with clean labels, and then deploying the model on field data from the same participants~\cite{ertin2011autosense, plarre2011continuous}. This systematic approach minimizes variability between training and deployment and generally results in better performance, as shown in recent works~\cite{toshnazarov2024sosw}. For greater between-person generalizability, researchers trained the model on one group of lab participants and tested it on a separate group of lab participants~\cite{Hovsepian-2015-cStress}. For lab-to-field generalizability, the model was tested on a third group of participants who collected data in field settings~\cite{Hovsepian-2015-cStress}. A further step toward real-world applicability involved training AI models directly on growing volumes of field data and testing them on the same participants using 10-fold cross-validation or cross-subject validation~\cite{mishra2018investigating,zhang2024reproducible}. This approach posed significant challenges due to the inherent noise and variability in real-world data, making model development more complex. The majority of these works~\cite{mishra2018investigating,zhang2024reproducible} report F1 scores of around 0.5 for stress classification when using PPG only data. 

A key limitation of task-specific models is their inflexibility—adapting to even a slightly different task (e.g., transitioning from binary to multi-class classification) often requires substantial retraining and modifications to the modeling pipeline. In addition, traditional task-specific models like Random Forest rely on hand-engineered features, and their performance is limited by the discriminative power of those features. In contrast, models like ResNet-26 learns features automatically from the data, leading to better performance within the problem space we examine, as can be observed from the respective rows for Random Forest and ResNet models in Table~\ref{tab:baseline_general_mean}. However, because powerful supervised models like ResNet-26 are trained on limited labeled data for each task, they often struggle to learn generalizable representations without much larger datasets. This is where foundation models can become particularly useful, unlike task-specific supervised models, by learning representations from large-scale unlabeled data, enabling them to generalize across diverse tasks without needing extensive labeled datasets for each task. The performance numbers of Pulse-PPG with a simple linear probe pipeline in Table~\ref{tab:baseline_general_mean} provide its evidence where Pulse-PPG outperforms both Random Forest and ResNet models across all performance metrics. Importantly, with the performance results from this experiment, the transition---from traditional hand-engineered models, to deep learning models that discover features with supervision, to general-purpose foundation models that learn without explicit task label, sets the future direction for AI in mHealth applications.

\subsection{Field-to-Lab Generalizability for mHealth Modeling from PPG Data}\label{sec:field_to_lab}
Cleaner data has been a major advantage of lab datasets. For detecting psychological states such as stress, another major advantage of lab data is clean labels. In the field conditions, the labels are usually self-reported. They do not temporally align with the physiological data due to a large time lag between how quickly physiology recovers versus the perception of stress. Further, unlike lab data, where an entire block of data corresponds to a stress or non-stress task, blocks of data receiving the same label can consist of a mixture of stress and non-stress segments, due to a lack of control over the experiences of a participant. These label impurities make it difficult to train task-specific models that use the labels together with the data.

Unlike traditional task-specific models~\cite{choi2022attention,hasanpoor2022stress,motaman2022stress,alshareef2022transformer,mitro2023ai}, self-supervised learning-based models are designed to reduce or eliminate the dependency on such explicit labels~\cite{spathis2022breaking,haresamudram2022assessing,rabbani2022contrastive}. This is achieved through their inherent structure, where the pre-training process derives labels directly from the data itself, without needing human-provided annotations. This self-supervised approach represents a significant departure from conventional methods, as it taps into the underlying patterns of the data rather than relying on predefined labels. 

Our results from the pre-trained foundation model on field data demonstrate this potential for PPG modeling. Pulse-PPG, pre-trained on field data, outperforms its counterpart trained on clinical data in 10/11 tasks. PaPaGei, which was originally trained on clinical datasets, hypothetically should perform better than field-trained PaPaGei, especially when evaluated on clinical datasets. But, \edit{a closer look at the results in Table~\ref{tab:pulseppg_vs_papagei_mean} shows that} field-trained PaPaGei performs better than clinically trained PaPaGei across all metrics. These results show that training on noisy field data with a well-matched pre-training approach can enhance field-to-lab generalizability for PPG-specific foundation models. 

To our knowledge, this is the first work to demonstrate the benefits of pre-training a foundation model on field data for potentially improving performance not just in real-world, field-based mHealth tasks but also in more traditional clinical and lab environments. Our preliminary findings challenge the conventional belief that PPG models must be trained on clean, highly curated datasets to perform well in clinical and lab settings.

\subsection{Optimizing Task Specific Performance}\label{sec:optimize_finetune}
Previous research primarily focused on task-specific performances. The foundation models, on the other hand, are general-purpose. But, an added benefit of foundation models is that they can also be optimized for enhancing the performance of specific tasks. When the focus is more on enhancing task-specific performance with pre-trained models, instead of using a linear probe approach to test generalizability (which keeps the pre-trained model weights frozen and simply adds a linear layer on top for training), fine-tuning is the more appropriate approach that involves training the model end-to-end, using the pre-trained model weights as a starting point. This allows the model to adapt to the task-specific characteristics of the data more closely, which enables it to perform better for that task. Our results show that when we fine-tuned Pulse-PPG for downstream tasks, the average F1 score increased by as much as $9.6\%$, and the MAE decreased by $19.78\%$ as seen in Table~\ref{tab:exp1}. Hence, our work provides a new starting point, enabling the community to build upon using the model weights of Pulse-PPG for more accurate task-specific results.

\subsection{Role of Relevant Data in Pre-training Foundation Models}\label{sec:role_relevant_data}
We evaluated Pulse-PPG against PaPaGei, another PPG-specific foundation model, and Chronos and MOMENT, two general-purpose time series foundation models, across a variety of downstream tasks. We draw several key insights from the results. First, two evaluation datasets—PPG-DaLiA (activity classification, heart rate estimation) and SDB (sleep disorder classification)—highlight the importance of training on relevant datasets that may represent physiological activation different from usual physiological activation. In sleep disorder classification, PaPaGei significantly outperforms Pulse-PPG. PaPaGei was exclusively trained on MESA, a sleep study involving 2,237 participants, enabling it to excel at this task. Conversely, in multiclass activity detection and heart rate estimation using the PPG-DaLiA dataset, which includes high-intensity activities like cycling and stair climbing, PaPaGei performs significantly worse than Pulse-PPG. PaPaGei's training data came from ICU patients, patients during surgery, and sleeping participants—contexts with minimal/no physical motion—whereas activity recognition tasks benefit from motion-rich pre-training. Pulse-PPG, although not explicitly trained on data from intense activities, demonstrates better generalization due to a much higher probability of activity-induced physiological activation arising from the natural environment in its pre-training data. These observations underline the importance of using relevant datasets during pre-training to enhance generalizability. 

Second, the absence of relevant data can potentially be compensated for if a foundation model is trained with a large variety of time-series data. This is demonstrated by the competitive performance of Chronos and MOMENT (see Table~\ref{tab:baseline_general_mean}). But, training with a large variety of datasets comes at a cost of significant time and computing resources as both of these models have 7x to 13.5x more parameters than Pulse-PPG.

\section{Limitations}\label{sec:limitations}

\subsection{Stress Downstream Task in Field Conditions}
One limitation of this study is that the stress \edit{and non-stress} episodes used for the field downstream task of stress vs.\ non-stress classification were obtained from a dataset~\cite{neupane2024momentarymoods} that used a commercial pre-trained model to obtain self-reported stress ratings. Other datasets have employed different approaches to obtain stress ratings~\cite{toshnazarov2024sosw,zhang2024reproducible}. Although the competing foundation model was also tested on the same dataset, testing the model performance on multiple large field datasets can demonstrate greater generalizability for stress detection in the field.

\subsection{Impact of Demographic Diversity on Foundation Models}\label{sec:lack_of_demographic}
While the field dataset we pretrained with included participants from diverse professional and educational backgrounds, it did not collect race or ethnicity. The lack of skin tone data limits our ability to assess how it may have impacted PPG signal computation\edit{~\cite{puranen2020effect, hu2022effect}}, and eventually the quality of learned representations. Although Pulse-PPG's motif-based architecture may have abstracted some of the challenges, future work should explicitly examine how demographic factors may influence signal quality and subsequently model's performance on downstream datasets. Selectively incorporating demographic factors, while ensuring fairness and minimizing bias, may provide valuable insights into individual differences in physiological responses for model pre-training. 

\subsection{Impact of Multiple Modalities on Foundation Models}\label{sec:lack_of_multi_modality}
In this work, we explore how a foundation model built solely from noisy raw PPG from the field can learn meaningful embeddings that enable it to perform better across various downstream tasks involving wearable field, lab, and clinical applications. Future work can explore integrating additional data modalities, such as accelerometer data, to fuse information from multiple sources, to more effectively learn how multiple signals interact with each other to impact human physiology. This could potentially have a positive impact on representation learning that can better isolate, e.g., mental stressors from physical activities that are considered as confounds.

\subsection{Expanding Pre-training Data for Broader Generalization}
The wearable field dataset used to pre-train the foundation models in this study originated from a single field study involving a diverse and general population from throughout the U.S. \citep{neupane2024momentarymoods}. While this approach demonstrated the model’s utility, relying on data from one study inherently limits the diversity of field conditions, stress episodes, and physiological responses. To be more inclusive and more robust to field conditions, future work should incorporate data from multiple studies across varied populations, geographic regions, and contexts to enhance generalizability. They can improve the model’s adaptability to new environments and ensure more robust and equitable performance across broader applications.

\subsection{Deployment Feasability of Pulse-PPG} \label{sec:deploy}

\edit{This work proposes a new foundation model to improve the performance of detecting multiple health conditions from PPG data collected in the noisy field environment. For such technologies to ultimately impact real life, they need to be implemented on wearabales. There are scenarios where PPG data may be transferred to a cloud-based server hosting powerful models such as the ones proposed here to obtain inferences. However, for low-latency applications that require real-time inferences or when data cannot leave the device due to privacy concerns, the models need to be adapted for running on battery-powered, resource-constrained wearables.

This work does not present methods for such translation. However, there is precedent for real-life translation of foundation models to run on edge devices.} For instance, in NLP, researchers first developed state-of-the-art foundation models (i.e., GPT-1 in 2018 \citep{radford2018improving}, BERT in 2018 \citep{devlin2018bert}), generalizable across a wide range of language tasks. \edit{Later on they designed} clever engineering methods to enable on-device utilization (i.e., ChatGPT app in 2022 \citep{fitria2023artificial}, LLaMa and DeepSeek on mobile devices in 2024 \citep{li2024large}). \edit{From a broad perspective though, for} mobile health systems, two core \edit{deployment approaches---}on-server and on-device\edit{, can be explored}.

For on-server deployment, prior work \edit{on} cloud-based foundation models has demonstrated the key benefit of leveraging full model performance without wearable hardware constraints. This paradigm \edit{is} already regulatory \edit{accepted}, as evidenced by the FDA-cleared EndoSleep system \citep{miller2025endosleep} for server-based sleep apnea diagnosis from raw PPG signals. While cloud processing introduces network latency, it is \edit{generally acceptable} for non-time-critical PPG applications, such as sleep staging or longitudinal hypertension risk screening \citep{elgendi2019use, lin2025longitudinal}, \edit{provided privacy concerns are adequately addressed for sensitive data transfer to servers. Apart from privacy, other considerations in} data transmission \edit{include} real-time compression \citep{banerjee2023new}, \edit{speed of data} transmission \citep{pvribil2021wearable}, and encryption \citep{he2023compression}.

For low-latency PPG applications, such as stress detection or instantaneous HR regression, on-device deployment becomes more relevant, and recent advances in model compression and hardware acceleration offer many interesting directions. Our Pulse-PPG foundation model utilizes a ResNet architecture as its encoder backbone, and prior works \edit{have investigated} ResNet model compression techniques explicitly for on-edge devices \citep{yao2023intelligent, kulkarni2021ai, chen2020deep}, as well as ResNet-specific hardware accelerator systems \citep{moreau2018leveraging, yang2024implementation, abtahi2018accelerating}.

\edit{The ultimate choice depends on use case needs---favoring server-based deployment for} accuracy and flexibility, or \edit{on-device implementation for} responsiveness \edit{and privacy}. Future work can explore hybrid architectures that \edit{adaptively split computation between edge and cloud} in mobile health monitoring.

\section{Conclusion}
Our work presents Pulse-PPG, the first open-source\footref{open} PPG foundation model pre-trained on field-collected data, underscoring the crucial role of real-world variability in enhancing model robustness and applicability. By outperforming state-of-the-art foundation models trained on wearable lab or clinical data across multiple downstream tasks, Pulse-PPG establishes the necessity of incorporating real-world variability into model training. Additionally, our findings indicate that field-trained models can perform competitively even in controlled lab datasets, challenging conventional assumptions about the superiority of lab-trained models.
The open-source\footref{open} release of Pulse-PPG will allow for further advancements in PPG-based health monitoring, enabling the research community to build upon a model that is inherently designed for real-world deployment and expand the applicability of foundation models across diverse physiological signals.

\edit{
\section{Acknowledgments}
Research reported here was supported by the National Institutes of Health (NIH) under award P41EB028242, by the National Science Foundation (NSF) under awards ACI-1640813 and CNS-1822935, and by the National Science Foundation Graduate Research Fellowship under Grant No. DGE-2039655. Additional support was provided by the Jump ARCHES  endowment for project P435 through the Health Care Engineering Systems Center at Illinois and the OSF Foundation. The opinions expressed in this article are the authors’ own and do not reflect the views of the NIH, NSF, or Jump ARCHES.
}

\bibliographystyle{ACM-Reference-Format}
\bibliography{biblio}


\begin{thebibliography}{119}

%%% the Web address.

\ifx \showCODEN    \undefined \def \showCODEN     #1{\unskip}     \fi
\ifx \showDOI      \undefined \def \showDOI       #1{#1}\fi
\ifx \showISBNx    \undefined \def \showISBNx     #1{\unskip}     \fi
\ifx \showISBNxiii \undefined \def \showISBNxiii  #1{\unskip}     \fi
\ifx \showISSN     \undefined \def \showISSN      #1{\unskip}     \fi
\ifx \showLCCN     \undefined \def \showLCCN      #1{\unskip}     \fi
\ifx \shownote     \undefined \def \shownote      #1{#1}          \fi
\ifx \showarticletitle \undefined \def \showarticletitle #1{#1}   \fi
\ifx \showURL      \undefined \def \showURL       {\relax}        \fi
% invisible to TeX
\providecommand\bibfield[2]{#2}
\providecommand\bibinfo[2]{#2}
\providecommand\natexlab[1]{#1}
\providecommand\showeprint[2][]{arXiv:#2}

\bibitem[Met(2025)]%
        {MetaDINOv2}
 \bibinfo{year}{Accessed April, 2025}\natexlab{}.
\newblock \bibinfo{booktitle}{\emph{How Inarix is using DINOv2 to revolutionize the agricultural supply chain}}.
\newblock
\urldef\tempurl%
\url{https://ai.meta.com/blog/inarix-agricultural-supply-chain-meta-dino-v2/}
\showURL{%
\tempurl}


\bibitem[App(2025)]%
        {AppleIntelligence}
 \bibinfo{year}{Accessed April, 2025}\natexlab{}.
\newblock \bibinfo{booktitle}{\emph{Introducing Apple’s On-Device and Server Foundation Models}}.
\newblock
\urldef\tempurl%
\url{https://machinelearning.apple.com/research/introducing-apple-foundation-models}
\showURL{%
\tempurl}


\bibitem[Abbaspourazad et~al\mbox{.}(2023)]%
        {abbaspourazad2023large}
\bibfield{author}{\bibinfo{person}{Salar Abbaspourazad}, \bibinfo{person}{Oussama Elachqar}, \bibinfo{person}{Andrew~C Miller}, \bibinfo{person}{Saba Emrani}, \bibinfo{person}{Udhyakumar Nallasamy}, {and} \bibinfo{person}{Ian Shapiro}.} \bibinfo{year}{2023}\natexlab{}.
\newblock \showarticletitle{Large-scale training of foundation models for wearable biosignals}.
\newblock \bibinfo{journal}{\emph{arXiv preprint arXiv:2312.05409}} (\bibinfo{year}{2023}).
\newblock


\bibitem[Abtahi et~al\mbox{.}(2018)]%
        {abtahi2018accelerating}
\bibfield{author}{\bibinfo{person}{Tahmid Abtahi}, \bibinfo{person}{Colin Shea}, \bibinfo{person}{Amey Kulkarni}, {and} \bibinfo{person}{Tinoosh Mohsenin}.} \bibinfo{year}{2018}\natexlab{}.
\newblock \showarticletitle{Accelerating convolutional neural network with FFT on embedded hardware}.
\newblock \bibinfo{journal}{\emph{IEEE Transactions on Very Large Scale Integration (VLSI) Systems}} \bibinfo{volume}{26}, \bibinfo{number}{9} (\bibinfo{year}{2018}), \bibinfo{pages}{1737--1749}.
\newblock


\bibitem[Ahmed et~al\mbox{.}(2022)]%
        {ahmed2022ppg}
\bibfield{author}{\bibinfo{person}{Solaiman Ahmed}, \bibinfo{person}{Tanveer~Ahmed Bhuiyan}, {and} \bibinfo{person}{Manabu Nii}.} \bibinfo{year}{2022}\natexlab{}.
\newblock \showarticletitle{PPG signal morphology-based method for distinguishing stress and non-stress conditions}.
\newblock \bibinfo{journal}{\emph{Journal of Advanced Computational Intelligence and Intelligent Informatics}} \bibinfo{volume}{26}, \bibinfo{number}{1} (\bibinfo{year}{2022}), \bibinfo{pages}{58--66}.
\newblock


\bibitem[Ajmal et~al\mbox{.}(2021)]%
        {ajmal2021monte}
\bibfield{author}{\bibinfo{person}{Ajmal}, \bibinfo{person}{Tananant Boonya-Ananta}, \bibinfo{person}{Andres~J Rodriguez}, \bibinfo{person}{VN Du~Le}, {and} \bibinfo{person}{Jessica~C Ramella-Roman}.} \bibinfo{year}{2021}\natexlab{}.
\newblock \showarticletitle{Monte Carlo analysis of optical heart rate sensors in commercial wearables: the effect of skin tone and obesity on the photoplethysmography (PPG) signal}.
\newblock \bibinfo{journal}{\emph{Biomedical optics express}} \bibinfo{volume}{12}, \bibinfo{number}{12} (\bibinfo{year}{2021}), \bibinfo{pages}{7445--7457}.
\newblock


\bibitem[Ali et~al\mbox{.}(2024)]%
        {ali2024comparison}
\bibfield{author}{\bibinfo{person}{Haider Ali}, \bibinfo{person}{Imran~Khan Niazi}, \bibinfo{person}{David White}, \bibinfo{person}{Malik~Naveed Akhter}, {and} \bibinfo{person}{Samaneh Madanian}.} \bibinfo{year}{2024}\natexlab{}.
\newblock \showarticletitle{Comparison of Machine Learning Models for Predicting Interstitial Glucose Using Smart Watch and Food Log}.
\newblock \bibinfo{journal}{\emph{Electronics}} \bibinfo{volume}{13}, \bibinfo{number}{16} (\bibinfo{year}{2024}), \bibinfo{pages}{3192}.
\newblock


\bibitem[Allen(2007)]%
        {allen2007photoplethysmography}
\bibfield{author}{\bibinfo{person}{John Allen}.} \bibinfo{year}{2007}\natexlab{}.
\newblock \showarticletitle{Photoplethysmography and its application in clinical physiological measurement}.
\newblock \bibinfo{journal}{\emph{Physiological measurement}} \bibinfo{volume}{28}, \bibinfo{number}{3} (\bibinfo{year}{2007}), \bibinfo{pages}{R1}.
\newblock


\bibitem[Alshareef et~al\mbox{.}(2022)]%
        {alshareef2022transformer}
\bibfield{author}{\bibinfo{person}{Moudy~Sharaf Alshareef}, \bibinfo{person}{Badraddin Alturki}, {and} \bibinfo{person}{Mona Jaber}.} \bibinfo{year}{2022}\natexlab{}.
\newblock \showarticletitle{A transformer-based model for effective and exportable IoMT-based stress detection}. In \bibinfo{booktitle}{\emph{GLOBECOM 2022-2022 IEEE Global Communications Conference}}. IEEE, \bibinfo{pages}{1158--1163}.
\newblock


\bibitem[Amari(1993)]%
        {amari1993backpropagation}
\bibfield{author}{\bibinfo{person}{Shun-ichi Amari}.} \bibinfo{year}{1993}\natexlab{}.
\newblock \showarticletitle{Backpropagation and stochastic gradient descent method}.
\newblock \bibinfo{journal}{\emph{Neurocomputing}} \bibinfo{volume}{5}, \bibinfo{number}{4-5} (\bibinfo{year}{1993}), \bibinfo{pages}{185--196}.
\newblock


\bibitem[Ansari et~al\mbox{.}(2024)]%
        {ansari2024chronos}
\bibfield{author}{\bibinfo{person}{Abdul~Fatir Ansari}, \bibinfo{person}{Lorenzo Stella}, \bibinfo{person}{Caner Turkmen}, \bibinfo{person}{Xiyuan Zhang}, \bibinfo{person}{Pedro Mercado}, \bibinfo{person}{Huibin Shen}, \bibinfo{person}{Oleksandr Shchur}, \bibinfo{person}{Syama~Sundar Rangapuram}, \bibinfo{person}{Sebastian~Pineda Arango}, \bibinfo{person}{Shubham Kapoor}, \bibinfo{person}{Jasper Zschiegner}, \bibinfo{person}{Danielle~C. Maddix}, \bibinfo{person}{Hao Wang}, \bibinfo{person}{Michael~W. Mahoney}, \bibinfo{person}{Kari Torkkola}, \bibinfo{person}{Andrew~Gordon Wilson}, \bibinfo{person}{Michael Bohlke-Schneider}, {and} \bibinfo{person}{Yuyang Wang}.} \bibinfo{year}{2024}\natexlab{}.
\newblock \bibinfo{title}{Chronos: Learning the Language of Time Series}.
\newblock
\newblock
\showeprint[arxiv]{2403.07815}~[cs.LG]
\urldef\tempurl%
\url{https://arxiv.org/abs/2403.07815}
\showURL{%
\tempurl}


\bibitem[Aqajari et~al\mbox{.}(2024)]%
        {aqajari2024enhancing}
\bibfield{author}{\bibinfo{person}{Seyed Amir~Hossein Aqajari}, \bibinfo{person}{Ziyu Wang}, \bibinfo{person}{Ali Tazarv}, \bibinfo{person}{Sina Labbaf}, \bibinfo{person}{Salar Jafarlou}, \bibinfo{person}{Brenda Nguyen}, \bibinfo{person}{Nikil Dutt}, \bibinfo{person}{Marco Levorato}, {and} \bibinfo{person}{Amir~M Rahmani}.} \bibinfo{year}{2024}\natexlab{}.
\newblock \showarticletitle{Enhancing performance and user engagement in everyday stress monitoring: A context-aware active reinforcement learning approach}.
\newblock \bibinfo{journal}{\emph{arXiv preprint arXiv:2407.08215}} (\bibinfo{year}{2024}).
\newblock


\bibitem[Banerjee and Singh(2023)]%
        {banerjee2023new}
\bibfield{author}{\bibinfo{person}{Soumyendu Banerjee} {and} \bibinfo{person}{Girish~Kumar Singh}.} \bibinfo{year}{2023}\natexlab{}.
\newblock \showarticletitle{A new real-time lossless data compression algorithm for ECG and PPG signals}.
\newblock \bibinfo{journal}{\emph{Biomedical Signal Processing and Control}}  \bibinfo{volume}{79} (\bibinfo{year}{2023}), \bibinfo{pages}{104127}.
\newblock


\bibitem[Bari et~al\mbox{.}(2020)]%
        {bari2020automated}
\bibfield{author}{\bibinfo{person}{Rummana Bari}, \bibinfo{person}{Md~Mahbubur Rahman}, \bibinfo{person}{Nazir Saleheen}, \bibinfo{person}{Megan~Battles Parsons}, \bibinfo{person}{Eugene~H Buder}, {and} \bibinfo{person}{Santosh Kumar}.} \bibinfo{year}{2020}\natexlab{}.
\newblock \showarticletitle{Automated detection of stressful conversations using wearable physiological and inertial sensors}.
\newblock \bibinfo{journal}{\emph{Proceedings of the ACM on interactive, mobile, wearable and ubiquitous technologies}} \bibinfo{volume}{4}, \bibinfo{number}{4} (\bibinfo{year}{2020}), \bibinfo{pages}{1--23}.
\newblock


\bibitem[B{\'e}n{\'e}dict et~al\mbox{.}(2021)]%
        {benedict2021sigmoidf1}
\bibfield{author}{\bibinfo{person}{Gabriel B{\'e}n{\'e}dict}, \bibinfo{person}{Vincent Koops}, \bibinfo{person}{Daan Odijk}, {and} \bibinfo{person}{Maarten de Rijke}.} \bibinfo{year}{2021}\natexlab{}.
\newblock \showarticletitle{SigmoidF1: A smooth F1 score surrogate loss for multilabel classification}.
\newblock \bibinfo{journal}{\emph{arXiv preprint arXiv:2108.10566}} (\bibinfo{year}{2021}).
\newblock


\bibitem[biosignalsplux(2019)]%
        {biosignalsplux}
\bibfield{author}{\bibinfo{person}{biosignalsplux}.} \bibinfo{year}{2019}\natexlab{}.
\newblock
\newblock
\urldef\tempurl%
\url{https://bio-medical.com/media/support/biosignalsplux_explorer_user_manual_v.1.0.pdf}
\showURL{%
\tempurl}


\bibitem[Bommasani et~al\mbox{.}(2021)]%
        {bommasani2021oppoandriskforfm}
\bibfield{author}{\bibinfo{person}{Rishi Bommasani}, \bibinfo{person}{Drew~A Hudson}, \bibinfo{person}{Ehsan Adeli}, \bibinfo{person}{Russ Altman}, \bibinfo{person}{Simran Arora}, \bibinfo{person}{Sydney von Arx}, \bibinfo{person}{Michael~S Bernstein}, \bibinfo{person}{Jeannette Bohg}, \bibinfo{person}{Antoine Bosselut}, \bibinfo{person}{Emma Brunskill}, {et~al\mbox{.}}} \bibinfo{year}{2021}\natexlab{}.
\newblock \showarticletitle{On the opportunities and risks of foundation models}.
\newblock \bibinfo{journal}{\emph{arXiv preprint arXiv:2108.07258}} (\bibinfo{year}{2021}).
\newblock


\bibitem[Chen et~al\mbox{.}(2015)]%
        {chen2015mesa}
\bibfield{author}{\bibinfo{person}{Xiaoli Chen}, \bibinfo{person}{Rui Wang}, \bibinfo{person}{Phyllis Zee}, \bibinfo{person}{Pamela~L Lutsey}, \bibinfo{person}{Sogol Javaheri}, \bibinfo{person}{Carmela Alc{\'a}ntara}, \bibinfo{person}{Chandra~L Jackson}, \bibinfo{person}{Michelle~A Williams}, {and} \bibinfo{person}{Susan Redline}.} \bibinfo{year}{2015}\natexlab{}.
\newblock \showarticletitle{Racial/ethnic differences in sleep disturbances: the Multi-Ethnic Study of Atherosclerosis (MESA)}.
\newblock \bibinfo{journal}{\emph{Sleep}} \bibinfo{volume}{38}, \bibinfo{number}{6} (\bibinfo{year}{2015}), \bibinfo{pages}{877--888}.
\newblock


\bibitem[Chen et~al\mbox{.}(2020)]%
        {chen2020deep}
\bibfield{author}{\bibinfo{person}{Yanming Chen}, \bibinfo{person}{Chao Li}, \bibinfo{person}{Luqi Gong}, \bibinfo{person}{Xiang Wen}, \bibinfo{person}{Yiwen Zhang}, {and} \bibinfo{person}{Weisong Shi}.} \bibinfo{year}{2020}\natexlab{}.
\newblock \showarticletitle{A deep neural network compression algorithm based on knowledge transfer for edge devices}.
\newblock \bibinfo{journal}{\emph{Computer Communications}}  \bibinfo{volume}{163} (\bibinfo{year}{2020}), \bibinfo{pages}{186--194}.
\newblock


\bibitem[Choi et~al\mbox{.}(2022)]%
        {choi2022attention}
\bibfield{author}{\bibinfo{person}{Jiho Choi}, \bibinfo{person}{Jun~Seong Lee}, \bibinfo{person}{Moonwook Ryu}, \bibinfo{person}{Gyutae Hwang}, \bibinfo{person}{Gyeongyeon Hwang}, {and} \bibinfo{person}{Sang~Jun Lee}.} \bibinfo{year}{2022}\natexlab{}.
\newblock \showarticletitle{Attention-lrcn: long-term recurrent convolutional network for stress detection from photoplethysmography}. In \bibinfo{booktitle}{\emph{2022 IEEE International Symposium on Medical Measurements and Applications (MeMeA)}}. IEEE, \bibinfo{pages}{1--6}.
\newblock


\bibitem[Cubas and Prado(2023)]%
        {cubas2023design}
\bibfield{author}{\bibinfo{person}{Percy Cubas} {and} \bibinfo{person}{Sixto Prado}.} \bibinfo{year}{2023}\natexlab{}.
\newblock \showarticletitle{Design of a PPG Signal Acquisition Platform Robust to Ambient Light}. In \bibinfo{booktitle}{\emph{Brazilian Technology Symposium}}. Springer, \bibinfo{pages}{206--216}.
\newblock


\bibitem[Das et~al\mbox{.}(2024)]%
        {das2024decoder}
\bibfield{author}{\bibinfo{person}{Abhimanyu Das}, \bibinfo{person}{Weihao Kong}, \bibinfo{person}{Rajat Sen}, {and} \bibinfo{person}{Yichen Zhou}.} \bibinfo{year}{2024}\natexlab{}.
\newblock \showarticletitle{A decoder-only foundation model for time-series forecasting}. In \bibinfo{booktitle}{\emph{Forty-first International Conference on Machine Learning}}.
\newblock


\bibitem[Davies et~al\mbox{.}(2024)]%
        {davies2024interpretable}
\bibfield{author}{\bibinfo{person}{Harry~J Davies}, \bibinfo{person}{James Monsen}, {and} \bibinfo{person}{Danilo~P Mandic}.} \bibinfo{year}{2024}\natexlab{}.
\newblock \showarticletitle{Interpretable Pre-Trained Transformers for Heart Time-Series Data}.
\newblock \bibinfo{journal}{\emph{arXiv preprint arXiv:2407.20775}} (\bibinfo{year}{2024}).
\newblock


\bibitem[Devlin et~al\mbox{.}(2018)]%
        {devlin2018bert}
\bibfield{author}{\bibinfo{person}{Jacob Devlin}, \bibinfo{person}{Ming-Wei Chang}, \bibinfo{person}{Kenton Lee}, {and} \bibinfo{person}{Kristina Toutanova}.} \bibinfo{year}{2018}\natexlab{}.
\newblock \showarticletitle{Bert: Pre-training of deep bidirectional transformers for language understanding}.
\newblock \bibinfo{journal}{\emph{arXiv preprint arXiv:1810.04805}} (\bibinfo{year}{2018}).
\newblock


\bibitem[Ding et~al\mbox{.}(2024)]%
        {ding2024siamquality}
\bibfield{author}{\bibinfo{person}{Cheng Ding}, \bibinfo{person}{Zhicheng Guo}, \bibinfo{person}{Zhaoliang Chen}, \bibinfo{person}{Randall~J Lee}, \bibinfo{person}{Cynthia Rudin}, {and} \bibinfo{person}{Xiao Hu}.} \bibinfo{year}{2024}\natexlab{}.
\newblock \showarticletitle{SiamQuality: a ConvNet-based foundation model for photoplethysmography signals}.
\newblock \bibinfo{journal}{\emph{Physiological Measurement}} \bibinfo{volume}{45}, \bibinfo{number}{8} (\bibinfo{year}{2024}), \bibinfo{pages}{085004}.
\newblock


\bibitem[Eldele et~al\mbox{.}(2023)]%
        {eldele2023tstcc}
\bibfield{author}{\bibinfo{person}{Emadeldeen Eldele}, \bibinfo{person}{Mohamed Ragab}, \bibinfo{person}{Zhenghua Chen}, \bibinfo{person}{Min Wu}, \bibinfo{person}{Chee-Keong Kwoh}, \bibinfo{person}{Xiaoli Li}, {and} \bibinfo{person}{Cuntai Guan}.} \bibinfo{year}{2023}\natexlab{}.
\newblock \showarticletitle{Self-supervised contrastive representation learning for semi-supervised time-series classification}.
\newblock \bibinfo{journal}{\emph{IEEE Transactions on Pattern Analysis and Machine Intelligence}} (\bibinfo{year}{2023}).
\newblock


\bibitem[Elgendi(2012)]%
        {elgendi2012analysis}
\bibfield{author}{\bibinfo{person}{Mohamed Elgendi}.} \bibinfo{year}{2012}\natexlab{}.
\newblock \showarticletitle{On the analysis of fingertip photoplethysmogram signals}.
\newblock \bibinfo{journal}{\emph{Current cardiology reviews}} \bibinfo{volume}{8}, \bibinfo{number}{1} (\bibinfo{year}{2012}), \bibinfo{pages}{14--25}.
\newblock


\bibitem[Elgendi et~al\mbox{.}(2019)]%
        {elgendi2019use}
\bibfield{author}{\bibinfo{person}{Mohamed Elgendi}, \bibinfo{person}{Richard Fletcher}, \bibinfo{person}{Yongbo Liang}, \bibinfo{person}{Newton Howard}, \bibinfo{person}{Nigel~H Lovell}, \bibinfo{person}{Derek Abbott}, \bibinfo{person}{Kenneth Lim}, {and} \bibinfo{person}{Rabab Ward}.} \bibinfo{year}{2019}\natexlab{}.
\newblock \showarticletitle{The use of photoplethysmography for assessing hypertension}.
\newblock \bibinfo{journal}{\emph{NPJ digital medicine}} \bibinfo{volume}{2}, \bibinfo{number}{1} (\bibinfo{year}{2019}), \bibinfo{pages}{60}.
\newblock


\bibitem[Elgendi et~al\mbox{.}(2022)]%
        {elgendi2022datasetppgbp}
\bibfield{author}{\bibinfo{person}{Mohamed Elgendi}, \bibinfo{person}{Valeria Galli}, \bibinfo{person}{Chakaveh Ahmadizadeh}, {and} \bibinfo{person}{Carlo Menon}.} \bibinfo{year}{2022}\natexlab{}.
\newblock \showarticletitle{Dataset of psychological scales and physiological signals collected for anxiety assessment using a portable device}.
\newblock \bibinfo{journal}{\emph{Data}} \bibinfo{volume}{7}, \bibinfo{number}{9} (\bibinfo{year}{2022}), \bibinfo{pages}{132}.
\newblock


\bibitem[Ertin et~al\mbox{.}(2011)]%
        {ertin2011autosense}
\bibfield{author}{\bibinfo{person}{Emre Ertin}, \bibinfo{person}{Nathan Stohs}, \bibinfo{person}{Santosh Kumar}, \bibinfo{person}{Andrew Raij}, \bibinfo{person}{Mustafa Al'Absi}, {and} \bibinfo{person}{Siddharth Shah}.} \bibinfo{year}{2011}\natexlab{}.
\newblock \showarticletitle{AutoSense: unobtrusively wearable sensor suite for inferring the onset, causality, and consequences of stress in the field}. In \bibinfo{booktitle}{\emph{Proceedings of the 9th ACM Conference on Embedded Networked Sensor Systems}}. \bibinfo{pages}{274--287}.
\newblock


\bibitem[Fischer et~al\mbox{.}(2016)]%
        {fischer2016algorithm}
\bibfield{author}{\bibinfo{person}{Christoph Fischer}, \bibinfo{person}{Benno D{\"o}mer}, \bibinfo{person}{Thomas Wibmer}, {and} \bibinfo{person}{Thomas Penzel}.} \bibinfo{year}{2016}\natexlab{}.
\newblock \showarticletitle{An algorithm for real-time pulse waveform segmentation and artifact detection in photoplethysmograms}.
\newblock \bibinfo{journal}{\emph{IEEE journal of biomedical and health informatics}} \bibinfo{volume}{21}, \bibinfo{number}{2} (\bibinfo{year}{2016}), \bibinfo{pages}{372--381}.
\newblock


\bibitem[Fitria(2023)]%
        {fitria2023artificial}
\bibfield{author}{\bibinfo{person}{Tira~Nur Fitria}.} \bibinfo{year}{2023}\natexlab{}.
\newblock \showarticletitle{Artificial intelligence (AI) technology in OpenAI ChatGPT application: A review of ChatGPT in writing English essay}. In \bibinfo{booktitle}{\emph{ELT Forum: Journal of English Language Teaching}}, Vol.~\bibinfo{volume}{12}. \bibinfo{pages}{44--58}.
\newblock


\bibitem[Fortino and Giamp{\`a}(2010)]%
        {fortino2010ppg}
\bibfield{author}{\bibinfo{person}{Giancarlo Fortino} {and} \bibinfo{person}{Valerio Giamp{\`a}}.} \bibinfo{year}{2010}\natexlab{}.
\newblock \showarticletitle{PPG-based methods for non invasive and continuous blood pressure measurement: An overview and development issues in body sensor networks}. In \bibinfo{booktitle}{\emph{2010 IEEE International Workshop on Medical Measurements and Applications}}. IEEE, \bibinfo{pages}{10--13}.
\newblock


\bibitem[Garde et~al\mbox{.}(2014)]%
        {garde2014development}
\bibfield{author}{\bibinfo{person}{Ainara Garde}, \bibinfo{person}{Parastoo Dehkordi}, \bibinfo{person}{Walter Karlen}, \bibinfo{person}{David Wensley}, \bibinfo{person}{J~Mark Ansermino}, {and} \bibinfo{person}{Guy~A Dumont}.} \bibinfo{year}{2014}\natexlab{}.
\newblock \showarticletitle{Development of a screening tool for sleep disordered breathing in children using the phone Oximeter™}.
\newblock \bibinfo{journal}{\emph{PloS one}} \bibinfo{volume}{9}, \bibinfo{number}{11} (\bibinfo{year}{2014}), \bibinfo{pages}{e112959}.
\newblock


\bibitem[Goswami et~al\mbox{.}(2024)]%
        {goswami2024moment}
\bibfield{author}{\bibinfo{person}{Mononito Goswami}, \bibinfo{person}{Konrad Szafer}, \bibinfo{person}{Arjun Choudhry}, \bibinfo{person}{Yifu Cai}, \bibinfo{person}{Shuo Li}, {and} \bibinfo{person}{Artur Dubrawski}.} \bibinfo{year}{2024}\natexlab{}.
\newblock \showarticletitle{Moment: A family of open time-series foundation models}.
\newblock \bibinfo{journal}{\emph{arXiv preprint arXiv:2402.03885}} (\bibinfo{year}{2024}).
\newblock


\bibitem[Haddad et~al\mbox{.}(2020)]%
        {haddad2020beat}
\bibfield{author}{\bibinfo{person}{Serj Haddad}, \bibinfo{person}{Assim Boukhayma}, {and} \bibinfo{person}{Antonino Caizzone}.} \bibinfo{year}{2020}\natexlab{}.
\newblock \showarticletitle{Beat-to-beat detection accuracy using the ultra low power senbiosys PPG sensor}. In \bibinfo{booktitle}{\emph{European Medical and Biological Engineering Conference}}. Springer, \bibinfo{pages}{178--188}.
\newblock


\bibitem[Haresamudram et~al\mbox{.}(2022)]%
        {haresamudram2022assessing}
\bibfield{author}{\bibinfo{person}{Harish Haresamudram}, \bibinfo{person}{Irfan Essa}, {and} \bibinfo{person}{Thomas Pl\"{o}tz}.} \bibinfo{year}{2022}\natexlab{}.
\newblock \showarticletitle{Assessing the State of Self-Supervised Human Activity Recognition Using Wearables}.
\newblock \bibinfo{journal}{\emph{Proc. ACM Interact. Mob. Wearable Ubiquitous Technol.}} \bibinfo{volume}{6}, \bibinfo{number}{3}, Article \bibinfo{articleno}{116} (\bibinfo{date}{sep} \bibinfo{year}{2022}), \bibinfo{numpages}{47}~pages.
\newblock
\urldef\tempurl%
\url{https://doi.org/10.1145/3550299}
\showDOI{\tempurl}


\bibitem[Hasanpoor et~al\mbox{.}(2022)]%
        {hasanpoor2022stress}
\bibfield{author}{\bibinfo{person}{Yasin Hasanpoor}, \bibinfo{person}{Bahram Tarvirdizadeh}, \bibinfo{person}{Khalil Alipour}, {and} \bibinfo{person}{Mohammad Ghamari}.} \bibinfo{year}{2022}\natexlab{}.
\newblock \showarticletitle{Stress Assessment with Convolutional Neural Network Using PPG Signals}. In \bibinfo{booktitle}{\emph{2022 10th RSI International Conference on Robotics and Mechatronics (ICRoM)}}. IEEE, \bibinfo{pages}{472--477}.
\newblock


\bibitem[He et~al\mbox{.}(2022)]%
        {he2022new}
\bibfield{author}{\bibinfo{person}{Jiayu He}, \bibinfo{person}{Jianlin Ou}, \bibinfo{person}{An He}, \bibinfo{person}{Lin Shu}, \bibinfo{person}{Tao Liu}, \bibinfo{person}{Ruowen Qu}, \bibinfo{person}{Xiangmin Xu}, \bibinfo{person}{Zhuoming Chen}, {and} \bibinfo{person}{Yifeng Yan}.} \bibinfo{year}{2022}\natexlab{}.
\newblock \showarticletitle{A new approach for daily life Blood-Pressure estimation using smart watch}.
\newblock \bibinfo{journal}{\emph{Biomedical Signal Processing and Control}}  \bibinfo{volume}{75} (\bibinfo{year}{2022}), \bibinfo{pages}{103616}.
\newblock


\bibitem[He et~al\mbox{.}(2023)]%
        {he2023compression}
\bibfield{author}{\bibinfo{person}{Peng He}, \bibinfo{person}{Shaoming Meng}, \bibinfo{person}{Yaping Cui}, \bibinfo{person}{Dapeng Wu}, {and} \bibinfo{person}{Ruyan Wang}.} \bibinfo{year}{2023}\natexlab{}.
\newblock \showarticletitle{Compression and encryption of heterogeneous signals for internet of medical things}.
\newblock \bibinfo{journal}{\emph{IEEE Journal of Biomedical and Health Informatics}} (\bibinfo{year}{2023}).
\newblock


\bibitem[Hoffmann et~al\mbox{.}(2022)]%
        {hoffmann2022metricrince}
\bibfield{author}{\bibinfo{person}{David~T Hoffmann}, \bibinfo{person}{Nadine Behrmann}, \bibinfo{person}{Juergen Gall}, \bibinfo{person}{Thomas Brox}, {and} \bibinfo{person}{Mehdi Noroozi}.} \bibinfo{year}{2022}\natexlab{}.
\newblock \showarticletitle{Ranking info noise contrastive estimation: Boosting contrastive learning via ranked positives}. In \bibinfo{booktitle}{\emph{Proceedings of the AAAI Conference on Artificial Intelligence}}, Vol.~\bibinfo{volume}{36}. \bibinfo{pages}{897--905}.
\newblock


\bibitem[Hong et~al\mbox{.}(2024)]%
        {hong2024spectralgpt}
\bibfield{author}{\bibinfo{person}{Danfeng Hong}, \bibinfo{person}{Bing Zhang}, \bibinfo{person}{Xuyang Li}, \bibinfo{person}{Yuxuan Li}, \bibinfo{person}{Chenyu Li}, \bibinfo{person}{Jing Yao}, \bibinfo{person}{Naoto Yokoya}, \bibinfo{person}{Hao Li}, \bibinfo{person}{Pedram Ghamisi}, \bibinfo{person}{Xiuping Jia}, {et~al\mbox{.}}} \bibinfo{year}{2024}\natexlab{}.
\newblock \showarticletitle{SpectralGPT: Spectral remote sensing foundation model}.
\newblock \bibinfo{journal}{\emph{IEEE Transactions on Pattern Analysis and Machine Intelligence}} (\bibinfo{year}{2024}).
\newblock


\bibitem[Hovsepian et~al\mbox{.}(2015)]%
        {Hovsepian-2015-cStress}
\bibfield{author}{\bibinfo{person}{Karen Hovsepian}, \bibinfo{person}{Mustafa Al'Absi}, \bibinfo{person}{Emre Ertin}, \bibinfo{person}{Thomas Kamarck}, \bibinfo{person}{Motohiro Nakajima}, {and} \bibinfo{person}{Santosh Kumar}.} \bibinfo{year}{2015}\natexlab{}.
\newblock \showarticletitle{cStress: towards a gold standard for continuous stress assessment in the mobile environment}.
\newblock \bibinfo{journal}{\emph{Proceedings of the 2015 ACM International Joint Conference on Pervasive and Ubiquitous Computing}} (\bibinfo{year}{2015}).
\newblock


\bibitem[Hu et~al\mbox{.}(2022)]%
        {hu2022effect}
\bibfield{author}{\bibinfo{person}{Hangxing Hu}, \bibinfo{person}{Jin Li}, {and} \bibinfo{person}{Xiang Chen}.} \bibinfo{year}{2022}\natexlab{}.
\newblock \showarticletitle{The effect of skin melanin concentration on wrist reflectance photoplethysmography based on Monte Carlo simulation}. In \bibinfo{booktitle}{\emph{2022 15th International Congress on Image and Signal Processing, BioMedical Engineering and Informatics (CISP-BMEI)}}. IEEE, \bibinfo{pages}{1--6}.
\newblock


\bibitem[Jeong et~al\mbox{.}(2023)]%
        {jeong2023event}
\bibfield{author}{\bibinfo{person}{Hyewon Jeong}, \bibinfo{person}{Nassim Oufattole}, \bibinfo{person}{Aparna Balagopalan}, \bibinfo{person}{Matthew Mcdermott}, \bibinfo{person}{Payal Chandak}, \bibinfo{person}{Marzyeh Ghassemi}, {and} \bibinfo{person}{Collin Stultz}.} \bibinfo{year}{2023}\natexlab{}.
\newblock \showarticletitle{Event-Based Contrastive Learning for Medical Time Series}.
\newblock \bibinfo{journal}{\emph{arXiv preprint arXiv:2312.10308}} (\bibinfo{year}{2023}).
\newblock


\bibitem[Jeong et~al\mbox{.}(2024)]%
        {jeong2024finding}
\bibfield{author}{\bibinfo{person}{Hyewon Jeong}, \bibinfo{person}{Suyeol Yun}, {and} \bibinfo{person}{Hammaad Adam}.} \bibinfo{year}{2024}\natexlab{}.
\newblock \showarticletitle{Finding" Good Views" of Electrocardiogram Signals for Inferring Abnormalities in Cardiac Condition}.
\newblock \bibinfo{journal}{\emph{arXiv preprint arXiv:2411.17702}} (\bibinfo{year}{2024}).
\newblock


\bibitem[Kan et~al\mbox{.}(2021)]%
        {kan2021metricroc}
\bibfield{author}{\bibinfo{person}{Shichao Kan}, \bibinfo{person}{Yigang Cen}, \bibinfo{person}{Yang Li}, \bibinfo{person}{Vladimir Mladenovic}, {and} \bibinfo{person}{Zhihai He}.} \bibinfo{year}{2021}\natexlab{}.
\newblock \showarticletitle{Relative order analysis and optimization for unsupervised deep metric learning}. In \bibinfo{booktitle}{\emph{Proceedings of the IEEE/CVF conference on computer vision and pattern recognition}}. \bibinfo{pages}{13999--14008}.
\newblock


\bibitem[Karlen et~al\mbox{.}(2011)]%
        {karlen2011human}
\bibfield{author}{\bibinfo{person}{Walter Karlen}, \bibinfo{person}{Guy Dumont}, \bibinfo{person}{Chris Petersen}, \bibinfo{person}{Jennifer Gow}, \bibinfo{person}{Joanne Lim}, \bibinfo{person}{Jules Sleiman}, {and} \bibinfo{person}{J~Mark Ansermino}.} \bibinfo{year}{2011}\natexlab{}.
\newblock \showarticletitle{HUMAN-CENTERED PHONE OXIMETER INTERFACE DESIGN FOR THE OPERATING ROOM-Pulse Oximeter Interfaced to a Mobile Device for Anesthesia Monitoring in the Developing World}. In \bibinfo{booktitle}{\emph{International Conference on Health Informatics}}, Vol.~\bibinfo{volume}{2}. SciTePress, \bibinfo{pages}{433--438}.
\newblock


\bibitem[Kavsao{\u{g}}lu et~al\mbox{.}(2016)]%
        {kavsaouglu2016innovative}
\bibfield{author}{\bibinfo{person}{Ahmet~Re{\c{s}}it Kavsao{\u{g}}lu}, \bibinfo{person}{Kemal Polat}, {and} \bibinfo{person}{Mehmet~Recep Bozkurt}.} \bibinfo{year}{2016}\natexlab{}.
\newblock \showarticletitle{An innovative peak detection algorithm for photoplethysmography signals: an adaptive segmentation method}.
\newblock \bibinfo{journal}{\emph{Turkish Journal of Electrical Engineering and Computer Sciences}} \bibinfo{volume}{24}, \bibinfo{number}{3} (\bibinfo{year}{2016}), \bibinfo{pages}{1782--1796}.
\newblock


\bibitem[Kim et~al\mbox{.}(2019)]%
        {kim2019metricbeyondlogratio}
\bibfield{author}{\bibinfo{person}{Sungyeon Kim}, \bibinfo{person}{Minkyo Seo}, \bibinfo{person}{Ivan Laptev}, \bibinfo{person}{Minsu Cho}, {and} \bibinfo{person}{Suha Kwak}.} \bibinfo{year}{2019}\natexlab{}.
\newblock \showarticletitle{Deep metric learning beyond binary supervision}. In \bibinfo{booktitle}{\emph{Proceedings of the IEEE/CVF Conference on Computer Vision and Pattern Recognition}}. \bibinfo{pages}{2288--2297}.
\newblock


\bibitem[Kirillov et~al\mbox{.}(2023)]%
        {kirillov2023segment}
\bibfield{author}{\bibinfo{person}{Alexander Kirillov}, \bibinfo{person}{Eric Mintun}, \bibinfo{person}{Nikhila Ravi}, \bibinfo{person}{Hanzi Mao}, \bibinfo{person}{Chloe Rolland}, \bibinfo{person}{Laura Gustafson}, \bibinfo{person}{Tete Xiao}, \bibinfo{person}{Spencer Whitehead}, \bibinfo{person}{Alexander~C Berg}, \bibinfo{person}{Wan-Yen Lo}, {et~al\mbox{.}}} \bibinfo{year}{2023}\natexlab{}.
\newblock \showarticletitle{Segment anything}. In \bibinfo{booktitle}{\emph{Proceedings of the IEEE/CVF International Conference on Computer Vision}}. \bibinfo{pages}{4015--4026}.
\newblock


\bibitem[Kontaxis et~al\mbox{.}(2020)]%
        {kontaxis2020photoplethysmographic}
\bibfield{author}{\bibinfo{person}{Spyridon Kontaxis}, \bibinfo{person}{Eduardo Gil}, \bibinfo{person}{Vaidotas Marozas}, \bibinfo{person}{Jesus Lazaro}, \bibinfo{person}{Esther Garcia}, \bibinfo{person}{Mar Posadas-de Miguel}, \bibinfo{person}{Sara Siddi}, \bibinfo{person}{Maria~Luisa Bernal}, \bibinfo{person}{Jordi Aguilo}, \bibinfo{person}{Josep~Maria Haro}, {et~al\mbox{.}}} \bibinfo{year}{2020}\natexlab{}.
\newblock \showarticletitle{Photoplethysmographic waveform analysis for autonomic reactivity assessment in depression}.
\newblock \bibinfo{journal}{\emph{IEEE Transactions on Biomedical Engineering}} \bibinfo{volume}{68}, \bibinfo{number}{4} (\bibinfo{year}{2020}), \bibinfo{pages}{1273--1281}.
\newblock


\bibitem[Kulkarni et~al\mbox{.}(2021)]%
        {kulkarni2021ai}
\bibfield{author}{\bibinfo{person}{Uday Kulkarni}, \bibinfo{person}{SM Meena}, \bibinfo{person}{Sunil~V Gurlahosur}, \bibinfo{person}{Pratiksha Benagi}, \bibinfo{person}{Atul Kashyap}, \bibinfo{person}{Ayub Ansari}, {and} \bibinfo{person}{Vinay Karnam}.} \bibinfo{year}{2021}\natexlab{}.
\newblock \showarticletitle{AI model compression for edge devices using optimization techniques}.
\newblock In \bibinfo{booktitle}{\emph{Modern Approaches in Machine Learning and Cognitive Science: A Walkthrough: Latest Trends in AI, Volume 2}}. \bibinfo{publisher}{Springer}, \bibinfo{pages}{227--240}.
\newblock


\bibitem[Lamichhane et~al\mbox{.}(2017)]%
        {lamichhane2017towards}
\bibfield{author}{\bibinfo{person}{Bishal Lamichhane}, \bibinfo{person}{Ulf Gro{\ss}ekath{\"o}fer}, \bibinfo{person}{Giuseppina Schiavone}, {and} \bibinfo{person}{Pierluigi Casale}.} \bibinfo{year}{2017}\natexlab{}.
\newblock \showarticletitle{Towards stress detection in real-life scenarios using wearable sensors: normalization factor to reduce variability in stress physiology}. In \bibinfo{booktitle}{\emph{eHealth 360°: International Summit on eHealth, Budapest, Hungary, June 14-16, 2016, Revised Selected Papers}}. Springer, \bibinfo{pages}{259--270}.
\newblock


\bibitem[Lazazzera et~al\mbox{.}(2020)]%
        {lazazzera2020detection}
\bibfield{author}{\bibinfo{person}{Remo Lazazzera}, \bibinfo{person}{Margot Deviaene}, \bibinfo{person}{Carolina Varon}, \bibinfo{person}{Bertien Buyse}, \bibinfo{person}{Dries Testelmans}, \bibinfo{person}{Pablo Laguna}, \bibinfo{person}{Eduardo Gil}, {and} \bibinfo{person}{Guy Carrault}.} \bibinfo{year}{2020}\natexlab{}.
\newblock \showarticletitle{Detection and classification of sleep apnea and hypopnea using PPG and SpO $ \_2 $ signals}.
\newblock \bibinfo{journal}{\emph{IEEE Transactions on Biomedical Engineering}} \bibinfo{volume}{68}, \bibinfo{number}{5} (\bibinfo{year}{2020}), \bibinfo{pages}{1496--1506}.
\newblock


\bibitem[Lee et~al\mbox{.}(2022)]%
        {lee2022SSLAPP}
\bibfield{author}{\bibinfo{person}{Harim Lee}, \bibinfo{person}{Eunseon Seong}, {and} \bibinfo{person}{Dong-Kyu Chae}.} \bibinfo{year}{2022}\natexlab{}.
\newblock \showarticletitle{Self-supervised learning with attention-based latent signal augmentation for sleep staging with limited labeled data}. In \bibinfo{booktitle}{\emph{Proceedings of the Thirty-First International Joint Conference on Artificial Intelligence, IJCAI-22, LD Raedt, Ed. International Joint Conferences on Artificial Intelligence Organization}}, Vol.~\bibinfo{volume}{7}. \bibinfo{pages}{3868--3876}.
\newblock


\bibitem[Li et~al\mbox{.}(2024)]%
        {li2024large}
\bibfield{author}{\bibinfo{person}{Xiang Li}, \bibinfo{person}{Zhenyan Lu}, \bibinfo{person}{Dongqi Cai}, \bibinfo{person}{Xiao Ma}, {and} \bibinfo{person}{Mengwei Xu}.} \bibinfo{year}{2024}\natexlab{}.
\newblock \showarticletitle{Large language models on mobile devices: Measurements, analysis, and insights}. In \bibinfo{booktitle}{\emph{Proceedings of the Workshop on Edge and Mobile Foundation Models}}. \bibinfo{pages}{1--6}.
\newblock


\bibitem[Li et~al\mbox{.}(2019)]%
        {li2019dice}
\bibfield{author}{\bibinfo{person}{Xiaoya Li}, \bibinfo{person}{Xiaofei Sun}, \bibinfo{person}{Yuxian Meng}, \bibinfo{person}{Junjun Liang}, \bibinfo{person}{Fei Wu}, {and} \bibinfo{person}{Jiwei Li}.} \bibinfo{year}{2019}\natexlab{}.
\newblock \showarticletitle{Dice loss for data-imbalanced NLP tasks}.
\newblock \bibinfo{journal}{\emph{arXiv preprint arXiv:1911.02855}} (\bibinfo{year}{2019}).
\newblock


\bibitem[Liang et~al\mbox{.}(2018)]%
        {liang2018new}
\bibfield{author}{\bibinfo{person}{Yongbo Liang}, \bibinfo{person}{Zhencheng Chen}, \bibinfo{person}{Guiyong Liu}, {and} \bibinfo{person}{Mohamed Elgendi}.} \bibinfo{year}{2018}\natexlab{}.
\newblock \showarticletitle{A new, short-recorded photoplethysmogram dataset for blood pressure monitoring in China}.
\newblock \bibinfo{journal}{\emph{Scientific data}} \bibinfo{volume}{5}, \bibinfo{number}{1} (\bibinfo{year}{2018}), \bibinfo{pages}{1--7}.
\newblock


\bibitem[Lin et~al\mbox{.}(2025)]%
        {lin2025longitudinal}
\bibfield{author}{\bibinfo{person}{Hui Lin}, \bibinfo{person}{Jiyang Li}, \bibinfo{person}{Ramy Hussein}, \bibinfo{person}{Xin Sui}, \bibinfo{person}{Xiaoyu Li}, \bibinfo{person}{Guangpu Zhu}, \bibinfo{person}{Aggelos~K Katsaggelos}, \bibinfo{person}{Zijing Zeng}, {and} \bibinfo{person}{Yelei Li}.} \bibinfo{year}{2025}\natexlab{}.
\newblock \showarticletitle{Longitudinal Wrist PPG Analysis for Reliable Hypertension Risk Screening Using Deep Learning}. In \bibinfo{booktitle}{\emph{ICASSP 2025-2025 IEEE International Conference on Acoustics, Speech and Signal Processing (ICASSP)}}. IEEE, \bibinfo{pages}{1--5}.
\newblock


\bibitem[Liu et~al\mbox{.}(2018)]%
        {liu2018partialconv}
\bibfield{author}{\bibinfo{person}{Guilin Liu}, \bibinfo{person}{Fitsum~A Reda}, \bibinfo{person}{Kevin~J Shih}, \bibinfo{person}{Ting-Chun Wang}, \bibinfo{person}{Andrew Tao}, {and} \bibinfo{person}{Bryan Catanzaro}.} \bibinfo{year}{2018}\natexlab{}.
\newblock \showarticletitle{Image inpainting for irregular holes using partial convolutions}. In \bibinfo{booktitle}{\emph{Proceedings of the European conference on computer vision (ECCV)}}. \bibinfo{pages}{85--100}.
\newblock


\bibitem[Lo~Grasso et~al\mbox{.}(2024)]%
        {lo2024advanced}
\bibfield{author}{\bibinfo{person}{Anna Lo~Grasso}, \bibinfo{person}{Pamela Zontone}, \bibinfo{person}{Roberto Rinaldo}, {and} \bibinfo{person}{Antonio Affanni}.} \bibinfo{year}{2024}\natexlab{}.
\newblock \showarticletitle{Advanced Necklace for Real-Time PPG Monitoring in Drivers}.
\newblock \bibinfo{journal}{\emph{Sensors}} \bibinfo{volume}{24}, \bibinfo{number}{18} (\bibinfo{year}{2024}), \bibinfo{pages}{5908}.
\newblock


\bibitem[McCarthy et~al\mbox{.}(2016)]%
        {mccarthy2016validation}
\bibfield{author}{\bibinfo{person}{Cameron McCarthy}, \bibinfo{person}{Nikhilesh Pradhan}, \bibinfo{person}{Calum Redpath}, {and} \bibinfo{person}{Andy Adler}.} \bibinfo{year}{2016}\natexlab{}.
\newblock \showarticletitle{Validation of the Empatica E4 wristband}. In \bibinfo{booktitle}{\emph{2016 IEEE EMBS international student conference (ISC)}}. IEEE, \bibinfo{pages}{1--4}.
\newblock


\bibitem[Meng et~al\mbox{.}(2023)]%
        {meng2023unsupervised}
\bibfield{author}{\bibinfo{person}{Qianwen Meng}, \bibinfo{person}{Hangwei Qian}, \bibinfo{person}{Yong Liu}, \bibinfo{person}{Yonghui Xu}, \bibinfo{person}{Zhiqi Shen}, {and} \bibinfo{person}{Lizhen Cui}.} \bibinfo{year}{2023}\natexlab{}.
\newblock \showarticletitle{Unsupervised representation learning for time series: A review}.
\newblock \bibinfo{journal}{\emph{arXiv preprint arXiv:2308.01578}} (\bibinfo{year}{2023}).
\newblock


\bibitem[Miller(2025)]%
        {miller2025endosleep}
\bibfield{author}{\bibinfo{person}{Steven Miller}.} \bibinfo{year}{2025}\natexlab{}.
\newblock \bibinfo{title}{Latest ENSODATA FDA 510(k) clearance enables AI-powered sleep diagnosis using pulse oximetry devices}.
\newblock
\newblock
\urldef\tempurl%
\url{https://www.ensodata.com/press/latest-ensodata-fda-510k-clearance-enables-ai-powered-sleep-diagnosis-using-pulse-oximetry-devices/}
\showURL{%
\tempurl}


\bibitem[Mishra et~al\mbox{.}(2018)]%
        {mishra2018investigating}
\bibfield{author}{\bibinfo{person}{Varun Mishra}, \bibinfo{person}{Tian Hao}, \bibinfo{person}{Si Sun}, \bibinfo{person}{Kimberly~N Walter}, \bibinfo{person}{Marion~J Ball}, \bibinfo{person}{Ching-Hua Chen}, {and} \bibinfo{person}{Xinxin Zhu}.} \bibinfo{year}{2018}\natexlab{}.
\newblock \showarticletitle{Investigating the role of context in perceived stress detection in the wild}. In \bibinfo{booktitle}{\emph{Proceedings of the 2018 ACM International Joint Conference and 2018 International Symposium on Pervasive and Ubiquitous Computing and Wearable Computers}}. \bibinfo{pages}{1708--1716}.
\newblock


\bibitem[Mitro et~al\mbox{.}(2023)]%
        {mitro2023ai}
\bibfield{author}{\bibinfo{person}{Nikos Mitro}, \bibinfo{person}{Katerina Argyri}, \bibinfo{person}{Lampros Pavlopoulos}, \bibinfo{person}{Dimitrios Kosyvas}, \bibinfo{person}{Lazaros Karagiannidis}, \bibinfo{person}{Margarita Kostovasili}, \bibinfo{person}{Fay Misichroni}, \bibinfo{person}{Eleftherios Ouzounoglou}, {and} \bibinfo{person}{Angelos Amditis}.} \bibinfo{year}{2023}\natexlab{}.
\newblock \showarticletitle{AI-Enabled Smart Wristband Providing Real-Time Vital Signs and Stress Monitoring}.
\newblock \bibinfo{journal}{\emph{Sensors}} \bibinfo{volume}{23}, \bibinfo{number}{5} (\bibinfo{year}{2023}), \bibinfo{pages}{2821}.
\newblock


\bibitem[Moreau et~al\mbox{.}(2018)]%
        {moreau2018leveraging}
\bibfield{author}{\bibinfo{person}{Thierry Moreau}, \bibinfo{person}{Tianqi Chen}, {and} \bibinfo{person}{Luis Ceze}.} \bibinfo{year}{2018}\natexlab{}.
\newblock \showarticletitle{Leveraging the vta-tvm hardware-software stack for fpga acceleration of 8-bit resnet-18 inference}.
\newblock In \bibinfo{booktitle}{\emph{Proceedings of the 1st on Reproducible Quality-Efficient Systems Tournament on Co-designing Pareto-efficient Deep Learning}}. \bibinfo{pages}{1}.
\newblock


\bibitem[Motaman et~al\mbox{.}(2022)]%
        {motaman2022stress}
\bibfield{author}{\bibinfo{person}{Koorosh Motaman}, \bibinfo{person}{Khalil Alipour}, \bibinfo{person}{Bahram Tarvirdizadeh}, {and} \bibinfo{person}{Mohammad Ghamari}.} \bibinfo{year}{2022}\natexlab{}.
\newblock \showarticletitle{A Stress Detection Model Based on LSTM Network Using Solely Raw PPG Signals}. In \bibinfo{booktitle}{\emph{2022 10th RSI International Conference on Robotics and Mechatronics (ICRoM)}}. IEEE, \bibinfo{pages}{485--490}.
\newblock


\bibitem[Neupane et~al\mbox{.}(2024)]%
        {neupane2024momentarymoods}
\bibfield{author}{\bibinfo{person}{Sameer Neupane}, \bibinfo{person}{Mithun Saha}, \bibinfo{person}{Nasir Ali}, \bibinfo{person}{Timothy Hnat}, \bibinfo{person}{Shahin~Alan Samiei}, \bibinfo{person}{Anandatirtha Nandugudi}, \bibinfo{person}{David~M Almeida}, {and} \bibinfo{person}{Santosh Kumar}.} \bibinfo{year}{2024}\natexlab{}.
\newblock \showarticletitle{Momentary Stressor Logging and Reflective Visualizations: Implications for Stress Management with Wearables}.
\newblock \bibinfo{journal}{\emph{arXiv preprint arXiv:2401.16307}} (\bibinfo{year}{2024}).
\newblock


\bibitem[Oquab et~al\mbox{.}(2023)]%
        {oquab2023dinov2}
\bibfield{author}{\bibinfo{person}{Maxime Oquab}, \bibinfo{person}{Timoth{\'e}e Darcet}, \bibinfo{person}{Th{\'e}o Moutakanni}, \bibinfo{person}{Huy Vo}, \bibinfo{person}{Marc Szafraniec}, \bibinfo{person}{Vasil Khalidov}, \bibinfo{person}{Pierre Fernandez}, \bibinfo{person}{Daniel Haziza}, \bibinfo{person}{Francisco Massa}, \bibinfo{person}{Alaaeldin El-Nouby}, {et~al\mbox{.}}} \bibinfo{year}{2023}\natexlab{}.
\newblock \showarticletitle{Dinov2: Learning robust visual features without supervision}.
\newblock \bibinfo{journal}{\emph{Transactions on Machine Learning Research}} (\bibinfo{year}{2023}).
\newblock


\bibitem[Ouyang et~al\mbox{.}(2017)]%
        {ouyang2017self}
\bibfield{author}{\bibinfo{person}{Han Ouyang}, \bibinfo{person}{Jingjing Tian}, \bibinfo{person}{Guanglong Sun}, \bibinfo{person}{Yang Zou}, \bibinfo{person}{Zhuo Liu}, \bibinfo{person}{Hu Li}, \bibinfo{person}{Luming Zhao}, \bibinfo{person}{Bojing Shi}, \bibinfo{person}{Yubo Fan}, \bibinfo{person}{Yifan Fan}, {et~al\mbox{.}}} \bibinfo{year}{2017}\natexlab{}.
\newblock \showarticletitle{Self-powered pulse sensor for antidiastole of cardiovascular disease}.
\newblock \bibinfo{journal}{\emph{Advanced Materials}} \bibinfo{volume}{29}, \bibinfo{number}{40} (\bibinfo{year}{2017}), \bibinfo{pages}{1703456}.
\newblock


\bibitem[Ozyurt et~al\mbox{.}(2022)]%
        {ozyurt2022cluda}
\bibfield{author}{\bibinfo{person}{Yilmazcan Ozyurt}, \bibinfo{person}{Stefan Feuerriegel}, {and} \bibinfo{person}{Ce Zhang}.} \bibinfo{year}{2022}\natexlab{}.
\newblock \showarticletitle{Contrastive learning for unsupervised domain adaptation of time series}.
\newblock \bibinfo{journal}{\emph{arXiv preprint arXiv:2206.06243}} (\bibinfo{year}{2022}).
\newblock


\bibitem[Pillai et~al\mbox{.}(2024)]%
        {pillai2024papagei}
\bibfield{author}{\bibinfo{person}{Arvind Pillai}, \bibinfo{person}{Dimitris Spathis}, \bibinfo{person}{Fahim Kawsar}, {and} \bibinfo{person}{Mohammad Malekzadeh}.} \bibinfo{year}{2024}\natexlab{}.
\newblock \showarticletitle{PaPaGei: Open Foundation Models for Optical Physiological Signals}.
\newblock \bibinfo{journal}{\emph{arXiv preprint arXiv:2410.20542}} (\bibinfo{year}{2024}).
\newblock


\bibitem[Plarre et~al\mbox{.}(2011)]%
        {plarre2011continuous}
\bibfield{author}{\bibinfo{person}{Kurt Plarre}, \bibinfo{person}{Andrew Raij}, \bibinfo{person}{Syed~Monowar Hossain}, \bibinfo{person}{Amin~Ahsan Ali}, \bibinfo{person}{Motohiro Nakajima}, \bibinfo{person}{Mustafa Al'Absi}, \bibinfo{person}{Emre Ertin}, \bibinfo{person}{Thomas Kamarck}, \bibinfo{person}{Santosh Kumar}, \bibinfo{person}{Marcia Scott}, {et~al\mbox{.}}} \bibinfo{year}{2011}\natexlab{}.
\newblock \showarticletitle{Continuous inference of psychological stress from sensory measurements collected in the natural environment}. In \bibinfo{booktitle}{\emph{Proceedings of the 10th ACM/IEEE international conference on information processing in sensor networks}}. IEEE, \bibinfo{pages}{97--108}.
\newblock


\bibitem[Poh et~al\mbox{.}(2018)]%
        {poh2018diagnostic}
\bibfield{author}{\bibinfo{person}{Ming-Zher Poh}, \bibinfo{person}{Yukkee~Cheung Poh}, \bibinfo{person}{Pak-Hei Chan}, \bibinfo{person}{Chun-Ka Wong}, \bibinfo{person}{Louise Pun}, \bibinfo{person}{Wangie Wan-Chiu Leung}, \bibinfo{person}{Yu-Fai Wong}, \bibinfo{person}{Michelle Man-Ying Wong}, \bibinfo{person}{Daniel Wai-Sing Chu}, {and} \bibinfo{person}{Chung-Wah Siu}.} \bibinfo{year}{2018}\natexlab{}.
\newblock \showarticletitle{Diagnostic assessment of a deep learning system for detecting atrial fibrillation in pulse waveforms}.
\newblock \bibinfo{journal}{\emph{Heart}} \bibinfo{volume}{104}, \bibinfo{number}{23} (\bibinfo{year}{2018}), \bibinfo{pages}{1921--1928}.
\newblock


\bibitem[Pollreisz and TaheriNejad(2022)]%
        {pollreisz2022detection}
\bibfield{author}{\bibinfo{person}{David Pollreisz} {and} \bibinfo{person}{Nima TaheriNejad}.} \bibinfo{year}{2022}\natexlab{}.
\newblock \showarticletitle{Detection and removal of motion artifacts in PPG signals}.
\newblock \bibinfo{journal}{\emph{Mobile Networks and Applications}} \bibinfo{volume}{27}, \bibinfo{number}{2} (\bibinfo{year}{2022}), \bibinfo{pages}{728--738}.
\newblock


\bibitem[P{\v{r}}ibil et~al\mbox{.}(2021)]%
        {pvribil2021wearable}
\bibfield{author}{\bibinfo{person}{Ji{\v{r}}{\'\i} P{\v{r}}ibil}, \bibinfo{person}{Anna P{\v{r}}ibilov{\'a}}, {and} \bibinfo{person}{Ivan Frollo}.} \bibinfo{year}{2021}\natexlab{}.
\newblock \showarticletitle{Wearable PPG sensor with bluetooth data transmission for continual measurement in low magnetic field environment}. In \bibinfo{booktitle}{\emph{2021 International Conference on Applied Electronics (AE)}}. IEEE, \bibinfo{pages}{1--4}.
\newblock


\bibitem[Puranen et~al\mbox{.}(2020)]%
        {puranen2020effect}
\bibfield{author}{\bibinfo{person}{Antti Puranen}, \bibinfo{person}{Tuomas Halkola}, \bibinfo{person}{Ole Kirkeby}, {and} \bibinfo{person}{Antti Vehkaoja}.} \bibinfo{year}{2020}\natexlab{}.
\newblock \showarticletitle{Effect of skin tone and activity on the performance of wrist-worn optical beat-to-beat heart rate monitoring}. In \bibinfo{booktitle}{\emph{2020 IEEE SENSORS}}. IEEE, \bibinfo{pages}{1--4}.
\newblock


\bibitem[Rabbani and Khan(2022)]%
        {rabbani2022contrastive}
\bibfield{author}{\bibinfo{person}{Suha Rabbani} {and} \bibinfo{person}{Naimul Khan}.} \bibinfo{year}{2022}\natexlab{}.
\newblock \showarticletitle{Contrastive self-supervised learning for stress detection from ecg data}.
\newblock \bibinfo{journal}{\emph{Bioengineering}} \bibinfo{volume}{9}, \bibinfo{number}{8} (\bibinfo{year}{2022}), \bibinfo{pages}{374}.
\newblock


\bibitem[Radford et~al\mbox{.}(2021)]%
        {radford2021learning}
\bibfield{author}{\bibinfo{person}{Alec Radford}, \bibinfo{person}{Jong~Wook Kim}, \bibinfo{person}{Chris Hallacy}, \bibinfo{person}{Aditya Ramesh}, \bibinfo{person}{Gabriel Goh}, \bibinfo{person}{Sandhini Agarwal}, \bibinfo{person}{Girish Sastry}, \bibinfo{person}{Amanda Askell}, \bibinfo{person}{Pamela Mishkin}, \bibinfo{person}{Jack Clark}, {et~al\mbox{.}}} \bibinfo{year}{2021}\natexlab{}.
\newblock \showarticletitle{Learning transferable visual models from natural language supervision}. In \bibinfo{booktitle}{\emph{International conference on machine learning}}. PMLR, \bibinfo{pages}{8748--8763}.
\newblock


\bibitem[Radford et~al\mbox{.}(2018)]%
        {radford2018improving}
\bibfield{author}{\bibinfo{person}{Alec Radford}, \bibinfo{person}{Karthik Narasimhan}, \bibinfo{person}{Tim Salimans}, \bibinfo{person}{Ilya Sutskever}, {et~al\mbox{.}}} \bibinfo{year}{2018}\natexlab{}.
\newblock \showarticletitle{Improving language understanding by generative pre-training}.
\newblock  (\bibinfo{year}{2018}).
\newblock


\bibitem[Rajakariar et~al\mbox{.}(2024)]%
        {rajakariar2024accuracy}
\bibfield{author}{\bibinfo{person}{Kevin Rajakariar}, \bibinfo{person}{Paul Buntine}, \bibinfo{person}{Andrew Ghaly}, \bibinfo{person}{Zheng~Cheng Zhu}, \bibinfo{person}{Vihangi Abeygunawardana}, \bibinfo{person}{Sarah Visakhamoorthy}, \bibinfo{person}{Patrick~J Owen}, \bibinfo{person}{Shaun Tham}, \bibinfo{person}{Liam Hackett}, \bibinfo{person}{Louise Roberts}, {et~al\mbox{.}}} \bibinfo{year}{2024}\natexlab{}.
\newblock \showarticletitle{Accuracy of Smartwatch Pulse Oximetry Measurements in Hospitalized Patients With Coronavirus Disease 2019}.
\newblock \bibinfo{journal}{\emph{Mayo Clinic Proceedings: Digital Health}} \bibinfo{volume}{2}, \bibinfo{number}{1} (\bibinfo{year}{2024}), \bibinfo{pages}{152--158}.
\newblock


\bibitem[Reiss et~al\mbox{.}(2019a)]%
        {reiss2019ppgdalia}
\bibfield{author}{\bibinfo{person}{Attila Reiss}, \bibinfo{person}{Ina Indlekofer}, \bibinfo{person}{Philip Schmidt}, {and} \bibinfo{person}{Kristof Van~Laerhoven}.} \bibinfo{year}{2019}\natexlab{a}.
\newblock \showarticletitle{Deep PPG: Large-scale heart rate estimation with convolutional neural networks}.
\newblock \bibinfo{journal}{\emph{Sensors}} \bibinfo{volume}{19}, \bibinfo{number}{14} (\bibinfo{year}{2019}), \bibinfo{pages}{3079}.
\newblock


\bibitem[Reiss et~al\mbox{.}(2019b)]%
        {reiss2019deep}
\bibfield{author}{\bibinfo{person}{Attila Reiss}, \bibinfo{person}{Ina Indlekofer}, \bibinfo{person}{Philip Schmidt}, {and} \bibinfo{person}{Kristof Van~Laerhoven}.} \bibinfo{year}{2019}\natexlab{b}.
\newblock \showarticletitle{Deep PPG: Large-scale heart rate estimation with convolutional neural networks}.
\newblock \bibinfo{journal}{\emph{Sensors}} \bibinfo{volume}{19}, \bibinfo{number}{14} (\bibinfo{year}{2019}), \bibinfo{pages}{3079}.
\newblock


\bibitem[Rinkevi{\v{c}}ius et~al\mbox{.}(2019)]%
        {rinkevivcius2019photoplethysmogram}
\bibfield{author}{\bibinfo{person}{Mantas Rinkevi{\v{c}}ius}, \bibinfo{person}{Spyridon Kontaxis}, \bibinfo{person}{Eduardo Gil}, \bibinfo{person}{Raquel Bail{\'o}n}, \bibinfo{person}{Jes{\'u}s L{\'a}zaro}, \bibinfo{person}{Pablo Laguna}, {and} \bibinfo{person}{Vaidotas Marozas}.} \bibinfo{year}{2019}\natexlab{}.
\newblock \showarticletitle{Photoplethysmogram signal morphology-based stress assessment}. In \bibinfo{booktitle}{\emph{2019 Computing in Cardiology (CinC)}}. IEEE, \bibinfo{pages}{Page--1}.
\newblock


\bibitem[Salah et~al\mbox{.}(2022)]%
        {salah2022beat}
\bibfield{author}{\bibinfo{person}{Mostafa Salah}, \bibinfo{person}{Osama~A Omer}, \bibinfo{person}{Loay Hassan}, \bibinfo{person}{Mohamed Ragab}, \bibinfo{person}{Ammar~Mostafa Hassan}, {and} \bibinfo{person}{Ahmed Abdelreheem}.} \bibinfo{year}{2022}\natexlab{}.
\newblock \showarticletitle{Beat-based PPG-ABP cleaning technique for blood pressure estimation}.
\newblock \bibinfo{journal}{\emph{IEEE Access}}  \bibinfo{volume}{10} (\bibinfo{year}{2022}), \bibinfo{pages}{55616--55626}.
\newblock


\bibitem[Saleheen et~al\mbox{.}(2021)]%
        {saleheen2021wristprint}
\bibfield{author}{\bibinfo{person}{Nazir Saleheen}, \bibinfo{person}{Md~Azim Ullah}, \bibinfo{person}{Supriyo Chakraborty}, \bibinfo{person}{Deniz~S Ones}, \bibinfo{person}{Mani Srivastava}, {and} \bibinfo{person}{Santosh Kumar}.} \bibinfo{year}{2021}\natexlab{}.
\newblock \showarticletitle{Wristprint: Characterizing user re-identification risks from wrist-worn accelerometry data}. In \bibinfo{booktitle}{\emph{Proceedings of the 2021 ACM SIGSAC Conference on Computer and Communications Security}}. \bibinfo{pages}{2807--2823}.
\newblock


\bibitem[Sarhaddi et~al\mbox{.}(2022)]%
        {sarhaddi2022comprehensive}
\bibfield{author}{\bibinfo{person}{Fatemeh Sarhaddi}, \bibinfo{person}{Kianoosh Kazemi}, \bibinfo{person}{Iman Azimi}, \bibinfo{person}{Rui Cao}, \bibinfo{person}{Hannakaisa Niela-Vil{\'e}n}, \bibinfo{person}{Anna Axelin}, \bibinfo{person}{Pasi Liljeberg}, {and} \bibinfo{person}{Amir~M Rahmani}.} \bibinfo{year}{2022}\natexlab{}.
\newblock \showarticletitle{A comprehensive accuracy assessment of Samsung smartwatch heart rate and heart rate variability}.
\newblock \bibinfo{journal}{\emph{PloS one}} \bibinfo{volume}{17}, \bibinfo{number}{12} (\bibinfo{year}{2022}), \bibinfo{pages}{e0268361}.
\newblock


\bibitem[Schmidt et~al\mbox{.}(2019)]%
        {schmidt2019multi}
\bibfield{author}{\bibinfo{person}{Philip Schmidt}, \bibinfo{person}{Robert D{\"u}richen}, \bibinfo{person}{Attila Reiss}, \bibinfo{person}{Kristof Van~Laerhoven}, {and} \bibinfo{person}{Thomas Pl{\"o}tz}.} \bibinfo{year}{2019}\natexlab{}.
\newblock \showarticletitle{Multi-target affect detection in the wild: an exploratory study}. In \bibinfo{booktitle}{\emph{Proceedings of the 2019 ACM International Symposium on Wearable Computers}}. \bibinfo{pages}{211--219}.
\newblock


\bibitem[Schmidt et~al\mbox{.}(2018a)]%
        {schmidt2018wesad}
\bibfield{author}{\bibinfo{person}{Philip Schmidt}, \bibinfo{person}{Attila Reiss}, \bibinfo{person}{Robert Duerichen}, \bibinfo{person}{Claus Marberger}, {and} \bibinfo{person}{Kristof Van~Laerhoven}.} \bibinfo{year}{2018}\natexlab{a}.
\newblock \showarticletitle{Introducing wesad, a multimodal dataset for wearable stress and affect detection}. In \bibinfo{booktitle}{\emph{Proceedings of the 20th ACM international conference on multimodal interaction}}. \bibinfo{pages}{400--408}.
\newblock


\bibitem[Schmidt et~al\mbox{.}(2018b)]%
        {schmidt2018introducing}
\bibfield{author}{\bibinfo{person}{Philip Schmidt}, \bibinfo{person}{Attila Reiss}, \bibinfo{person}{Robert Duerichen}, \bibinfo{person}{Claus Marberger}, {and} \bibinfo{person}{Kristof Van~Laerhoven}.} \bibinfo{year}{2018}\natexlab{b}.
\newblock \showarticletitle{Introducing wesad, a multimodal dataset for wearable stress and affect detection}. In \bibinfo{booktitle}{\emph{Proceedings of the 20th ACM international conference on multimodal interaction}}. \bibinfo{pages}{400--408}.
\newblock


\bibitem[Shin and Min(2017)]%
        {shin2017feasibility}
\bibfield{author}{\bibinfo{person}{Hangsik Shin} {and} \bibinfo{person}{Se~Dong Min}.} \bibinfo{year}{2017}\natexlab{}.
\newblock \showarticletitle{Feasibility study for the non-invasive blood pressure estimation based on ppg morphology: Normotensive subject study}.
\newblock \bibinfo{journal}{\emph{Biomedical engineering online}}  \bibinfo{volume}{16} (\bibinfo{year}{2017}), \bibinfo{pages}{1--14}.
\newblock


\bibitem[Smets et~al\mbox{.}(2018)]%
        {smets2018large}
\bibfield{author}{\bibinfo{person}{Elena Smets}, \bibinfo{person}{Emmanuel Rios~Velazquez}, \bibinfo{person}{Giuseppina Schiavone}, \bibinfo{person}{Imen Chakroun}, \bibinfo{person}{Ellie D’Hondt}, \bibinfo{person}{Walter De~Raedt}, \bibinfo{person}{Jan Cornelis}, \bibinfo{person}{Olivier Janssens}, \bibinfo{person}{Sofie Van~Hoecke}, \bibinfo{person}{Stephan Claes}, {et~al\mbox{.}}} \bibinfo{year}{2018}\natexlab{}.
\newblock \showarticletitle{Large-scale wearable data reveal digital phenotypes for daily-life stress detection}.
\newblock \bibinfo{journal}{\emph{NPJ digital medicine}} \bibinfo{volume}{1}, \bibinfo{number}{1} (\bibinfo{year}{2018}), \bibinfo{pages}{67}.
\newblock


\bibitem[Spathis et~al\mbox{.}(2022)]%
        {spathis2022breaking}
\bibfield{author}{\bibinfo{person}{Dimitris Spathis}, \bibinfo{person}{Ignacio Perez-Pozuelo}, \bibinfo{person}{Laia Marques-Fernandez}, {and} \bibinfo{person}{Cecilia Mascolo}.} \bibinfo{year}{2022}\natexlab{}.
\newblock \showarticletitle{Breaking away from labels: The promise of self-supervised machine learning in intelligent health}.
\newblock \bibinfo{journal}{\emph{Patterns}} \bibinfo{volume}{3}, \bibinfo{number}{2} (\bibinfo{year}{2022}).
\newblock


\bibitem[Tang et~al\mbox{.}(2020)]%
        {tang2020exploring}
\bibfield{author}{\bibinfo{person}{Chi~Ian Tang}, \bibinfo{person}{Ignacio Perez-Pozuelo}, \bibinfo{person}{Dimitris Spathis}, {and} \bibinfo{person}{Cecilia Mascolo}.} \bibinfo{year}{2020}\natexlab{}.
\newblock \showarticletitle{Exploring contrastive learning in human activity recognition for healthcare}.
\newblock \bibinfo{journal}{\emph{arXiv preprint arXiv:2011.11542}} (\bibinfo{year}{2020}).
\newblock


\bibitem[Tonekaboni et~al\mbox{.}(2021)]%
        {tonekaboni2021tnc}
\bibfield{author}{\bibinfo{person}{Sana Tonekaboni}, \bibinfo{person}{Danny Eytan}, {and} \bibinfo{person}{Anna Goldenberg}.} \bibinfo{year}{2021}\natexlab{}.
\newblock \showarticletitle{Unsupervised representation learning for time series with temporal neighborhood coding}.
\newblock \bibinfo{journal}{\emph{International Conference of Learning Representations}} (\bibinfo{year}{2021}).
\newblock


\bibitem[Toshnazarov et~al\mbox{.}(2024)]%
        {toshnazarov2024sosw}
\bibfield{author}{\bibinfo{person}{Kobiljon Toshnazarov}, \bibinfo{person}{Uichin Lee}, \bibinfo{person}{Byung~Hyung Kim}, \bibinfo{person}{Varun Mishra}, \bibinfo{person}{Lismer Andres~Caceres Najarro}, {and} \bibinfo{person}{Youngtae Noh}.} \bibinfo{year}{2024}\natexlab{}.
\newblock \showarticletitle{SOSW: Stress Sensing with Off-the-shelf Smartwatches in the Wild}.
\newblock \bibinfo{journal}{\emph{IEEE Internet of Things Journal}} (\bibinfo{year}{2024}).
\newblock


\bibitem[Truslow et~al\mbox{.}(2024)]%
        {truslow2024understanding}
\bibfield{author}{\bibinfo{person}{James Truslow}, \bibinfo{person}{Angela Spillane}, \bibinfo{person}{Huiming Lin}, \bibinfo{person}{Katherine Cyr}, \bibinfo{person}{Adeeti Ullal}, \bibinfo{person}{Edith Arnold}, \bibinfo{person}{Ron Huang}, \bibinfo{person}{Laura Rhodes}, \bibinfo{person}{Jennifer Block}, \bibinfo{person}{Jamie Stark}, {et~al\mbox{.}}} \bibinfo{year}{2024}\natexlab{}.
\newblock \showarticletitle{Understanding activity and physiology at scale: The Apple Heart \& Movement Study}.
\newblock \bibinfo{journal}{\emph{npj Digital Medicine}} \bibinfo{volume}{7}, \bibinfo{number}{1} (\bibinfo{year}{2024}), \bibinfo{pages}{242}.
\newblock


\bibitem[van Gilst et~al\mbox{.}(2019)]%
        {van2019protocol}
\bibfield{author}{\bibinfo{person}{Merel~M van Gilst}, \bibinfo{person}{Johannes~P van Dijk}, \bibinfo{person}{Roy Krijn}, \bibinfo{person}{Bertram Hoondert}, \bibinfo{person}{Pedro Fonseca}, \bibinfo{person}{Ruud~JG van Sloun}, \bibinfo{person}{Bruno Arsenali}, \bibinfo{person}{Nele Vandenbussche}, \bibinfo{person}{Sigrid Pillen}, \bibinfo{person}{Henning Maass}, {et~al\mbox{.}}} \bibinfo{year}{2019}\natexlab{}.
\newblock \showarticletitle{Protocol of the SOMNIA project: an observational study to create a neurophysiological database for advanced clinical sleep monitoring}.
\newblock \bibinfo{journal}{\emph{BMJ open}} \bibinfo{volume}{9}, \bibinfo{number}{11} (\bibinfo{year}{2019}), \bibinfo{pages}{e030996}.
\newblock


\bibitem[Vaswani et~al\mbox{.}(2017)]%
        {vaswani2017attnisallyouneedtransformer}
\bibfield{author}{\bibinfo{person}{Ashish Vaswani}, \bibinfo{person}{Noam Shazeer}, \bibinfo{person}{Niki Parmar}, \bibinfo{person}{Jakob Uszkoreit}, \bibinfo{person}{Llion Jones}, \bibinfo{person}{Aidan~N Gomez}, \bibinfo{person}{{\L}ukasz Kaiser}, {and} \bibinfo{person}{Illia Polosukhin}.} \bibinfo{year}{2017}\natexlab{}.
\newblock \showarticletitle{Attention is all you need}.
\newblock \bibinfo{journal}{\emph{Advances in neural information processing systems}}  \bibinfo{volume}{30} (\bibinfo{year}{2017}).
\newblock


\bibitem[Wang et~al\mbox{.}(2024)]%
        {wang2024classifying}
\bibfield{author}{\bibinfo{person}{Jie Wang}, \bibinfo{person}{Tuantuan Lu}, \bibinfo{person}{Ruogu Huang}, {and} \bibinfo{person}{Yongxiang Zhao}.} \bibinfo{year}{2024}\natexlab{}.
\newblock \showarticletitle{Classifying engagement in E-learning through GRU-TCN model using photoplethysmography signals}.
\newblock \bibinfo{journal}{\emph{Biomedical Signal Processing and Control}}  \bibinfo{volume}{90} (\bibinfo{year}{2024}), \bibinfo{pages}{105903}.
\newblock


\bibitem[Wilson et~al\mbox{.}(2018)]%
        {wilson2018couples}
\bibfield{author}{\bibinfo{person}{Stephanie~J Wilson}, \bibinfo{person}{Brittney~E Bailey}, \bibinfo{person}{Lisa~M Jaremka}, \bibinfo{person}{Christopher~P Fagundes}, \bibinfo{person}{Rebecca Andridge}, \bibinfo{person}{William~B Malarkey}, \bibinfo{person}{Kathleen~M Gates}, {and} \bibinfo{person}{Janice~K Kiecolt-Glaser}.} \bibinfo{year}{2018}\natexlab{}.
\newblock \showarticletitle{When couples’ hearts beat together: Synchrony in heart rate variability during conflict predicts heightened inflammation throughout the day}.
\newblock \bibinfo{journal}{\emph{Psychoneuroendocrinology}}  \bibinfo{volume}{93} (\bibinfo{year}{2018}), \bibinfo{pages}{107--116}.
\newblock


\bibitem[Woo et~al\mbox{.}(2022)]%
        {woo2022cost}
\bibfield{author}{\bibinfo{person}{Gerald Woo}, \bibinfo{person}{Chenghao Liu}, \bibinfo{person}{Doyen Sahoo}, \bibinfo{person}{Akshat Kumar}, {and} \bibinfo{person}{Steven Hoi}.} \bibinfo{year}{2022}\natexlab{}.
\newblock \showarticletitle{CoST: Contrastive learning of disentangled seasonal-trend representations for time series forecasting}.
\newblock \bibinfo{journal}{\emph{arXiv preprint arXiv:2202.01575}} (\bibinfo{year}{2022}).
\newblock


\bibitem[Xu et~al\mbox{.}(2022)]%
        {xu2022pulseimpute}
\bibfield{author}{\bibinfo{person}{Maxwell Xu}, \bibinfo{person}{Alexander Moreno}, \bibinfo{person}{Supriya Nagesh}, \bibinfo{person}{Varol Aydemir}, \bibinfo{person}{David Wetter}, \bibinfo{person}{Santosh Kumar}, {and} \bibinfo{person}{James~M Rehg}.} \bibinfo{year}{2022}\natexlab{}.
\newblock \showarticletitle{PulseImpute: A Novel Benchmark Task for Pulsative Physiological Signal Imputation}.
\newblock \bibinfo{journal}{\emph{Advances in Neural Information Processing Systems Dataset and Benchmarks Track}}  \bibinfo{volume}{35} (\bibinfo{year}{2022}), \bibinfo{pages}{26874--26888}.
\newblock


\bibitem[Xu et~al\mbox{.}(2024a)]%
        {xurebar}
\bibfield{author}{\bibinfo{person}{Maxwell Xu}, \bibinfo{person}{Alexander Moreno}, \bibinfo{person}{Hui Wei}, \bibinfo{person}{Benjamin Marlin}, {and} \bibinfo{person}{James~Matthew Rehg}.} \bibinfo{year}{2024}\natexlab{a}.
\newblock \showarticletitle{REBAR: Retrieval-Based Reconstruction for Time-series Contrastive Learning}. In \bibinfo{booktitle}{\emph{The Twelfth International Conference on Learning Representations}}.
\newblock


\bibitem[Xu et~al\mbox{.}(2024b)]%
        {xu2024rebar}
\bibfield{author}{\bibinfo{person}{Maxwell Xu}, \bibinfo{person}{Alexander Moreno}, \bibinfo{person}{Hui Wei}, \bibinfo{person}{Benjamin Marlin}, {and} \bibinfo{person}{James~Matthew Rehg}.} \bibinfo{year}{2024}\natexlab{b}.
\newblock \showarticletitle{REBAR: Retrieval-Based Reconstruction for Time-series Contrastive Learning}. In \bibinfo{booktitle}{\emph{The Twelfth International Conference on Learning Representations}}.
\newblock


\bibitem[Xu et~al\mbox{.}(2024c)]%
        {xu2024relcon}
\bibfield{author}{\bibinfo{person}{Maxwell~A Xu}, \bibinfo{person}{Jaya Narain}, \bibinfo{person}{Gregory Darnell}, \bibinfo{person}{Haraldur Hallgrimsson}, \bibinfo{person}{Hyewon Jeong}, \bibinfo{person}{Darren Forde}, \bibinfo{person}{Richard Fineman}, \bibinfo{person}{Karthik~J Raghuram}, \bibinfo{person}{James~M Rehg}, {and} \bibinfo{person}{Shirley Ren}.} \bibinfo{year}{2024}\natexlab{c}.
\newblock \showarticletitle{RelCon: Relative Contrastive Learning for a Motion Foundation Model for Wearable Data}.
\newblock \bibinfo{journal}{\emph{arXiv preprint arXiv:2411.18822}} (\bibinfo{year}{2024}).
\newblock


\bibitem[Yang et~al\mbox{.}(2024)]%
        {yang2024implementation}
\bibfield{author}{\bibinfo{person}{Jiamei Yang}, \bibinfo{person}{Yu Wang}, \bibinfo{person}{Hui Wang}, \bibinfo{person}{Peng Zhou}, \bibinfo{person}{Xinyou Li}, {and} \bibinfo{person}{Jianbin Zheng}.} \bibinfo{year}{2024}\natexlab{}.
\newblock \showarticletitle{Implementation of FPGA-based ResNet accelerator for vehicle detection}. In \bibinfo{booktitle}{\emph{Ninth International Symposium on Advances in Electrical, Electronics, and Computer Engineering (ISAEECE 2024)}}, Vol.~\bibinfo{volume}{13291}. SPIE, \bibinfo{pages}{100--105}.
\newblock


\bibitem[Yang and Hong(2022)]%
        {yang2022btsf}
\bibfield{author}{\bibinfo{person}{Ling Yang} {and} \bibinfo{person}{Shenda Hong}.} \bibinfo{year}{2022}\natexlab{}.
\newblock \showarticletitle{Unsupervised time-series representation learning with iterative bilinear temporal-spectral fusion}. In \bibinfo{booktitle}{\emph{International Conference on Machine Learning}}. PMLR, \bibinfo{pages}{25038--25054}.
\newblock


\bibitem[Yang et~al\mbox{.}(2022)]%
        {yang2022timeclr}
\bibfield{author}{\bibinfo{person}{Xinyu Yang}, \bibinfo{person}{Zhenguo Zhang}, {and} \bibinfo{person}{Rongyi Cui}.} \bibinfo{year}{2022}\natexlab{}.
\newblock \showarticletitle{Timeclr: A self-supervised contrastive learning framework for univariate time series representation}.
\newblock \bibinfo{journal}{\emph{Knowledge-Based Systems}}  \bibinfo{volume}{245} (\bibinfo{year}{2022}), \bibinfo{pages}{108606}.
\newblock


\bibitem[Yao et~al\mbox{.}(2023)]%
        {yao2023intelligent}
\bibfield{author}{\bibinfo{person}{Bowen Yao}, \bibinfo{person}{Liansheng Liu}, \bibinfo{person}{Yu Peng}, {and} \bibinfo{person}{Xiyuan Peng}.} \bibinfo{year}{2023}\natexlab{}.
\newblock \showarticletitle{Intelligent measurement on edge devices using hardware memory-aware joint compression enabled neural networks}.
\newblock \bibinfo{journal}{\emph{IEEE Transactions on Instrumentation and Measurement}}  \bibinfo{volume}{73} (\bibinfo{year}{2023}), \bibinfo{pages}{1--13}.
\newblock


\bibitem[Yue et~al\mbox{.}(2022)]%
        {yue2022ts2vec}
\bibfield{author}{\bibinfo{person}{Zhihan Yue}, \bibinfo{person}{Yujing Wang}, \bibinfo{person}{Juanyong Duan}, \bibinfo{person}{Tianmeng Yang}, \bibinfo{person}{Congrui Huang}, \bibinfo{person}{Yunhai Tong}, {and} \bibinfo{person}{Bixiong Xu}.} \bibinfo{year}{2022}\natexlab{}.
\newblock \showarticletitle{Ts2vec: Towards universal representation of time series}. In \bibinfo{booktitle}{\emph{Proceedings of the AAAI Conference on Artificial Intelligence}}, Vol.~\bibinfo{volume}{36}. \bibinfo{pages}{8980--8987}.
\newblock


\bibitem[Yun et~al\mbox{.}(2024)]%
        {yun2024unsupervised}
\bibfield{author}{\bibinfo{person}{Taedong Yun}, \bibinfo{person}{Justin Cosentino}, \bibinfo{person}{Babak Behsaz}, \bibinfo{person}{Zachary~R McCaw}, \bibinfo{person}{Davin Hill}, \bibinfo{person}{Robert Luben}, \bibinfo{person}{Dongbing Lai}, \bibinfo{person}{John Bates}, \bibinfo{person}{Howard Yang}, \bibinfo{person}{Tae-Hwi Schwantes-An}, {et~al\mbox{.}}} \bibinfo{year}{2024}\natexlab{}.
\newblock \showarticletitle{Unsupervised representation learning on high-dimensional clinical data improves genomic discovery and prediction}.
\newblock \bibinfo{journal}{\emph{Nature Genetics}} \bibinfo{volume}{56}, \bibinfo{number}{8} (\bibinfo{year}{2024}), \bibinfo{pages}{1604--1613}.
\newblock


\bibitem[Zhang et~al\mbox{.}(2024a)]%
        {zhang2024reproducible}
\bibfield{author}{\bibinfo{person}{Panyu Zhang}, \bibinfo{person}{Gyuwon Jung}, \bibinfo{person}{Jumabek Alikhanov}, \bibinfo{person}{Uzair Ahmed}, {and} \bibinfo{person}{Uichin Lee}.} \bibinfo{year}{2024}\natexlab{a}.
\newblock \showarticletitle{A Reproducible Stress Prediction Pipeline with Mobile Sensor Data}.
\newblock \bibinfo{journal}{\emph{Proceedings of the ACM on interactive, mobile, wearable and ubiquitous technologies}} \bibinfo{volume}{8}, \bibinfo{number}{3} (\bibinfo{year}{2024}), \bibinfo{pages}{1--35}.
\newblock


\bibitem[Zhang et~al\mbox{.}(2022)]%
        {zhang2022tfc}
\bibfield{author}{\bibinfo{person}{Xiang Zhang}, \bibinfo{person}{Ziyuan Zhao}, \bibinfo{person}{Theodoros Tsiligkaridis}, {and} \bibinfo{person}{Marinka Zitnik}.} \bibinfo{year}{2022}\natexlab{}.
\newblock \showarticletitle{Self-supervised contrastive pre-training for time series via time-frequency consistency}.
\newblock \bibinfo{journal}{\emph{Advances in Neural Information Processing Systems}}  \bibinfo{volume}{35} (\bibinfo{year}{2022}), \bibinfo{pages}{3988--4003}.
\newblock


\bibitem[Zhang et~al\mbox{.}(2024b)]%
        {zhang2024general}
\bibfield{author}{\bibinfo{person}{Zexing Zhang}, \bibinfo{person}{Huimin Lu}, \bibinfo{person}{Songzhe Ma}, \bibinfo{person}{Jianzhong Peng}, \bibinfo{person}{Chenglin Lin}, \bibinfo{person}{Niya Li}, {and} \bibinfo{person}{Bingwang Dong}.} \bibinfo{year}{2024}\natexlab{b}.
\newblock \showarticletitle{A general framework for generative self-supervised learning in non-invasive estimation of physiological parameters using photoplethysmography}.
\newblock \bibinfo{journal}{\emph{Biomedical Signal Processing and Control}}  \bibinfo{volume}{98} (\bibinfo{year}{2024}), \bibinfo{pages}{106788}.
\newblock


\bibitem[Zhou et~al\mbox{.}(2024)]%
        {zhou2024comprehensive}
\bibfield{author}{\bibinfo{person}{Ce Zhou}, \bibinfo{person}{Qian Li}, \bibinfo{person}{Chen Li}, \bibinfo{person}{Jun Yu}, \bibinfo{person}{Yixin Liu}, \bibinfo{person}{Guangjing Wang}, \bibinfo{person}{Kai Zhang}, \bibinfo{person}{Cheng Ji}, \bibinfo{person}{Qiben Yan}, \bibinfo{person}{Lifang He}, {et~al\mbox{.}}} \bibinfo{year}{2024}\natexlab{}.
\newblock \showarticletitle{A comprehensive survey on pretrained foundation models: A history from bert to chatgpt}.
\newblock \bibinfo{journal}{\emph{International Journal of Machine Learning and Cybernetics}} (\bibinfo{year}{2024}), \bibinfo{pages}{1--65}.
\newblock


\bibitem[Zhu et~al\mbox{.}(2023)]%
        {zhu2023stress}
\bibfield{author}{\bibinfo{person}{Lili Zhu}, \bibinfo{person}{Petros Spachos}, \bibinfo{person}{Pai~Chet Ng}, \bibinfo{person}{Yuanhao Yu}, \bibinfo{person}{Yang Wang}, \bibinfo{person}{Konstantinos Plataniotis}, {and} \bibinfo{person}{Dimitrios Hatzinakos}.} \bibinfo{year}{2023}\natexlab{}.
\newblock \showarticletitle{Stress detection through wrist-based electrodermal activity monitoring and machine learning}.
\newblock \bibinfo{journal}{\emph{IEEE Journal of Biomedical and Health Informatics}} \bibinfo{volume}{27}, \bibinfo{number}{5} (\bibinfo{year}{2023}), \bibinfo{pages}{2155--2165}.
\newblock


\end{thebibliography}

\newpage
\appendix
\section{Appendix}
\subsection{Evaluation Dataset Details} \label{sec:evaldata}
Our evaluation benchmarking approach demonstrates the versatility of our PPG foundation model, showcasing its ability to adapt to the domain shift of various datasets and generalize well across diverse health-related tasks, notably after learning from noisy PPG data from a field dataset. \edit{We describe each downstream paradigm and associated dataset below: }

\textbf{Cardiovascular health \edit{(PPG-BP \citep{elgendi2022datasetppgbp})}} is a key area of focus, with biomarkers such as systolic blood pressure, diastolic blood pressure, and heart rate serving as vital indicators. The PPG-BP dataset includes short PPG recordings from 219 participants, most of whom have hypertension, to acquire information on the basic physiology of individuals. As part of the data collection process, participants were asked to relax for about 10 minutes. After that, both blood pressure and finger PPG were collected within the next 3 minutes with 3 short PPG recordings of 2.1 seconds each per participant, along with corresponding heart rate measurements~\cite{liang2018new}. Using this dataset, we estimate these biomarkers and predict hypertension.

\textbf{Stress and affective states \edit{(WESAD \citep{schmidt2018introducing})}}, which significantly impact physical and mental health, are evaluated using the WESAD dataset~\cite{schmidt2018introducing}. The WESAD dataset collected physiological signals from 15 participants (12 males, 3 females) using RespiBAN Professional \citep{biosignalsplux} on the chest and an Empatica E4 \citep{mccarthy2016validation} on the wrist. \edit{From E4,} the dataset \edit{collected Accelerometer (ACC)}, Blood Volume Pulse (BVP), Electrodermal Activity (EDA), and Skin Temperature (TEMP) \edit{data} at \edit{32} Hz, 64 Hz, 4 Hz, and 4 Hz, respectively. The study protocol had five sessions: Baseline, Stress, Amusement, Meditation, and Rest. In the baseline session, participants engaged in neutral activities for 20 minutes. Stress was induced using the Trier Social Stress for about 10 minutes. During the session, participants had to give a public speech and complete a mental arithmetic task for 5 minutes each. For the amusement session, participants watched funny video clips for 6.5 minutes. To facilitate recovery, a meditation session followed both the stress and amusement sessions, along with an additional 10-minute rest period after the stress session. 

As part of the downstream evaluation tasks, we perform both binary and multi-class classification using only PPG data collected during the sessions. For binary classification, we combined baseline and amusement sessions to form the non-stress class, while the stress class was the original stress session. For multi-class classification, we use Stress, Baseline, Amusement, and Mediation sessions. For both types of classification, each session was divided into non-overlapping 1-minute windows. Hence, the stress vs. non-stress class had approximately 10 vs. 26 windows for the binary classification, while for the multi-class classification, the window distribution was 20, 10, 6, and 7 for Baseline, Stress, Amusement, and Mediation, respectively.

\textbf{Physical activity \edit{(PPG-DaLiA \citep{reiss2019deep})}}, which is a key indicator of one's healthy lifestyle, is assessed using the PPG-DaLiA dataset~\cite{reiss2019deep}. 
This publicly available multimodal dataset comprises physiological and motion data collected from wrist- and chest-worn wearable devices. It includes recordings from 15 participants (seven males and eight females) aged 21 to 55 years, performing eight distinct activities (sitting still, ascending/descending stairs, playing table soccer, cycling, driving, taking lunch breaks, walking, and working). Additionally, transition periods between activities were labeled as a separate ‘zero’ activity. All activities were conducted in conditions closely resembling real-life scenarios, with each participant’s recorded session lasting approximately 2.5 hours.
Data were collected using the RespiBAN Professional and Empatica E4 wearable devices. The dataset includes PPG recordings and heart rate measurements during various activities. We utilize this dataset for two key tasks: activity classification (9 classes) and instantaneous heart rate estimation.

\textbf{Sleep-disordered breathing \edit{(SDB \citep{garde2014development})}} can lead to significant health complications in both adults and children. In children, SDB is associated with issues ranging from daytime sleepiness to severe conditions like developmental delays and growth failure. To examine its impact, the SDB dataset~\cite{garde2014development} collected physiological data from 146 children using a Phone Oximeter~\cite{karlen2011human}, which recorded PPG at 62.5 Hz and SpO$_2$ at 1 Hz during six hours of overnight sleep. Participants were categorized into SDB-positive and SDB-negative groups based on their Apnea-Hypopnea Index (AHI) scores, which quantify the severity of obstructive sleep apnea. Following the approach in~\cite{pillai2024papagei}, we utilize these AHI ratings to frame SDB prediction as a binary classification task.

\textbf{Stress Detection and Physical Activity in the Field  \edit{(MOODS \citep{neupane2024momentarymoods})}}
In the field study, participants received on average 5.2 prompts/day from which they responded to 3.9, using a Likert scale---\emph{`Not stressed,' `Probably not stressed,' `Unsure,' `Probably Stressed,' `Stressed'}, yielding a response ratio of 75\%. If the \edit{response} was either \emph{`Stressed,' `Probably Stressed,' or `Unsure'}, the app \edit{required} them to provide \edit{both} semantic location and the likely stressor. From a total of 128,006 physiological events detected by an AI model running on the smartwatch, 35,785 events were \edit{prompted} to participants, of which they \edit{responded} to 26,682 events. Following prior research~\cite{toshnazarov2024sosw, mishra2018investigating}, we \edit{divide these annotated events into} 1-minute windows for the downstream \edit{task of detecting} stress vs. non-stress. \edit{However}, we remove \edit{data of 3} participants who exhibit no variation in their stress ratings to address label noise \edit{and 16 participants} with a low compliance rate in submitting stress reports \edit{(less than one standard deviation of population mean responses). This exclusion results} in a revised train/validation/test split of 74/14/13 participants. For classification, we group \emph{`Stressed,' `Probably Stressed, `Unsure'} into the Stress class, \edit{and} \emph{`Not stressed,' `Probably not stressed'} \edit{into the} Non-Stress class \edit{with a distribution of} approximately 42\% and 58\%, respectively, \edit{for} a total of 206,088 one-minute stress and non-stress windows.

For the other task \edit{of physical activity}, we chose stationary vs. \edit{non-stationary} as the downstream task. We use the first 10 days of each participant's data, assuming this duration \edit{sufficiently captures} an individual's stationary and motion behavior. Since the field dataset lacks explicit activity labels, we \edit{apply} a pre-trained activity recognition model~\cite{saleheen2021wristprint} to classify 20-second \edit{windows} of \edit{motion} data into five categories: \emph{`Stationary,' `Walking', `Stairs', `Sports', and `Exercise'}. We group \emph{`Walking,' `Stairs,' `Sports,' and `Exercise'} into the non-stationary class. From these 10 days of data, we obtain a total of 3,559,429 20-second windows, with a class distribution of 15\% stationary and 85\% non-stationary. We extract raw PPG data corresponding to these 20-second windows for each participant. To be consistent with prior experiments, we use the same original \edit{participants'} split of 84/18/18 to partition 3,559,429 windows \edit{into train, validation, and test sets} for the downstream task.

\begin{table*}[!htbp]
    \centering
    \LARGE
    \color{black}
    \captionsetup{
        labelfont={color=black},
        textfont={color=black},
        font={small}
    }
    \caption{\textbf{Light Pulse-PPG vs. PaPaGei\edit{, Controlling for Parameters and Input Size.}} \edit{Mean} Performance across \edit{Classification and Regression Tasks. The best is bolded, and the second is underlined. Full results in Table~\ref{tab:exp23}. Light Pulse-PPG matches PaPaGei's parameter count, as well as input window length, and continues to outperform PaPaGei.}
    }
    \label{tab:light_pulse_ppg_mean}

    \resizebox{\textwidth}{!}{%
    \begin{tabular}{lccccccccc}
        \midrule
        & \multicolumn{6}{c}{\textbf{Avg. Classification Performance}} 
        & \multicolumn{3}{c}{\textbf{Avg. Regression Performance}} \\
        \cmidrule(lr){2-7} \cmidrule(lr){8-10}
        \textbf{Model} & F1 Score & Accuracy & Precision & Recall & AUPRC & AUROC 
        & MAE & MSE & MAPE \\
        \midrule
        Light Pulse-PPG  & \edit{$\boldsymbol{\textcolor{black}{0.5205} \pm 0.16}$} & \edit{$\boldsymbol{\textcolor{black}{0.6726} \pm 0.16}$} & \edit{$\boldsymbol{\textcolor{black}{0.5873} \pm 0.15}$} & \edit{$\boldsymbol{\textcolor{black}{0.5360} \pm 0.15}$} & \edit{$\boldsymbol{\textcolor{black}{0.4612} \pm 0.18}$} & \edit{$\boldsymbol{\textcolor{black}{0.6997} \pm 0.14}$} & \edit{$\boldsymbol{\textcolor{black}{9.4940} \pm 6.00}$} & \edit{$\boldsymbol{\textcolor{black}{171.25} \pm 197.09}$} & \edit{$\boldsymbol{\textcolor{black}{0.1082} \pm 0.05}$} \\
        
        PaPaGei      & \edit{\underline{$0.4994 \pm 0.14$}} & 
        \edit{\underline{$0.6361 \pm 0.15$}} & 
        \edit{\underline{$0.5530 \pm 0.14$}} & 
        \edit{\underline{$0.5210 \pm 0.14$}} & 
        \edit{\underline{$0.4164 \pm 0.12$}} & 
        \edit{\underline{$0.6840 \pm 0.10$}} & 
        \edit{\underline{$10.7488 \pm 7.75$}} & 
        \edit{\underline{$204.6825 \pm 237.78$}} & 
        \edit{\underline{$0.1288 \pm 0.10$}} \\
        \bottomrule
    \end{tabular}%
    }
\end{table*}

\subsection{Experiment \edit{5: Light Pulse-PPG vs. PaPaGei, Controlling for Parameters and Input Size}}\label{sec:light_pulse_ppg_vs_papagei}
\subsubsection{Background} The original Pulse-PPG has 28.5 million parameters (to better capture the noise and variability in field-collected PPG), while PaPaGei-S, a model from PaPaGei's family of models against which we are comparing, had 5.7 million parameters \edit{with an input window of 10 seconds}. Hence, to isolate the impact of field data pre-training from the impact of these architectural and window-size differences, which would ensure a fair comparison between Pulse-PPG and PaPaGei, we train a lighter version of Pulse-PPG with a significantly reduced parameter count \edit{and window size}.

\subsubsection{Experimental Design:} We segmented the field PPG timeseries of each participant into 10-second non-overlapping windows. For architectural and hyperparameter alignment with PaPaGei, we adopted a ResNet 18-block architecture, reduced the kernel size to 3, and adjusted the initial base filters to 32. These modifications resulted in a Light Pulse-PPG model with approximately 4.73 million parameters, significantly reduced from the original model’s 28.5 million parameters. We kept the train, val, and test splits of pre-training data the same as the original 4-minute Pulse-PPG model. We then pre-trained and evaluated this, which we refer to as the Light Pulse-PPG model, using the same linear probe evaluation pipeline as the
original 4-minute model.

\subsubsection{Results} We summarize mean performance metrics across classification and regression tasks for the Light Pulse-PPG and PaPaGei models, trained on field and clinical datasets, respectively, in Table \ref{tab:light_pulse_ppg_mean}. Our findings indicate that the parameter-matched and window-aligned Light Pulse-PPG model outperforms PaPaGei trained on clinical data across all mean performance metrics. This indicates that even with controlled experiments where the model becomes comparable in scale and input context, there is still more value in pre-training on field data. This becomes even more significant because it is increasingly hard to learn meaningful patterns from the field data when the window size becomes as small as 10 seconds, due to the inherent increased noisiness.

\begin{table*}[!htbp]
    \centering
    % \fontsize{4}{5}\selectfont % Custom font size (4pt) with 5pt baseline skip
    \LARGE
    \color{black}
    \captionsetup{
        labelfont={color=black}, 
        textfont={color=black}, 
        font={small}
    }
    \caption{\textbf{Input Ablation Study.} \edit{Mean} Performance across \edit{Classification and Regression Tasks. The best is bolded, and the second is underlined. Full results in Table~\ref{tab:ablation_results}. Results show that normalization consistently improves performance, and 4-minute input windows are the best.}
    }
    \label{tab:ablation_mean}
    \resizebox{\textwidth}{!}{%
    \begin{tabular}{@{}lccccccccc@{}}
        \toprule
        & \multicolumn{6}{c}{\textbf{Avg. Classification Performance}} 
        & \multicolumn{3}{c}{\textbf{Avg. Regression Performance}} \\
        \cmidrule(lr){2-7} \cmidrule(lr){8-10}
        \textbf{Pulse-PPG Variant} &  F1 Score & Accuracy & Precision & Recall & AUPRC & AUROC 
        & MAE & MSE & MAPE \\
        \midrule
        (4 min, norm) & \edit{$\boldsymbol{\textcolor{black}{0.5870} \pm 0.17}$} & \edit{$\boldsymbol{\textcolor{black}{0.6840} \pm 0.17}$} & \edit{\underline{$\textcolor{black}{0.6237} \pm 0.16$}} & \edit{\underline{$\textcolor{black}{0.5894} \pm 0.17$}} & \edit{$\boldsymbol{\textcolor{black}{0.5149} \pm 0.20}$} & \edit{$\boldsymbol{\textcolor{black}{0.7481} \pm 0.17}$} & \edit{$\boldsymbol{\textcolor{black}{8.8593} \pm 6.25}$} & \edit{$\boldsymbol{\textcolor{black}{145.2550} \pm 171.08}$} & \edit{$\boldsymbol{\textcolor{black}{0.1009} \pm 0.05}$} \\
        
        (1 min, norm) & \edit{\underline{$\textcolor{black}{0.5851} \pm 0.17$}} & \edit{\underline{$\textcolor{black}{0.6785} \pm 0.17$}} & \edit{$\textcolor{black}{0.6198} \pm 0.16$} & \edit{$\boldsymbol{\textcolor{black}{0.5904} \pm 0.18}$} & \edit{$\textcolor{black}{0.5065} \pm 0.18$} & \edit{\underline{$\textcolor{black}{0.7441} \pm 0.16$}} & \edit{\underline{$\textcolor{black}{8.9284} \pm 6.22$}} & \edit{\underline{$\textcolor{black}{148.7335} \pm 177.05$}} & \edit{$\textcolor{black}{0.1028} \pm 0.05$} \\
        
        (1 min, no norm) & \edit{$\textcolor{black}{0.5620} \pm 0.16$} & \edit{$\textcolor{black}{0.6494} \pm 0.16$} & \edit{$\textcolor{black}{0.5900} \pm 0.16$} & \edit{$\textcolor{black}{0.5682} \pm 0.16$} & \edit{$\textcolor{black}{0.4939} \pm 0.20$} & \edit{$\textcolor{black}{0.7136} \pm 0.18$} & \edit{$\textcolor{black}{9.0028} \pm 6.10$} & \edit{$\textcolor{black}{155.8258} \pm 194.71$} & \edit{\underline{$\textcolor{black}{0.1025} \pm 0.04$}} \\
        
        (2 min, norm) & \edit{$\textcolor{black}{0.5800} \pm 0.17$} & \edit{$\textcolor{black}{0.6763} \pm 0.17$} & \edit{$\boldsymbol{\textcolor{black}{0.6347} \pm 0.16}$} & \edit{$\textcolor{black}{0.5802} \pm 0.17$} & \edit{\underline{$\textcolor{black}{0.5066} \pm 0.21$}} & \edit{$\textcolor{black}{0.7342} \pm 0.16$} & \edit{$\textcolor{black}{9.1154} \pm 6.94$} & \edit{$\textcolor{black}{154.9225} \pm 192.61$} & \edit{$\textcolor{black}{0.1038} \pm 0.05$} \\
        
        (2 min, no norm) & \edit{$\textcolor{black}{0.5671} \pm 0.17$} & \edit{$\textcolor{black}{0.6645} \pm 0.17$} & \edit{$\textcolor{black}{0.5951} \pm 0.17$} & \edit{$\textcolor{black}{0.5700} \pm 0.17$} & \edit{$\textcolor{black}{0.4994} \pm 0.20$} & \edit{$\textcolor{black}{0.7365} \pm 0.17$} & \edit{$\textcolor{black}{9.3761} \pm 7.05$} & \edit{$\textcolor{black}{169.5480} \pm 231.95$} & \edit{$\textcolor{black}{0.1060} \pm 0.05$} \\
        \bottomrule
    \end{tabular}%
    }
\end{table*}
\subsection{Experiment \edit{6}: Input Ablation Study} \label{sec:ablation}

\subsubsection{Background}\label{sec:background_exp4} To better understand the robustness and design choices of our foundation model, we conducted a series of studies on the input types. This is a common and essential practice when working with foundation models, as performance can be highly sensitive to implementation decisions\edit{~\cite{hong2024spectralgpt, das2024decoder}}. For instance, although we adopted a 4-minute PPG input window based on prior literature, other works have used shorter windows, which could impact temporal resolution and noise robustness. Similarly, we explore the role of person-specific normalization, a common preprocessing step in physiological signal modeling, to assess its contribution to model performance. These studies not only validate our design choices but also provide insights into how different parameter settings influence the quality and generalizability of the learned representations. 

\subsubsection{Experimental Design:} We segmented the PPG time series into non-overlapping 1-minute and 2-minute windows. First, we retrained Pulse-PPG using 1-minute windows with person-specific normalization ($z$-scoring). We then repeated the training using the same 1-minute windows, but without normalization. In both cases, we followed the same evaluation pipeline as used for the original 4-minute Pulse-PPG model. We applied the same process to 2-minute windows, retraining Pulse-PPG with and without normalization to evaluate the effect of window length and preprocessing on model performance.

\subsubsection{Results:}
We provide detailed results of these experiments in Table~\ref{tab:ablation_results}, including the original 4-minute Pulse-PPG model results for reference. For each model variant, we computed the mean performance across all downstream tasks using task-specific metrics and present our results in Table~\ref{tab:ablation_mean}.

Several key observations emerge from this study:
(1) Normalization consistently improves performance: for both 1-minute and 2-minute window settings, models trained with person-specific normalization outperform their non-normalized counterparts in most of the metrics.
(2) The original 4-minute normalized Pulse-PPG model still achieves the best overall performance, outperforming both the 1- and 2-minute normalized models across most metrics and always appears in the top 2 positions.
(3) The 1-minute normalized model slightly outperforms the 2-minute normalized model, indicating that performance does not necessarily scale linearly with window size.
(4) Both the 1-minute and 2-minute normalized models outperform PaPaGei when evaluated under the same downstream tasks. These findings suggest that while Pulse-PPG benefits from longer input windows and normalization, it remains robust even when using shorter input segments---a positive result for future research aiming at latency-sensitive or real-time applications.

\subsection{\edit{Loss Curve Visualization}}

\begin{figure}[h]
    \centering
    \includegraphics[width=\linewidth]{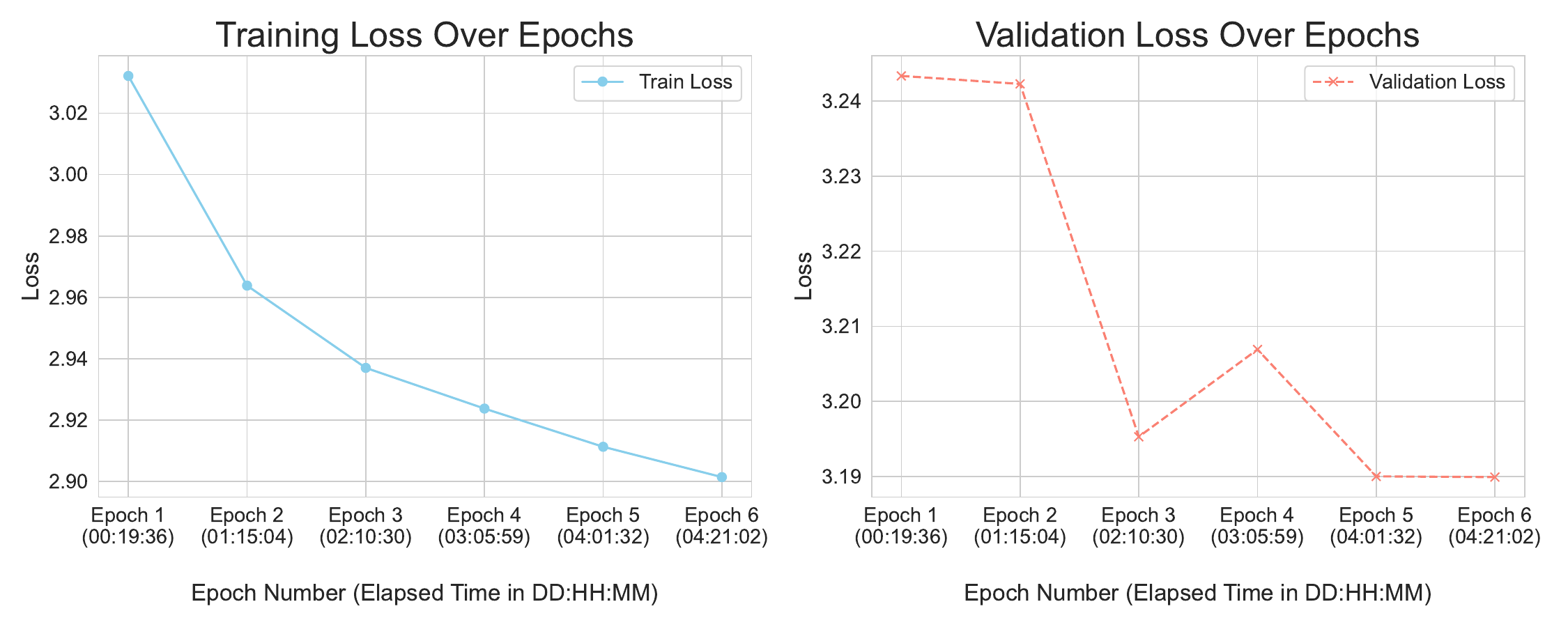}
    \Description{Convergence curve of the loss function over the training and validation sets, showing a steady decline in both, with the validation loss plateauing after epoch 5.}
    \caption{\edit{\textbf{Convergence curve of the loss function, over the training and validation sets}. The left plot shows that the training loss decreased monotonically, while the right plot shows the validation loss exhibited an overall downward trend with minor fluctuations over six epochs. Both losses converged steadily, with validation loss plateauing after epoch 5. See further discussion in Section \ref{sec:training}.}}
    \label{fig:losscurvel}
\end{figure}

\subsection{\edit{Full Result Tables for Experiments}}
\edit{In this section, we include the full results table of our experiments. Each table includes a short takeaway conclusion, with a reference to the relevant experiment sections for further detail discussion.}

\begin{table}[!htbp]
  \tiny
  \color{black}
  \captionsetup{labelfont={color=black},textfont={color=black}, font={small}}
  \caption{\edit{\textbf{Full Results of Pulse-PPG vs. PPG FM, General Time-series FM, and Supervised Baselines.} Mean performance in Table \ref{tab:baseline_general_mean}. Pulse-PPG achieves strongest consistent performance. See experiments in Section \ref{sec:exppapagei} and \ref{sec:expbaselinegeneralfoundation} for further discussion.}}
  \label{tab:baseline_general_TS}
  \resizebox{0.95\textwidth}{!}{
  \begin{tabular}{>{\raggedright}p{.08cm}>{\centering}p{2cm}>{\raggedright}p{.8cm}>{\raggedleft}p{.1cm}cccccc}
  \toprule
  & & & & \multicolumn{2}{c}{PPG FM} & \multicolumn{2}{c}{General Time-series FM} & \multicolumn{2}{c}{Supervised} \\
  \cmidrule(lr){5-6} \cmidrule(lr){7-8} \cmidrule(lr){9-10} & \multicolumn{3}{r}{{}} & \textbf{Pulse-PPG} & \textbf{PaPaGei} & \textbf{Chronos} & \textbf{MOMENT} & \textbf{ResNet-26} & \textbf{Rand Forest} \\
    \cmidrule(lr){5-5} \cmidrule(lr){6-6} \cmidrule(lr){7-7} \cmidrule(lr){8-8}  \cmidrule(lr){9-9} \cmidrule(lr){10-10}  & \multicolumn{3}{r}{\textbf{Eval Method:} } & \multicolumn{4}{c}{Linear Probe} & \multicolumn{2}{c}{-}  \\
    \cmidrule(lr){5-8} \cmidrule(lr){9-10}
  % \midrule
    
  \multirow{24}{*}{\textbf{\rotatebox{90}{\centering Wearable Lab}}}
  & \multirow{3}{*}{\centering\parbox{2.2cm}{\centering \textbf{Instant HR (R)} \\ PPG-DaLiA}}
  & MAE & \multirow{3}{*}{\rotatebox[origin=c]{-90}{\parbox{.3cm}{\rightarrowfill}}} & 8.936 & 15.22 & \edit{\underline{\textcolor{black}{8.170}}} & 7.508 & \textcolor{black}{\textbf{5.141}} & 15.95 \\
  & & MSE & & 149.3 & 338.0 & \edit{\underline{\textcolor{black}{122.9}}} & \textcolor{black}{\textbf{116.2}} & 126.7 & 366.9 \\
  & & MAPE & & 0.1166 & 0.2104 & 0.1089 & \edit{\underline{\textcolor{black}{0.0987}}} & \textcolor{black}{\textbf{0.0672}} & 0.2191 \\
  \addlinespace[-1pt]
  \cmidrule(lr){2-10}
  \addlinespace[-3pt]
  & \multirow{6}{*}{\centering \parbox{2cm}{\centering \textbf{Activity (9)} \\ PPG-DaLiA}}
  & F1 Score & \multirow{6}{*}{\rotatebox[origin=c]{90}{\parbox{.5cm}{\rightarrowfill}}} & 0.3039 & 0.2344 & \edit{\underline{\textcolor{black}{0.3866}}} & 0.3361 & \textcolor{black}{\textbf{0.3922}} & 0.2205 \\
  & & Accuracy & & 0.4104 & 0.3456 & \edit{\underline{\textcolor{black}{0.4500}}} & 0.4267 & \textcolor{black}{\textbf{0.4586}} & 0.3410 \\
  & & Precision & & 0.3882 & 0.2914 & \textcolor{black}{\textbf{0.4684}} & 0.4123 & \edit{\underline{\textcolor{black}{0.4543}}} & 0.2713 \\
  & & Recall & & 0.3107 & 0.2544 & \edit{\underline{\textcolor{black}{0.3724}}} & 0.3363 & \textcolor{black}{\textbf{0.3859}} & 0.2296 \\
  & & AUPRC & & 0.3504 & 0.2591 & \textcolor{black}{\textbf{0.4353}} & 0.3888 & \edit{\underline{\textcolor{black}{0.4230}}} & 0.2425 \\
  & & AUROC & & 0.8051 & 0.7237 & \textcolor{black}{\textbf{0.8497}} & 0.8263 & \edit{\underline{\textcolor{black}{0.8470}}} & 0.7146 \\
  \addlinespace[-1pt]
  \cmidrule(lr){2-10}
  \addlinespace[-3pt]
  & \multirow{6}{*}{\centering \parbox{2cm}{\centering \textbf{Stress (2)} \\ WESAD}}
  & F1 Score & \multirow{6}{*}{\rotatebox[origin=c]{90}{\parbox{.5cm}{\rightarrowfill}}} & \textcolor{black}{\textbf{0.8759}} & 0.6896 & 0.7972 & \edit{\underline{\textcolor{black}{0.8517}}} & 0.4065 & 0.3799 \\
  & & Accuracy & & \textcolor{black}{\textbf{0.8904}} & 0.7260 & 0.8082 & \edit{\underline{\textcolor{black}{0.8767}}} & 0.6849 & 0.5479 \\
  & & Precision & & \textcolor{black}{\textbf{0.8688}} & 0.6863 & 0.7932 & \edit{\underline{\textcolor{black}{0.8684}}} & 0.3425 & 0.3613 \\
  & & Recall & & \textcolor{black}{\textbf{0.8848}} & 0.6943 & 0.8365& \edit{\underline{\textcolor{black}{0.8396}}} & 0.5000 & 0.4117 \\
  & & AUPRC & & \textcolor{black}{\textbf{0.9410}} & 0.6465 & 0.7530 & \edit{\underline{\textcolor{black}{0.9196}}} & 0.2590 & 0.3173 \\
  & & AUROC & & \textcolor{black}{\textbf{0.9687}} & 0.8043 & 0.8878 & \edit{\underline{\textcolor{black}{0.9583}}} & 0.3974 & 0.5374 \\
  \addlinespace[-1pt]
  \cmidrule(lr){2-10}
  \addlinespace[-3pt]
  & \multirow{6}{*}{\centering \parbox{2cm}{\centering \textbf{Stress (4)} \\ WESAD}}
  & F1 Score & \multirow{6}{*}{\rotatebox[origin=c]{90}{\parbox{.5cm}{\rightarrowfill}}} & \edit{\underline{\textcolor{black}{0.5431}}} & 0.4010 & \textcolor{black}{\textbf{0.6398}} & 0.5118 & 0.1998 & 0.3540 \\
  & & Accuracy & & \edit{\underline{\textcolor{black}{0.6517}}} & 0.5169 & \textcolor{black}{\textbf{0.7416}} & 0.6180 & 0.4045 & 0.4719 \\
  & & Precision & & \edit{\underline{\textcolor{black}{0.6003}}} & 0.3947 & \textcolor{black}{\textbf{0.8022}} & 0.5767 & 0.1865 & 0.4274 \\
  & & Recall & & \edit{\underline{\textcolor{black}{0.5716}}} & 0.4079 & \textcolor{black}{\textbf{0.6390}} & 0.5252 & 0.2640 & 0.3573 \\
  & & AUPRC & & \edit{\underline{\textcolor{black}{0.5671}}} & 0.5202 & \textcolor{black}{\textbf{0.7443}} & 0.5304 & 0.2641 & 0.4811 \\
  & & AUROC & & \edit{\underline{\textcolor{black}{0.7807}}} & 0.7684 & \textcolor{black}{\textbf{0.8751}} & 0.7491 & 0.4662 & 0.7060 \\
 
  \midrule
   \addlinespace[-0.5pt]
 
  \multirow{12}{*}{\textbf{\rotatebox{90}{\centering Wearable Field}}}
  & \multirow{6}{*}{\centering \parbox{2cm}{\centering \textbf{Stress (2)} \\ MOODS}}
  & F1 Score & \multirow{6}{*}{\rotatebox[origin=c]{90}{\parbox{.5cm}{\rightarrowfill}}} & \textcolor{black}{\textbf{0.5398}} & 0.4398 & \edit{\underline{\textcolor{black}{0.5334}}} & 0.4308 & 0.5244 & 0.4725 \\
  & & Accuracy & & 0.6156 & \textcolor{black}{\textbf{0.6257}} & 0.6111 & 0.6167 & \edit{\underline{\textcolor{black}{0.6176}}} & 0.5998 \\
  & & Precision & & \edit{\underline{\textcolor{black}{0.5606}}} & 0.5538 & 0.5571 & 0.5131 & \textcolor{black}{\textbf{0.5609}} & 0.5139 \\
  & & Recall & & \textcolor{black}{\textbf{0.5460}} & 0.5114 & \edit{\underline{\textcolor{black}{0.5420}}} & 0.5030 & 0.5395 & 0.5071 \\
  & & AUPRC & & 0.4275 & 0.4030 & \edit{\underline{\textcolor{black}{0.4474}}} & 0.3883 & \textcolor{black}{\textbf{0.4480}} & 0.3929 \\
  & & AUROC & & 0.5691 & 0.5246 & \edit{\underline{\textcolor{black}{0.5810}}} & 0.5256 & \textcolor{black}{\textbf{0.5982}} & 0.5254 \\
  \addlinespace[-1pt]
  \cmidrule(lr){2-10}
  \addlinespace[-3pt]
  & \multirow{6}{*}{\centering\parbox{2cm}{\centering \textbf{Activity (2)} \\ MOODS}}
  & F1 Score & \multirow{6}{*}{\rotatebox[origin=c]{90}{\parbox{.5cm}{\rightarrowfill}}} & \edit{\underline{\textcolor{black}{0.6859}}} & 0.5318 & 0.6554 & 0.5701 & \textcolor{black}{\textbf{0.7903}} & 0.4621 \\
  & & Accuracy & & \edit{\underline{\textcolor{black}{0.8705}}} & 0.8458 & 0.8649 & 0.8531 & \textcolor{black}{\textbf{0.8991}} & 0.8428 \\
  & & Precision & & \edit{\underline{\textcolor{black}{0.7835}}} & 0.7058 & 0.7748 & 0.7604 & \textcolor{black}{\textbf{0.8261}} & 0.6543 \\
  & & Recall & & \edit{\underline{\textcolor{black}{0.6509}}} & 0.5365 & 0.6234 & 0.5601 & \textcolor{black}{\textbf{0.7649}} & 0.5019 \\
  & & AUPRC & & \edit{\underline{\textcolor{black}{0.5835}}} & 0.3410 & 0.5353 & 0.4225 & \textcolor{black}{\textbf{0.7339}} & 0.3036 \\
  & & AUROC & & \edit{\underline{\textcolor{black}{0.8722}}} & 0.6941 & 0.8349 & 0.7659 & \textcolor{black}{\textbf{0.9178}} & 0.6986 \\
 
  \midrule
    \addlinespace[-0.5pt]

  \multirow{21}{*}{\textbf{\rotatebox{90}{\centering Clinical}}}
  & \multirow{6}{*}{\centering\parbox{2cm}{\centering \textbf{Sleep Disturbance (2)} \\ SDB}}
  & F1 Score & \multirow{6}{*}{\rotatebox[origin=c]{90}{\parbox{.5cm}{\rightarrowfill}}} & 0.4630 & \textcolor{black}{\textbf{0.5588}} & 0.4852 & 0.4711 & 0.4770 & \edit{\underline{\textcolor{black}{0.4897}}} \\
  & & Accuracy & & 0.5364 & \textcolor{black}{\textbf{0.6609}} & 0.5456 & 0.5524 & 0.4790 & \edit{\underline{\textcolor{black}{0.5862}}} \\
  & & Precision & & 0.4633 & \textcolor{black}{\textbf{0.6079}} & 0.4856 & 0.4733 & \edit{\underline{\textcolor{black}{0.5063}}} & 0.5005 \\
  & & Recall & & 0.4673 & \textcolor{black}{\textbf{0.5672}} & 0.4865 & 0.4774 & \edit{\underline{\textcolor{black}{0.5067}}} & 0.5004 \\
  & & AUPRC & & 0.3085 & \textcolor{black}{\textbf{0.4236}} & 0.3459 & 0.3220 & 0.3090 & \edit{\underline{\textcolor{black}{0.3722}}} \\
  & & AUROC & & 0.4437 & \textcolor{black}{\textbf{0.5528}} & 0.4792 & 0.4624 & 0.4680 & \edit{\underline{\textcolor{black}{0.5009}}} \\
  \addlinespace[-1pt]
  \cmidrule(lr){2-10}
  \addlinespace[-3pt]
  & \multirow{6}{*}{\centering\parbox{2cm}{\centering \textbf{Hypertension (2)} \\ PPG-BP}}
  & F1 Score & \multirow{6}{*}{\rotatebox[origin=c]{90}{\parbox{.5cm}{\rightarrowfill}}} & \textcolor{black}{\textbf{0.6975}} & 0.6403 & 0.5939 & \edit{\underline{\textcolor{black}{0.6473}}} & 0.4459 & 0.5277 \\
  & & Accuracy & & \textcolor{black}{\textbf{0.8130}} & 0.7317 & 0.7073 & 0.7398 & \edit{\underline{\textcolor{black}{0.8049}}} & 0.7480 \\
  & & Precision & & \textcolor{black}{\textbf{0.7009}} & 0.6311 & 0.5887 & \edit{\underline{\textcolor{black}{0.637}}} & 0.4024 & 0.5407 \\
  & & Recall & & \textcolor{black}{\textbf{0.6944}} & 0.6755 & 0.6130 & \edit{\underline{\textcolor{black}{0.6806}}} & 0.5000 & 0.5278 \\
  & & AUPRC & & 0.4266 & 0.3214 & 0.3711 & \textcolor{black}{\textbf{0.4601}} & 0.3803 & 0.2201 \\
  & & AUROC & & \textcolor{black}{\textbf{0.7971}} & 0.7201 & 0.6229 & \edit{\underline{\textcolor{black}{0.7811}}} & 0.6620 & 0.5011 \\
  \addlinespace[-1pt]
  \cmidrule(lr){2-10}
  \addlinespace[-3pt]
  & \multirow{3}{*}{\centering\parbox{2cm}{\centering \textbf{Systolic BP (R)} \\ PPG-BP}}
  & MAE & \multirow{3}{*}{\rotatebox[origin=c]{-90}{\parbox{.3cm}{\rightarrowfill}}} & \edit{\underline{\textcolor{black}{13.62}}} & 13.99 & \textcolor{black}{\textbf{13.26}} & 14.36 & 31.65 & 17.28 \\
  & & MSE & & \edit{\underline{\textcolor{black}{286.5}}} & 321.1 & \textcolor{black}{\textbf{283.4}} & 348.7 & 1458 & 466.7 \\
  & & MAPE & & \edit{\underline{\textcolor{black}{0.1065}}} & 0.1096 & \textcolor{black}{\textbf{0.1052}} & 0.113 & 0.229 & 0.136 \\
  \addlinespace[-1pt]
  \cmidrule(lr){2-10}
  \addlinespace[-3pt]
  & \multirow{3}{*}{\centering\parbox{2cm}{\centering \textbf{Diastolic BP (R)} \\ PPG-BP}}
  & MAE & \multirow{3}{*}{\rotatebox[origin=c]{-90}{\parbox{.3cm}{\rightarrowfill}}} & \textcolor{black}{\textbf{8.878}} & \edit{\underline{\textcolor{black}{9.246}}} & 9.588 & 9.447 & 11.21 & 9.904 \\
  & & MSE & & \textcolor{black}{\textbf{118.2}} & \edit{\underline{\textcolor{black}{127.0}}} & 142.1 & 130.2 & 245.8 & 150.9 \\
  & & MAPE & & \textcolor{black}{\textbf{0.1232}} & \edit{\underline{\textcolor{black}{0.1290}}} & 0.1328 & 0.1325 & 0.1387 & 0.1379 \\
  \addlinespace[-1pt]
  \cmidrule(lr){2-10}
  \addlinespace[-3pt]
  & \multirow{3}{*}{\centering\parbox{2cm}{\centering \textbf{Average HR (R)} \\ PPG-BP}}
  & MAE & \multirow{3}{*}{\rotatebox[origin=c]{-90}{\parbox{.3cm}{\rightarrowfill}}} & \textcolor{black}{\textbf{4.003}} & 4.549 & 4.648 & \edit{\underline{\textcolor{black}{4.087}}} & 21.09 & 7.127 \\
  & & MSE & & \textcolor{black}{\textbf{27.02}} & 32.59 & 34.99 & \edit{\underline{\textcolor{black}{28.50}}} & 491.1 & 88.42 \\
  & & MAPE & & \textcolor{black}{\textbf{0.0573}} & 0.0662 & 0.0669 & \edit{\underline{\textcolor{black}{0.0586}}} & 0.2862 & 0.1048 \\
  \bottomrule
  \end{tabular}%
  }
\end{table}

\begin{table}[!htbp]
    \tiny 
     	\caption{\edit{\textbf{Full Results of Pulse-PPG vs. PaPaGei vs. Light Pulse-PPG across Wearable Field PPG vs. Clinical Clean PPG pre-training data.} Mean performance in Table   \ref{tab:pulseppg_vs_papagei_mean}, \ref{tab:light_pulse_ppg_mean}. Pulse-PPG achieves stronger performance than PaPaGei, a prior PPG FM, even when controlling for parameters (i.e. Light Pulse-PPG). Surprisingly, pre-training on wearable field data achieves stronger generalizable performance on Wearable and Clinical settings. See Sections \ref{sec:exppapagei}, \ref{sec:expfieldtolab}, \ref{sec:light_pulse_ppg_vs_papagei} for further detail.}}
    % \caption{Comparing \textbf{Pulse-PPG and \edit{Light Pulse-PPG} vs. PaPaGei} and \textbf{Wearable Field PPG vs. Clinical Clean PPG} pre-training data.} 
    \label{tab:exp23}
    \resizebox{0.85\textwidth}{!}{%
    \begin{tabular}{>{\raggedright}p{.05cm}>{\centering}p{2cm}>{\raggedright}p{.8cm}>{\raggedleft}p{.1cm}ccccc} 
    
    \toprule
     &  \multicolumn{3}{r}{{}} & \multicolumn{2}{c}{\textcolor{black}{\textbf{Pulse-PPG}}} & \multicolumn{2}{c}{\textbf{PaPaGei}} & \textbf{\edit{Light Pulse-PPG}} \\
    \cmidrule(lr){5-6} \cmidrule(lr){7-8} \cmidrule(lr){9-9} 
    & \multicolumn{3}{r}{\textbf{Pre-train Data:}} & Wearable Field & Clinical & Wearable Field & Clinical & \edit{Wearable Field} \\
        \cmidrule(lr){5-5} \cmidrule(lr){6-6} \cmidrule(lr){7-7} \cmidrule(lr){8-8} \cmidrule(lr){9-9} 
        % \cmidrule(lr){5-9}
    & \multicolumn{3}{r}{\textbf{Eval Method:} } & \multicolumn{5}{c}{Linear Probe} \\
    \cmidrule(lr){5-9}
    \multirow{21}{*}{\textbf{\rotatebox{90}{\centering Wearable Lab}}}
    & \multirow{3}{*}{\centering\parbox{2.2cm}{\centering \textbf{Instant HR (R)} \\ PPG-DaLiA}}
    & MAE  & \multirow{3}{*}{\rotatebox[origin=c]{-90}{\parbox{.3cm}{\rightarrowfill}}} & \textbf{8.936} & 18.60 & \edit{12.08} & \edit{15.22} & \edit{\underline{10.21}} \\
    & & MSE  &  & \textbf{149.3} & 498.2 & \edit{234.51} & \edit{338.0} & \edit{\underline{180.1}} \\
    & & MAPE  & & \textbf{0.1166} & 0.2581 & \edit{0.1597} & \edit{0.2104} & \edit{\underline{0.1324}} \\
    \addlinespace[-1pt]
    \cmidrule(lr){2-9}
    \addlinespace[-3pt]
    & \multirow{6}{*}{\centering \parbox{2cm}{\centering \textbf{Activity (9)} \\ PPG-DaLiA}}
    & F1 Score  & \multirow{6}{*}{\rotatebox[origin=c]{90}{\parbox{.5cm}{\rightarrowfill}}} & \textbf{0.3039} & 0.2250 & \edit{0.2796} & \edit{0.2344} & \edit{\underline{0.2956}} \\
    & & Accuracy  &  & \edit{\underline{\textcolor{black}{0.4104}}} & 0.3276 & \edit{0.4010} & \edit{0.3456} & \edit{\textbf{0.4112}} \\
    & & Precision & & \textbf{0.3882} & 0.2631 & \edit{0.3690} & \edit{0.2914} & \edit{\underline{0.3856}} \\
    & & Recall  & & \textbf{0.3107} & 0.2457 & \edit{0.2987} & \edit{0.2544} & \edit{\underline{0.3087}} \\
    & & AUPRC  & & \textbf{0.3504} & 0.2527 & \edit{0.3236} & \edit{0.2591} & \edit{\underline{0.3438}} \\
    & & AUROC  & & \textbf{0.8051} & 0.7741 & \edit{0.7716} & \edit{0.7237} & \edit{\underline{0.7936}} \\
    \addlinespace[-1pt]
    \cmidrule(lr){2-9}
    \addlinespace[-3pt]
    & \multirow{6}{*}{\centering \parbox{2cm}{\centering \textbf{Stress (2)} \\ WESAD}}
    & F1 Score  & \multirow{6}{*}{\rotatebox[origin=c]{90}{\parbox{.5cm}{\rightarrowfill}}} & \textbf{0.8759} & 0.8009 & 0.8237 & 0.6896 & \edit{\underline{0.8698}} \\
    & & Accuracy  & & \textbf{0.8904} & 0.8493 & 0.8356 & 0.7260 & \edit{\textbf{0.8904}} \\
    & & Precision & & 0.8688 & \edit{\underline{\textcolor{black}{0.8795}}} & 0.8149 & 0.6863 & \edit{\textbf{0.8804}} \\
    & & Recall  & & \textbf{0.8848} & 0.7726 & 0.8565 & 0.6943 & \edit{\underline{0.8613}} \\
    & & AUPRC  & & \textbf{0.9410} & 0.8724 & 0.7518 & 0.6465 & \edit{\underline{0.8764}} \\
    & & AUROC  & & \textbf{0.9687} & 0.9374 & 0.9130 & 0.8043 & \edit{\underline{0.9504}} \\
    \addlinespace[-1pt]
    \cmidrule(lr){2-9}
    \addlinespace[-3pt]
    & \multirow{6}{*}{\centering \parbox{2cm}{\centering \textbf{Stress (4)} \\ WESAD}}
    & F1 Score  & \multirow{6}{*}{\rotatebox[origin=c]{90}{\parbox{.5cm}{\rightarrowfill}}} & \textbf{0.5431} & \edit{\underline{\textcolor{black}{0.5126}}} & 0.4976 & 0.4010 & \edit{0.4596} \\
    & & Accuracy  & & \edit{\underline{\textcolor{black}{0.6517}}} & \textbf{0.6629} & 0.5393 & 0.5169 & \edit{0.5730} \\
    & & Precision & & \textbf{0.6003} & \edit{\underline{\textcolor{black}{0.5124}}} & 0.4976 & 0.3947 & \edit{0.4773} \\
    & & Recall  & & \textbf{0.5716} & \edit{\underline{\textcolor{black}{0.5387}}} & 0.5258 & 0.4079 & \edit{0.4722} \\
    & & AUPRC  & & \edit{\underline{\textcolor{black}{0.5671}}} & \textbf{0.6931} & 0.4813 & 0.5202 & \edit{0.4667} \\
    & & AUROC  & & \edit{\underline{\textcolor{black}{0.7807}}} & \textbf{0.8456} & 0.7257 & 0.7684 & \edit{0.6655} \\
    
    \midrule
    \addlinespace[-0.5pt]
    \multirow{12}{*}{\textbf{\rotatebox{90}{\centering Wearable Field}}}
    & \multirow{6}{*}{\centering \parbox{2cm}{\centering \textbf{Stress (2)} \\ MOODS}}
    & F1 Score  & \multirow{6}{*}{\rotatebox[origin=c]{90}{\parbox{.5cm}{\rightarrowfill}}} & \textbf{0.5398} & \edit{\underline{\textcolor{black}{0.4550}}} & 0.4141 & 0.4398 & \edit{0.4383} \\
    & & Accuracy  & & 0.6156 & 0.6135 & 0.6251 & \edit{\underline{\textcolor{black}{0.6257}}} & \edit{\textbf{0.6342}} \\
    & & Precision &  & \edit{\underline{\textcolor{black}{0.5606}}} & 0.5244 & 0.5355 & 0.5538 & \edit{\textbf{0.5709}} \\
    & & Recall  & & \textbf{0.5460} & 0.5085 & 0.5042 & 0.5114 & \edit{\underline{0.5127}} \\
    & & AUPRC  & & \textbf{0.4275} & 0.3960 & 0.3923 & 0.4030 & \edit{\underline{0.4065}} \\
    & & AUROC  & & \textbf{0.5691} & \edit{\underline{\textcolor{black}{0.5382}}} & 0.5275 & 0.5246 & \edit{0.5341} \\
    \addlinespace[-1pt]
    \cmidrule(lr){2-9}
    \addlinespace[-3pt]
    & \multirow{6}{*}{\centering\parbox{2cm}{\centering \textbf{Activity (2)} \\ MOODS}}
    & F1 Score  & \multirow{6}{*}{\rotatebox[origin=c]{90}{\parbox{.5cm}{\rightarrowfill}}} & \textbf{0.6859} & \edit{\underline{\textcolor{black}{0.5641}}} & 0.5375 & 0.5318 & \edit{0.5320} \\
    & & Accuracy  & & \textbf{0.8705} & \edit{\underline{\textcolor{black}{0.8508}}} & 0.8483 & 0.8458 & \edit{0.8468} \\
    & & Precision &  & \textbf{0.7835} & \edit{\underline{\textcolor{black}{0.7401}}} & 0.7349 & 0.7058 & \edit{0.7188} \\
    & & Recall  & & \textbf{0.6509} & \edit{\underline{\textcolor{black}{0.5562}}} & 0.5401 & 0.5365 & \edit{0.5368} \\
    & & AUPRC  & & \textbf{0.5835} & \edit{\underline{\textcolor{black}{0.3916}}} & 0.3673 & 0.3410 & \edit{0.3517} \\
    & & AUROC  & & \textbf{0.8722} & 0.7205 & 0.7302 & 0.6941 & \edit{\underline{0.7328}} \\
    
    \midrule
    \addlinespace[-0.5pt]
    \multirow{21}{*}{\textbf{\rotatebox{90}{\centering Clinical}}} 
    & \multirow{6}{*}{\centering\parbox{2cm}{\centering \textbf{Sleep Disturbance (2)} \\ SDB}} 
    & F1 Score  & \multirow{6}{*}{\rotatebox[origin=c]{90}{\parbox{.5cm}{\rightarrowfill}}} & 0.4630 & 0.5005 & \edit{\underline{\textcolor{black}{0.5012}}} & \textbf{0.5588} & \edit{0.4723} \\
    & & Accuracy  & & 0.5364 & \edit{\underline{\textcolor{black}{0.5898}}} & 0.5808 & \textbf{0.6609} & \edit{0.5881} \\
    & & Precision & & 0.4633 & \edit{\underline{\textcolor{black}{0.5108}}} & 0.5075 & \textbf{0.6079} & \edit{0.4867} \\
    & & Recall  & & 0.4673 & \edit{\underline{\textcolor{black}{0.5084}}} & 0.5062 & \textbf{0.5672} & \edit{0.4912} \\
    & & AUPRC  & & 0.3085 & \edit{\underline{\textcolor{black}{0.3636}}} & 0.3577 & \textbf{0.4236} & \edit{0.3371} \\
    & & AUROC  & & 0.4437 & \edit{\underline{\textcolor{black}{0.5122}}} & 0.5088 & \textbf{0.5528} & \edit{0.4964} \\
    \addlinespace[-1pt]
    \cmidrule(lr){2-9}
    \addlinespace[-3pt]
    & \multirow{6}{*}{\centering\parbox{2cm}{\centering \textbf{Hypertension (2)} \\ PPG-BP}} 
    & F1 Score  & \multirow{6}{*}{\rotatebox[origin=c]{90}{\parbox{.5cm}{\rightarrowfill}}} & \textbf{0.6975} & 0.5939 & 0.5839 & \edit{\underline{\textcolor{black}{0.6403}}} & \edit{0.5756} \\
    & & Accuracy  & & \textbf{0.8130} & 0.7073 & 0.6829 & 0.7317 & \edit{\underline{0.7642}} \\
    & & Precision & & \textbf{0.7009} & 0.5887 & 0.5824 & \edit{\underline{\textcolor{black}{0.6311}}} & \edit{0.5915} \\
    & & Recall  & & \textbf{0.6944} & 0.6130 & 0.6136 & \edit{\underline{\textcolor{black}{0.6755}}} & \edit{0.5694} \\
    & & AUPRC  & & \edit{\underline{\textcolor{black}{0.4266}}} & 0.3051 & 0.3413 & 0.3214 & \edit{\textbf{0.4465}} \\
    & & AUROC  & & \textbf{0.7971} & 0.6936 & 0.7100 & 0.7201 & \edit{\underline{0.7251}} \\
    \addlinespace[-1pt]
    \cmidrule(lr){2-9}
    \addlinespace[-3pt]
    & \multirow{3}{*}{\centering\parbox{2cm}{\centering \textbf{Systolic BP (R)} \\ PPG-BP}} 
    & MAE  & \multirow{3}{*}{\rotatebox[origin=c]{-90}{\parbox{.3cm}{\rightarrowfill}}} & \textbf{13.62} & 14.60 & 14.49 & 13.99 & \edit{\underline{13.70}} \\
    & & MSE  & & \textbf{286.5} & 339.0 & 338.2 & \edit{\underline{\textcolor{black}{321.1}}} & \edit{331.5} \\
    & & MAPE  & & \edit{\underline{\textcolor{black}{0.1065}}} & 0.1138 & 0.1131 & 0.1096 & \edit{\textbf{0.1050}} \\
    \addlinespace[-1pt]
    \cmidrule(lr){2-9}
    \addlinespace[-3pt]
    & \multirow{3}{*}{\centering\parbox{2cm}{\centering \textbf{Diastolic BP (R)} \\ PPG-BP}} 
    & MAE  & \multirow{3}{*}{\rotatebox[origin=c]{-90}{\parbox{.3cm}{\rightarrowfill}}} & \textbf{8.878} & 9.415 & 9.848 & \edit{\underline{\textcolor{black}{9.246}}} & \edit{9.510} \\
    & & MSE  & & \textbf{118.2} & 136.7 & 135.3 & \edit{\underline{\textcolor{black}{127.0}}} & \edit{141.3} \\
    & & MAPE  & & \textbf{0.1232} & 0.1299 & 0.1374 & \edit{\underline{\textcolor{black}{0.1290}}} & \edit{0.1302} \\
    \addlinespace[-1pt]
    \cmidrule(lr){2-9}
    \addlinespace[-3pt]
    & \multirow{3}{*}{\centering\parbox{2cm}{\centering \textbf{Average HR (R)} \\ PPG-BP}} 
    & MAE  & \multirow{3}{*}{\rotatebox[origin=c]{-90}{\parbox{.3cm}{\rightarrowfill}}} & \edit{\underline{\textcolor{black}{
4.003}}} & 4.258 & \textbf{3.971} & 4.549 & \edit{4.554} \\
    & & MSE  & & \edit{\underline{\textcolor{black}{
27.02}}} & 27.60 & \textbf{25.74} & 32.59 & \edit{32.07} \\
    & & MAPE  & & \textbf{0.0573} & 0.0607 & \edit{\underline{\textcolor{black}{
0.0575}}} & 0.0662 & \edit{0.0652} \\
    \bottomrule
    \end{tabular}%
   } 
\end{table}

\begin{table}[!htbp]
    \tiny
    \color{black}
    \captionsetup{labelfont={color=black},textfont={color=black}}
    % \caption{\textbf{Ablation Study Results for Pulse-PPG}} 
    \caption{\edit{\textbf{Full Results of Ablation Study on Input Window Size and Z-normalization Status.} Mean performance in Table \ref{tab:ablation_mean}. 4 minutes is the best, and normalizing improves performance. See Section \ref{sec:ablation} for more.}}
    \label{tab:ablation_results}
    \resizebox{0.95\textwidth}{!}{
    \begin{tabular}{>{\raggedright}p{.08cm}>{\centering}p{2cm}>{\raggedright}p{.8cm}>{\raggedleft}p{.1cm}ccccc} 
    \toprule
     &  &  & & \multicolumn{5}{c}{\textbf{Pulse-PPG Model Variants}} \\
         \cmidrule(lr){5-9}
    & \multicolumn{3}{r}{\textbf{Pre-Train Data:}} & \multicolumn{5}{c}{Wearable Field} \\
    \cmidrule(lr){5-9}
    & \multicolumn{3}{r}{\textbf{Eval Method:}} & \multicolumn{5}{c}{Linear Probe} \\
    \cmidrule(lr){5-9}
             & \multicolumn{3}{r}{\textbf{(Window Size, Z-Norm Status):}} & (4 min, norm) & (1 min, norm) & (1 min, no norm) & (2 min, norm) & (2 min, no norm) \\     \cmidrule(lr){5-5} \cmidrule(lr){6-6} \cmidrule(lr){7-7} \cmidrule(lr){8-8} \cmidrule(lr){9-9}
    % \midrule
    % \\
    \multirow{24}{*}{\textbf{\rotatebox{90}{\centering Wearable Lab}}}
    & \multirow{3}{*}{\centering\parbox{2.2cm}{\centering \textbf{Instant HR (R)} \\ PPG-DaLiA}}
    & MAE  & \multirow{3}{*}{\rotatebox[origin=c]{-90}{\parbox{.3cm}{\rightarrowfill}}} & 8.936 & 8.717 & \textcolor{black}{\textbf{8.303}} & 8.883 & \edit{\underline{\textcolor{black}{8.582}}} \\
    & & MSE  &  & 149.3 & 137.7 & \textcolor{black}{\textbf{131.1}} & 144.5 & \edit{\underline{\textcolor{black}{136.9}}} \\
    & & MAPE  & & 0.1166 & 0.117 & \textcolor{black}{\textbf{0.1086}} & 0.1171 & \edit{\underline{\textcolor{black}{0.1122}}} \\
    \addlinespace[-1pt]
    \cmidrule(lr){2-9}
    \addlinespace[-3pt]
    & \multirow{6}{*}{\centering \parbox{2cm}{\centering \textbf{Activity (9)} \\ PPG-DaLiA}}
    & F1 Score  & \multirow{6}{*}{\rotatebox[origin=c]{90}{\parbox{.5cm}{\rightarrowfill}}} & 0.3039 & 0.3212 & \textcolor{black}{\textbf{0.3423}} & 0.3269 & \edit{\underline{\textcolor{black}{0.3270}}} \\
    & & Accuracy  &  & 0.4104 & 0.4144 & \textcolor{black}{\textbf{0.4323}} & \edit{\underline{\textcolor{black}{0.4211}}} & 0.4206 \\
    & & Precision & & 0.3882 & 0.4034 & \textcolor{black}{\textbf{0.4270}} & \edit{\underline{\textcolor{black}{0.4091}}} & 0.4035 \\
    & & Recall  & & 0.3107 & 0.3226 & \textcolor{black}{\textbf{0.3459}} & 0.3289 & \edit{\underline{\textcolor{black}{0.3300}}} \\
    & & AUPRC  & & 0.3504 & 0.3687 & \textcolor{black}{\textbf{0.3907}} & 0.3602 & \edit{\underline{\textcolor{black}{0.3773}}} \\
    & & AUROC  & & 0.8051 & 0.8144 & \textcolor{black}{\textbf{0.8269}} & 0.8072 & \edit{\underline{\textcolor{black}{0.8200}}} \\
    \addlinespace[-1pt]
    \cmidrule(lr){2-9}
    \addlinespace[-3pt]
    & \multirow{6}{*}{\centering \parbox{2cm}{\centering \textbf{Stress (2)} \\ WESAD}}
    & F1 Score  & \multirow{6}{*}{\rotatebox[origin=c]{90}{\parbox{.5cm}{\rightarrowfill}}} & 0.8759 & \edit{\underline{\textcolor{black}{0.9087}}} & 0.8643 & \textcolor{black}{\textbf{0.9215}} & 0.9068 \\
    & & Accuracy  & & 0.8904 & \edit{\underline{\textcolor{black}{0.9178}}} & 0.8767 & \textcolor{black}{\textbf{0.9315}} & \edit{\underline{\textcolor{black}{0.9178}}} \\
    & & Precision & & 0.8688 & 0.8965 & 0.8527 & \textcolor{black}{\textbf{0.9170}} & \edit{\underline{\textcolor{black}{0.8991}}} \\
    & & Recall  & & 0.8848 & \textcolor{black}{\textbf{0.9282}} & 0.8865 & \edit{\underline{\textcolor{black}{0.9265}}} & 0.9165 \\
    & & AUPRC  & & \edit{\underline{\textcolor{black}{0.9410}}} & 0.8836 & 0.9106 & \textcolor{black}{\textbf{0.9658}} & 0.9409 \\
    & & AUROC  & & 0.9687 & 0.9530 & 0.9539 & \textcolor{black}{\textbf{0.9817}} & \edit{\underline{\textcolor{black}{0.9739}}} \\
    \addlinespace[-1pt]
    \cmidrule(lr){2-9}
    \addlinespace[-2pt]
    & \multirow{6}{*}{\centering \parbox{2cm}{\centering \textbf{Stress (4)} \\ WESAD}}
    & F1 Score  & \multirow{6}{*}{\rotatebox[origin=c]{90}{\parbox{.5cm}{\rightarrowfill}}} & \textcolor{black}{\textbf{0.5431}} & \edit{\underline{\textcolor{black}{0.5225}}} & 0.5208 & 0.5028 & 0.4603 \\
    & & Accuracy  & & \textcolor{black}{\textbf{0.6517}} & \edit{\underline{\textcolor{black}{0.6516}}} & 0.5955 & 0.6067 & 0.5617 \\
    & & Precision & & \edit{\underline{\textcolor{black}{0.6003}}} & 0.5883 & 0.5182 & \textcolor{black}{\textbf{0.6895}} & 0.4599 \\
    & & Recall  & & \textcolor{black}{\textbf{0.5716}} & \edit{\underline{\textcolor{black}{0.5397}}} & 0.5396 & 0.5138 & 0.4790 \\
    & & AUPRC  & & \edit{\underline{\textcolor{black}{0.5671}}} & 0.5601 & 0.5635 & \textcolor{black}{\textbf{0.5893}} & 0.5620 \\
    & & AUROC  & & \edit{\underline{\textcolor{black}{0.7807}}} & 0.7663 & 0.7599 & 0.7677 & \textcolor{black}{\textbf{0.7961}} \\

    \midrule
    \addlinespace[-.5 pt]
    
    \multirow{12}{*}{\textbf{\rotatebox{90}{\centering Wearable Field}}}
    & \multirow{6}{*}{\centering \parbox{2cm}{\centering \textbf{Stress (2)} \\ MOODS}}
    & F1 Score  & \multirow{6}{*}{\rotatebox[origin=c]{90}{\parbox{.5cm}{\rightarrowfill}}} & 0.5398 & \textcolor{black}{\textbf{0.5458}} & \edit{\underline{\textcolor{black}{0.5447}}} & 0.5437 & 0.5147 \\
    & & Accuracy  & & \edit{\underline{\textcolor{black}{0.6156}}} & \textcolor{black}{\textbf{0.6161}} & 0.6108 & 0.6135 & 0.6004 \\
    & & Precision &  & \edit{\underline{\textcolor{black}{0.5606}}} & \textcolor{black}{\textbf{0.5637}} & 0.5590 & 0.5605 & 0.5352 \\
    & & Recall  & & 0.5460 & \textcolor{black}{\textbf{0.5501}} & \edit{\underline{\textcolor{black}{0.5480}}} & \edit{\underline{\textcolor{black}{0.5480}}} & 0.5255 \\
    & & AUPRC  & & 0.4275 & \textcolor{black}{\textbf{0.4457}} & 0.4289 & \edit{\underline{\textcolor{black}{0.4373}}} & 0.4098 \\
    & & AUROC  & & 0.5691 & \textcolor{black}{\textbf{0.5812}} & 0.5657 & \edit{\underline{\textcolor{black}{0.5717}}} & 0.5556 \\    
    \addlinespace[-1pt]
    \cmidrule(lr){2-9}
    \addlinespace[-3pt]
    & \multirow{6}{*}{\centering\parbox{2cm}{\centering \textbf{Activity (2)} \\ MOODS}}
    & F1 Score  & \multirow{6}{*}{\rotatebox[origin=c]{90}{\parbox{.5cm}{\rightarrowfill}}} & \textcolor{black}{\textbf{0.6859}} & \edit{\underline{\textcolor{black}{0.6785}}} & 0.6781 & 0.6757 & 0.6668 \\
    & & Accuracy  & & \textcolor{black}{\textbf{0.8705}} & 0.8689 & \edit{\underline{\textcolor{black}{0.8701}}} & 0.8684 & 0.8672 \\
    & & Precision &  & \edit{\underline{\textcolor{black}{0.7835}}} & 0.7809 & \textcolor{black}{\textbf{0.7883}} & 0.7806 & 0.7802 \\
    & & Recall  & & \textcolor{black}{\textbf{0.6509}} & \edit{\underline{\textcolor{black}{0.6440}}} & 0.6427 & 0.6414 & 0.6331 \\
    & & AUPRC  & & \textcolor{black}{\textbf{0.5835}} & \edit{\underline{\textcolor{black}{0.5692}}} & 0.5685 & 0.5674 & 0.5570 \\
    & & AUROC  & & \textcolor{black}{\textbf{0.8722}} & \edit{\underline{\textcolor{black}{0.8690}}} & 0.8649 & 0.8680 & 0.8646 \\
    \midrule

    \addlinespace[-3pt]
    \multirow{21}{*}{\textbf{\rotatebox{90}{\centering Clinical}}} 
    & \multirow{6}{*}{\centering\parbox{2cm}{\centering \textbf{Sleep Disturbance (2)} \\ SDB}} 
    & F1 Score  & \multirow{6}{*}{\rotatebox[origin=c]{90}{\parbox{.5cm}{\rightarrowfill}}} & \edit{\underline{\textcolor{black}{0.4630}}} & 0.4499 & 0.4273 & \textcolor{black}{\textbf{0.4861}} & 0.4600 \\
    & & Accuracy  & & \edit{\underline{\textcolor{black}{0.5364}}} & 0.5166 & 0.5016 & \textcolor{black}{\textbf{0.5614}} & 0.5275 \\
    & & Precision & & \edit{\underline{\textcolor{black}{0.4633}}} & 0.4491 & 0.4254 & \textcolor{black}{\textbf{0.4889}} & 0.4597 \\
    & & Recall  & & \edit{\underline{\textcolor{black}{0.4673}}} & 0.4526 & 0.4321 & \textcolor{black}{\textbf{0.4904}} & 0.4629 \\
    & & AUPRC  & & 0.3085 & 0.3033 & 0.2870 & \textcolor{black}{\textbf{0.3313}} & \edit{\underline{\textcolor{black}{0.3092}}} \\
    & & AUROC  & & 0.4437 & 0.4336 & 0.3933 & \textcolor{black}{\textbf{0.4772}} & \edit{\underline{\textcolor{black}{0.4470}}} \\
    \addlinespace[-1pt]
    \cmidrule(lr){2-9}
    \addlinespace[-3pt]
    & \multirow{6}{*}{\centering\parbox{2cm}{\centering \textbf{Hypertension (2)} \\ PPG-BP}} 
    & F1 Score  & \multirow{6}{*}{\rotatebox[origin=c]{90}{\parbox{.5cm}{\rightarrowfill}}} & \textcolor{black}{\textbf{0.6975}} & \edit{\underline{\textcolor{black}{0.6688}}} & 0.5564 & 0.6031 & 0.6342 \\
    & & Accuracy  & & \textcolor{black}{\textbf{0.8130}} & \edit{\underline{\textcolor{black}{0.7642}}} & 0.6585 & 0.7317 & 0.7560 \\
    & & Precision & & \textcolor{black}{\textbf{0.7009}} & \edit{\underline{\textcolor{black}{0.6565}}} & 0.5591 & 0.5979 & 0.6280 \\
    & & Recall  & & \edit{\underline{\textcolor{black}{0.6944}}} & \textcolor{black}{\textbf{0.6957}} & 0.5827 & 0.6123 & 0.6433 \\
    & & AUPRC  & & \textcolor{black}{\textbf{0.4266}} & \edit{\underline{\textcolor{black}{0.4154}}} & 0.3072 & 0.2950 & 0.3398 \\
    & & AUROC  & & \textcolor{black}{\textbf{0.7971}} & \edit{\underline{\textcolor{black}{0.7912}}} & 0.6304 & 0.6658 & 0.6982 \\
    \addlinespace[-1pt]
    \cmidrule(lr){2-9}
    \addlinespace[-3pt]
    & \multirow{3}{*}{\centering\parbox{2cm}{\centering \textbf{Systolic BP (R)} \\ PPG-BP}} 
    & MAE  & \multirow{3}{*}{\rotatebox[origin=c]{-90}{\parbox{.3cm}{\rightarrowfill}}} & \textcolor{black}{\textbf{13.62}} & \edit{\underline{\textcolor{black}{13.69}}} & 13.97 & 14.53 & 15.06 \\
    & & MSE  & & \textcolor{black}{\textbf{286.5}} & \edit{\underline{\textcolor{black}{298.8}}} & 327.0 & 318.1 & 374.5 \\
    & & MAPE  & & \edit{\underline{\textcolor{black}{0.1065}}} & \textcolor{black}{\textbf{0.1062}} & 0.1080 & 0.1135 & 0.1166 \\
    \addlinespace[-1pt]
    \cmidrule(lr){2-9}
    \addlinespace[-3pt]
    & \multirow{3}{*}{\centering\parbox{2cm}{\centering \textbf{Diastolic BP (R)} \\ PPG-BP}} 
    & MAE  & \multirow{3}{*}{\rotatebox[origin=c]{-90}{\parbox{.3cm}{\rightarrowfill}}} & \textcolor{black}{\textbf{8.878}} & 9.169 & \edit{\underline{\textcolor{black}{9.068}}} & 9.193 & 9.573 \\
    & & MSE  & & \textcolor{black}{\textbf{118.2}} & \edit{\underline{\textcolor{black}{128.1}}} & 128.7 & 131.0 & 137.2 \\
    & & MAPE  & & \textcolor{black}{\textbf{0.1232}} & 0.1286 & \edit{\underline{\textcolor{black}{0.1259}}} & 0.1288 & 0.1326 \\
    \addlinespace[-1pt]
    \cmidrule(lr){2-9}
    \addlinespace[-3pt]
    & \multirow{3}{*}{\centering\parbox{2cm}{\centering \textbf{Average HR (R)} \\ PPG-BP}} 
    & MAE  & \multirow{3}{*}{\rotatebox[origin=c]{-90}{\parbox{.3cm}{\rightarrowfill}}} & \edit{\underline{\textcolor{black}{4.003}}} & 4.133 & 4.664 & \textcolor{black}{\textbf{3.851}} & 4.288 \\
    & & MSE  & & \edit{\underline{\textcolor{black}{27.02}}} & 29.44 & 36.37 & \textcolor{black}{\textbf{25.93}} & 30.11 \\
    & & MAPE  & & \edit{\underline{\textcolor{black}{0.0573}}} & 0.0594 & 0.0674 & \textcolor{black}{\textbf{0.0556}} & 0.0624 \\
    \bottomrule
    \end{tabular}%
    }
\end{table}

\newpage

\subsection{Additional \edit{Mean Performance Result Tables for Experiments}}
\edit{In this section, we include the additional average performance tables for our experiments. Each average table will represent the summary performance of the corresponding full experiment result. They each include a short takeaway conclusion, with a reference to the relevant experiment sections for further detail discussion.}

\begin{table*}[!htbp]
    \centering
    \LARGE
    \color{black}
    \captionsetup{labelfont={color=black}, textfont={color=black}, font={small}}
    \caption{\textbf{Pulse-PPG vs. PaPaGei.} Mean Performance \edit{across Classification and Regression Tasks. The best is bolded, and the second is underlined. Full results in Table~\ref{tab:exp23}}. See Section \ref{sec:exppapagei} and \ref{sec:expfieldtolab} for further discussion.}
    \label{tab:pulseppg_vs_papagei_mean}

    \resizebox{\textwidth}{!}{%
    \begin{tabular}{@{}llccccccccc@{}}
        \toprule
        & & \multicolumn{6}{c}{\textbf{Avg. Classification Performance}} 
        & \multicolumn{3}{c}{\textbf{Avg. Regression Performance}} \\
        \cmidrule(lr){3-8} \cmidrule(lr){9-11}
        \textbf{Model} & \textbf{Pre-Train Data} & {F1 Score} & {Accuracy} & {Precision} & {Recall} & {AUPRC} & {AUROC} & {MAE} & {MSE} & {MAPE} \\
        \midrule
        \multirow{2}{*}{Pulse-PPG} 
        & Wearable Field & \edit{$\boldsymbol{\textcolor{black}{0.5870} \pm 0.17}$} & \edit{$\boldsymbol{\textcolor{black}{0.6840} \pm 0.17}$} & \edit{$\boldsymbol{\textcolor{black}{0.6237} \pm 0.16}$} & \edit{$\boldsymbol{\textcolor{black}{0.5894} \pm 0.17}$} & \edit{$\boldsymbol{\textcolor{black}{0.5149} \pm 0.20}$} & \edit{$\boldsymbol{\textcolor{black}{0.7481} \pm 0.17}$} & \edit{$\boldsymbol{\textcolor{black}{8.8593} \pm 6.25}$} & \edit{$\boldsymbol{\textcolor{black}{145.2550} \pm 171.08}$} & \edit{$\boldsymbol{\textcolor{black}{0.1009} \pm 0.05}$} \\
        
        & Clinical & \edit{\underline{$\textcolor{black}{0.5217} \pm 0.16$}} & \edit{\underline{$\textcolor{black}{0.6573} \pm 0.17$}} & \edit{$\textcolor{black}{0.5741} \pm 0.18$} & \edit{$\textcolor{black}{0.5347} \pm 0.15$} & \edit{\underline{$\textcolor{black}{0.4678} \pm 0.21$}} & \edit{\underline{$\textcolor{black}{0.7174} \pm 0.14$}} & \edit{$\textcolor{black}{11.7183} \pm 9.92$} & \edit{$\textcolor{black}{250.3750} \pm 333.55$} & \edit{$\textcolor{black}{0.1406} \pm 0.13$} \\
        
        \midrule
        \multirow{2}{*}{PaPaGei} 
        & Wearable Field & \edit{$0.5196 \pm 0.15$} & \edit{$0.6447 \pm 0.14$} & \edit{\underline{$0.5774 \pm 0.14$}} & \edit{\underline{$0.5493 \pm 0.14$}} & \edit{$0.4307 \pm 0.15$} & \edit{$0.6981 \pm 0.14$} & \edit{\underline{$10.0973 \pm 7.16$}} & \edit{\underline{$183.4350 \pm 212.98$}} & \edit{\underline{$0.1169 \pm 0.07$}} \\
        
        & Clinical & \edit{$0.4994 \pm 0.14$} & \edit{$0.6361 \pm 0.15$} & \edit{$0.5530 \pm 0.14$} & \edit{$0.5210 \pm 0.14$} & \edit{$0.4164 \pm 0.12$} & \edit{$0.6840 \pm 0.10$} & \edit{$10.7488 \pm 7.75$} & \edit{$204.6825 \pm 237.78$} & \edit{$0.1288 \pm 0.10$} \\
        
        \bottomrule
    \end{tabular}
    }
\end{table*}

\begin{table*}[!htbp]
    \centering
    \LARGE
    \color{black}
    \captionsetup{
        labelfont={color=black},
        textfont={color=black},
        font={small}
    }
    \caption{\textbf{Naive vs. Pulse-PPG Linear Probe vs. Pulse-PPG 
 Fine-Tuning.} Mean Performance \edit{across Classification and Regression Tasks. The best is bolded, and the second is underlined. Full results in Table~\ref{tab:exp1}}. See Section \ref{sec:finetuneeval} for further discussion.}
    \label{tab:linearprobe_vs_finetuning_vs_naive}

    \resizebox{\textwidth}{!}{%
    \begin{tabular}{llccccccccc}
        \toprule
        & & \multicolumn{6}{c}{\textbf{Avg. Classification Performance}} 
        & \multicolumn{3}{c}{\textbf{Avg. Regression Performance}} \\
        \cmidrule(lr){3-8} \cmidrule(lr){9-11}
        \textbf{Eval Method} & \textbf{Pre-Train Data} & {F1 Score} & {Accuracy} & {Precision} & {Recall} & {AUPRC} & {AUROC} & {MAE} & {MSE} & {MAPE} \\
        \midrule
        Naive          & -     & \edit{$\textcolor{black}{0.3265} \pm 0.15$} & 
        \edit{$\textcolor{black}{0.6139} \pm 0.19$} & 
        \edit{$\textcolor{black}{0.2774} \pm 0.14$} & 
        \edit{$\textcolor{black}{0.4087} \pm 0.15$} & 
        \edit{$\textcolor{black}{0.2455} \pm 0.10$} & 
        \edit{$\textcolor{black}{0.5000} \pm 0.00$} & 
        \edit{$\textcolor{black}{13.6587} \pm 8.71$} & 
        \edit{$\textcolor{black}{306.1450} \pm 340.85$} & 
        \edit{$\textcolor{black}{0.1668} \pm 0.12$}  \\
        Linear Probe   & Wearable Field & \edit{\underline{$\textcolor{black}{0.5870} \pm 0.17$}} & 
        \edit{\underline{$\textcolor{black}{0.6840} \pm 0.17$}} & 
        \edit{\underline{$\textcolor{black}{0.6237} \pm 0.16$}} & 
        \edit{\underline{$\textcolor{black}{0.5894} \pm 0.17$}} & 
        \edit{\underline{$\textcolor{black}{0.5149} \pm 0.20$}} & 
        \edit{\underline{$\textcolor{black}{0.7481} \pm 0.17$}} & 
        \edit{\underline{$\textcolor{black}{8.8593} \pm 6.25$}} & 
        \edit{\underline{$\textcolor{black}{145.2550} \pm 171.08$}} & 
        \edit{\underline{$\textcolor{black}{0.1009} \pm 0.05$}}  \\
        Finetuning     & Wearable Field & \edit{$\boldsymbol{\textcolor{black}{0.6497} \pm 0.17}$} & 
        \edit{$\boldsymbol{\textcolor{black}{0.7191} \pm 0.16}$} & 
        \edit{$\boldsymbol{\textcolor{black}{0.6571} \pm 0.15}$} & 
        \edit{$\boldsymbol{\textcolor{black}{0.6621} \pm 0.18}$} & 
        \edit{$\boldsymbol{\textcolor{black}{0.6769} \pm 0.18}$} & 
        \edit{$\boldsymbol{\textcolor{black}{0.7905} \pm 0.13}$} & 
        \edit{$\boldsymbol{\textcolor{black}{7.1068} \pm 6.69}$} & 
        \edit{$\boldsymbol{\textcolor{black}{120.8925} \pm 180.97}$} & 
        \edit{$\boldsymbol{\textcolor{black}{0.0791} \pm 0.06}$} \\
        \bottomrule
    \end{tabular}%
    }
\end{table*}

\end{document}